\documentclass[journal]{IEEEtran}
\usepackage{times}
\usepackage{dsfont}
\usepackage{epsfig}
\usepackage{graphics}
\usepackage{amsmath}
\usepackage{float}
\usepackage{amssymb}
\usepackage{subfigure}
\usepackage{indentfirst}
\usepackage{fancyhdr}
\usepackage{multicol}
\usepackage{color}
\usepackage{fullpage,authblk}
\begin{document}
\title{Early Fire Detection Using HEP and Space-time Analysis}
\author{CHEN~Junzhou, YOU Yong}

\affil{School of Information Science \& Technology, Southwest Jiaotong University,\\
  Chengdu, Sichuan, 610031, China\\
 Email:jzchen@swjtu.edu.cn}

\date{}
\maketitle \thispagestyle{empty}
\pagestyle{empty}

\begin{abstract}
In this article, a video base early fire alarm system is developed by monitoring the smoke in the scene. There are two major contributions in this work. First, to find the best texture feature for smoke detection, a general framework, named \emph{Histograms of Equivalent Patterns} (HEP), is adopted to achieve an extensive evaluation of various kinds of texture features. Second, the \emph{Block based Inter-Frame Difference} (BIFD) and  a improved version of LBP-TOP are proposed and ensembled to describe the space-time characteristics of the smoke. In order to reduce the false alarms, the \emph{Smoke History Image} (SHI) is utilized to register the recent classification results of candidate smoke blocks. Experimental results using SVM show that the proposed method can achieve better accuracy and less false alarm compared with the state-of-the-art technologies.
\end{abstract}
\begin{IEEEkeywords}
smoke detection, dynamic texture, BIFD, improved LBP-TOPs, SHI.
\end{IEEEkeywords}
\section{Introduction}
As one of the leading hazards, fire causes grave losses of life and property around the world each year. In order to detect and eliminate fire at very early stage, during the past decades, different kinds of technologies are developed by researchers. However, traditional methods usually depend on heat sensors, optical sensors (ultraviolet, infrared), ion sensors, or thermocouples. These sensors suffer from the transport delay of the fire (heat or smoke particles) to the sensor, and cannot work effectively if the fire is far away. Moreover, these sensors can not provide the situation and development of the fire for validation and rescue.

With the increasingly wide usage of video surveillance systems, in recent years, video based fire detection technologies received much attentions from researchers. Smoke is usually generated before flames and can be observed from a great distance, it is an important sign for early fire detection. Since the video camera can accomplish real-time monitoring in a large range, the smoke of uncontrolled fire can be easily observed by it. Therefore, video based technologies are much more competent in fire detection especially for large and open areas.

Many methods are proposed for fire detection in the field of view of the camera \cite{Getin20131827}. Color models of fire and smoke pixels was introduced by Chen et al. \cite{chen2004early} \cite{chen2006smoke}, the estimated candidate fire and smoke regions were verified by dynamical measuring of their growth and disorder. False alarms are inevitable if the scene has fire-colored moving objects due to no usage of texture information. Toreyin et al. \cite{Toreyin2005wavelet} proposed a wavelet based real-time smoke detection algorithm, in which both temporal and spatial wavelet transformations were employed. The temporal wavelet transformation was used to analysis the flicker of smoke like objects, while the spatial wavelet transformation was implemented to calculate the decrease in high frequency content corresponding to edges caused by the blurring effect of smoke. The boundary roughness \cite{Toreyin2006contour} was introduced to improve the performance of \cite{Toreyin2005wavelet}, three-state Markov models were trained to discriminate between smoke and non-smoke pixels. Liu and Ahuja \cite{celik2007fire} proposed spectral, spatial and temporal models for fire detection. Fourier Descriptors (FD) were used to represent fire shapes but FD is sensitive to noise. Celik et al. Yuan \cite{yuan2008fast} presented a real-time video smoke detection method using an accumulative motion model, but it can not detect smoke drifting in any direction.
In \cite{Zhang2009dynamic} Zhang et al. presented a real-time forest fire detection system using dynamic characteristics of fire regions. Maruta et al. \cite{Maruta2010smoke} considered that the image information of smoke is a self-affine fractal, local Hurst exponent was extracted to analyze the self-similarity of suspect smoke region. In \cite{yuan2011lbpv}, Yuan proposed an effective feature vector by concatenating the histogram sequences of Local Binary Pattern (LBP) and Local Binary Pattern Variance (LBPV) pyramids, and a BP neural network was used for smoke detection. Ko et al. \cite{ko2009fire} detected candidate fire regions by motion and color, and created a luminance map to remove non-fire pixels. SVM classifier was used for final decision. In \cite{Yuan2012double}, a double mapping framework is proposed by Yuan to extract partition based features of smoke, AdaBoost was employed to enhance the performance of classification.

Due to the arbitrary shapes of smoke, illumination and intra-class variations, occlusions and etc., video base smoke detection is still a challenging task. Being enlightened on the fact that human can easily recognize the existence of smoke only according to dynamic characteristics observed, in this paper, we concentrate on study the dynamic characteristics feature extraction methods for video based smoke detection. The remainder of this paper is organized as follows. Section \ref{sec:method} presents the details of the algorithms. Section \ref{sec:result} illustrates the performance of the proposed method is evaluated and compared with some state-of-the-art technologies. In section \ref{sec:conclusion}, a brief conclusion is presented.
\section{Algorithms} \label{sec:method}
The proposed video smoke detection algorithm is depicted as the following:
\begin{enumerate}
  \item Detect the candidate smoke blocks by color and motion.
  \item Verify the candidate smoke blocks by the accumulative motion orientation.
  \item Verify the candidate smoke blocks by the texture feature based on \emph{Histograms of Equivalent Patterns} (HEP).
  \item Verify the candidate smoke blocks by space-time feature.
  \item Verify the candidate smoke blocks by \emph{Smoke History Image} (SHI).
\end{enumerate}

It worth to mentioned that we used the SVM \cite{chang2011libsvm} classifier for the experimental evaluation, the accuracy of the SVM is largely dependent on the selection of its parameters, particularly $C$ and $Gamma$. Therefore, we adopted the five couples of $C$ and $Gamma$ which have been used in texture classification problems. These five couples include Kim's \cite{kim2002svm} (2, 100), Rajpoot's \cite{rajpoot2004wavelets} (0.001, 1), Li's \cite{li2003texture} (50, 1000), (0.5, 1000) and (0.02, 1000).
At the same time, we randomly select half of dataset for training and the left for test. Each couple parameters repeat 10 times, then compute the mean accuracy as the final accuracy.

\subsection{Candidate smoke block detection} \label{sec:csbd}
In order to get candidate smoke regions for further estimation, moving detection and characteristics of smoke color are used. Each video frame is divided into non-overlapped blocks. Supposed that image width, image height, block width and block height are $I_w$, $I_h$, $B_w$ and $B_h$, respectively. The row number $N_r$ and column number $N_c$ of blocks in each frame are calculated by:
\begin{equation}\label{eq:NrNc}
    N_r = \lfloor\frac{I_h}{B_h}\rfloor
\end{equation}

\begin{equation}\label{eq:NrNc}
    N_c = \lfloor\frac{I_w}{B_w}\rfloor
\end{equation}

where $\lfloor\;\rfloor$ is an operator of truncating a floating point number to get an integer number.

Block $B_{ij}$ of $t$-th frame is regard as a moving block if
\begin{equation}\label{eq:block_diff}
\sum_{(x,j)\in B_{ij}}{[f(x,y,t)-f(x,y,t-1)]} > T_B
\end{equation}

where $T_B$ is a predetermined threshold, and block $B_{ij}$ is on the $i$-th row and $j$-th column in the video.

Then, the estimated moving blocks are verified by several rules of smoke color in RGB color space \cite{yuan2008fast}.

\subsection{Accumulative Motion Orientation} \label{sec:amo}
\begin{figure}
\center
  \includegraphics[width=3in]{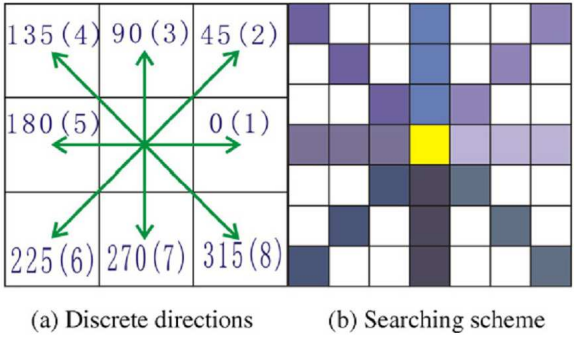}\\
  \caption{Discrete direction  and searching scheme of AOM}\label{fig:aom}
\end{figure}

Since the smoke of fire commonly move upward due to high temperature, motion orientation information is helpful for fire detection. In this paper, the block based Accumulative Motion Orientation (AMO) proposed in \cite{yuan2008fast} is adopted to analysis motion of the candidate smoke regions. The orientation of motion is discretized into 8-directions: 0, 45, 90, 135, 180, 225, 270 and 315 degrees. These discrete directions are coded as 1, 2, 3, 4, 5, 6, 7 and 8, respectively. The 3-pixel searching displacement as described in Fig.\ref{fig:aom} is applied at each direction to find the sift orientation of the candidate smoke regions. The estimated temporal motion orientation histogram $\mathcal{H}(i, j, t)$ were accumulated in a sliding time window $W_t$ to enhance the accuracy. The ratio of the sum of frequencies at upward motion directions to the sum at all directions are calculated by equation (\ref{eq:umr}), where $\mathcal{H}_B$ is the histogram of accumulative motion orientation of block $B$. The block with $UMR<T_U$ is regard as non-smoke. Where $T_U$ is a predefined threshold. In our implementation, $T_U = 0.55$ is used.

\begin{equation}\label{eq:umr}
    UMR = \frac{\sum_{\theta = 2}^4 \mathcal{H}_B(\theta)}{\sum_{\theta = 1}^8\mathcal{H}_B(\theta)}
\end{equation}

\subsection{Texture descriptor for smoke block classification} \label{sec:hep}
The candidate smoke blocks are further inspected by using texture feature. In order to find a proper texture descriptor, \emph{Histograms of Equivalent Patterns} (HEP) \cite{fernandez2013texture}, is adopted to achieve a comprehensive evaluation of texture features for smoke detection. In HEP, a texture descriptor is defined as a function $F$ that receives an image $I$ and returns a vector $h$.
\begin{equation}\label{eq:texture_desc}
    h = F(I)
\end{equation}

The k-th element of $h$ can be expressed as the following:
\begin{equation}\label{eq:hep_def}
h_{k}={\frac{1}{D}}{{\sum_{m=m_{min}}^{m_{max}}\sum_{n=n_{min}}^{n_{max}}\delta[f(x^\Omega_{m,n},\theta)-k]}}
\end{equation}
Where $m$ and $n$ represent row- and column-wise pixel indices, $x^\Omega_{m,n}$ is the neighbourhood $\Omega_{m,n}$ around (m,n). $\theta$ a vector of parameters computed from the whole image. $D$ a normalizing factor, $f$ a generic function that returns an integer between $0$ and $K$. $\delta$ the function defined as bellow:
\begin{equation}\label{eq:SHID}
\begin{split}
     \delta(x) = &\left\{\begin{aligned}
     1 & & x=0 \\
     0 & & Otherwise \\
                    \end{aligned}\right.
\end{split}
\end{equation}

\begin{figure*}
\centering
\subfigure[]
{
      \includegraphics[width=0.5in]{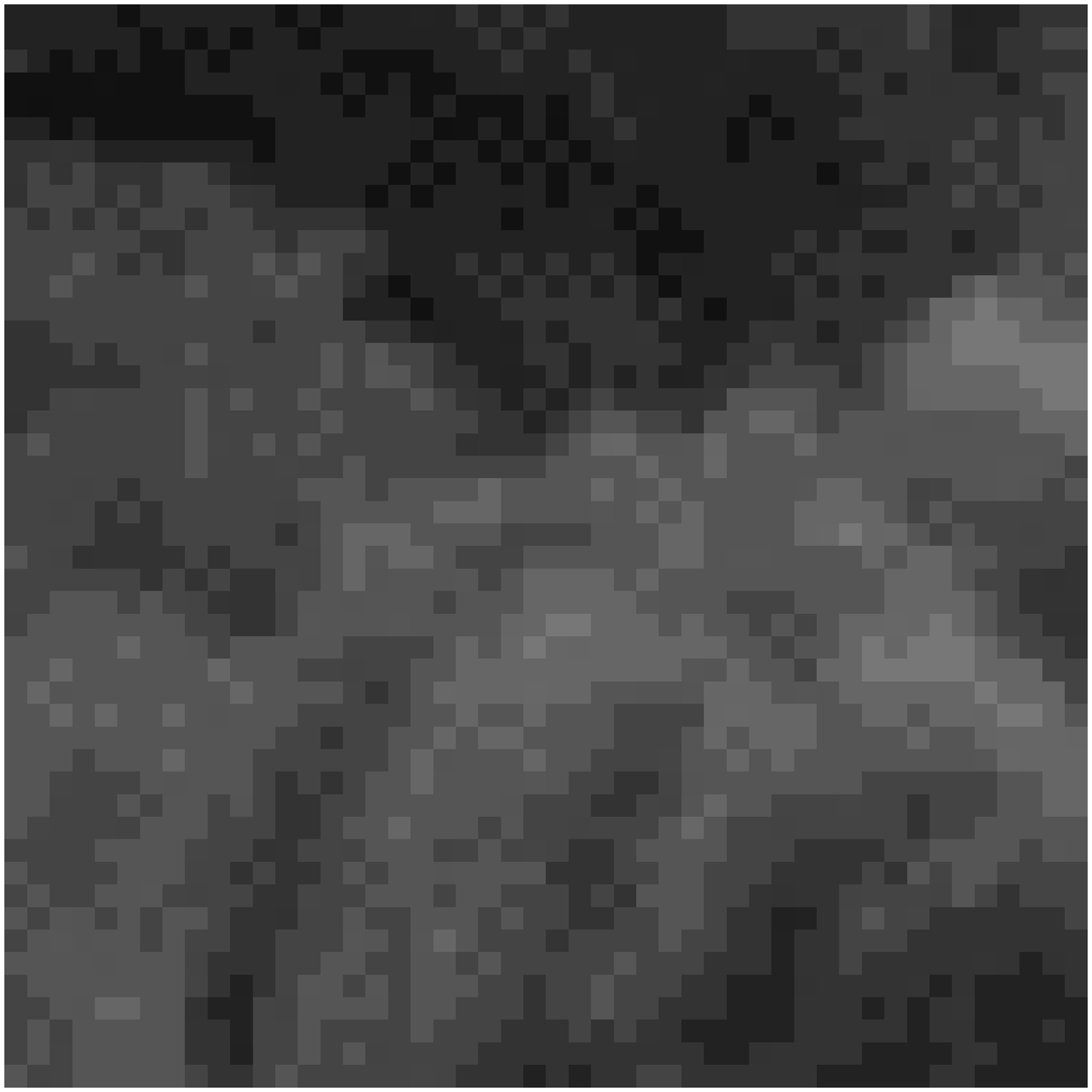}
      \label{fig_hep_dataset_1}
}
\subfigure[]
{
      \includegraphics[width=0.5in]{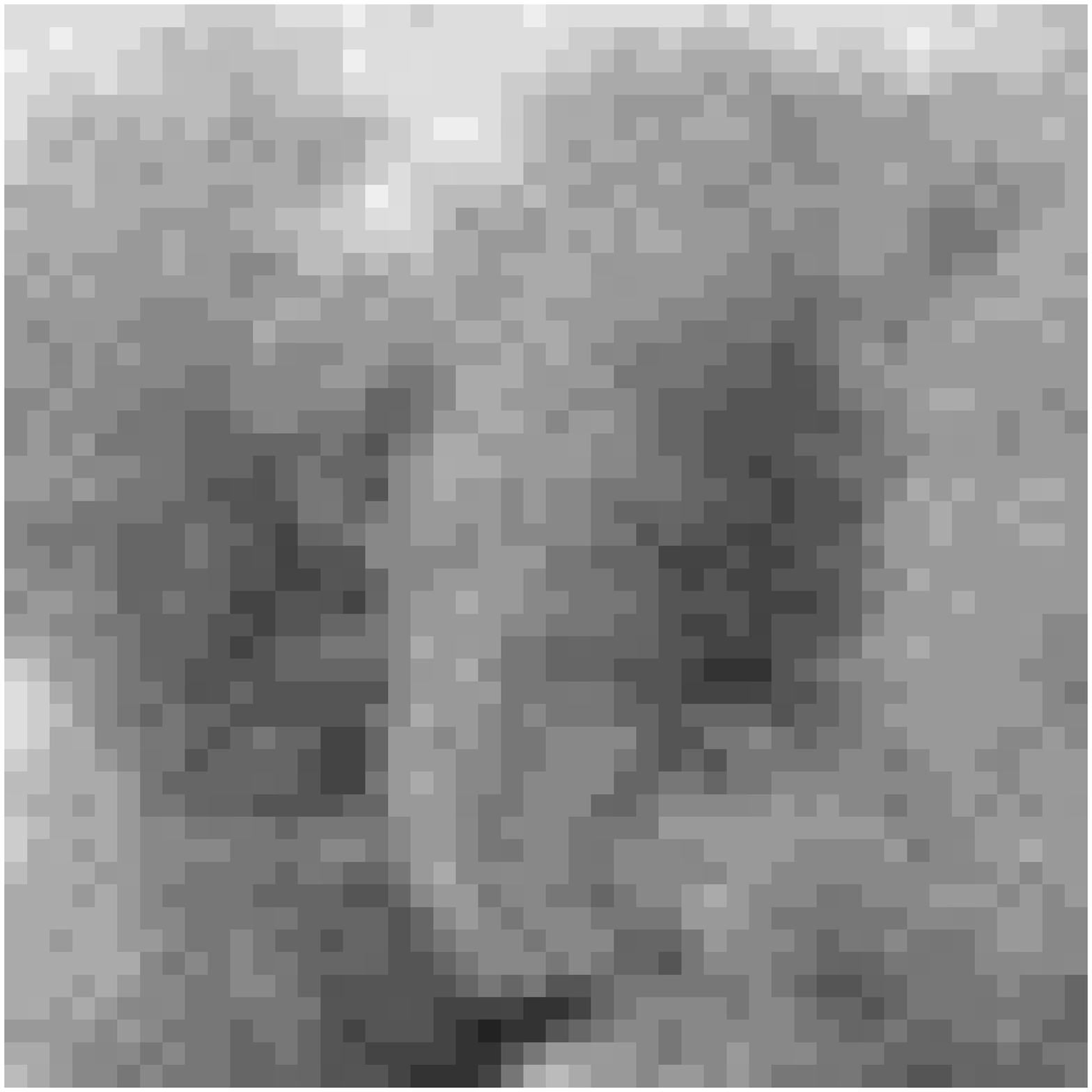}
      \label{fig_hep_dataset_2}
}
\subfigure[]
{
      \includegraphics[width=0.5in]{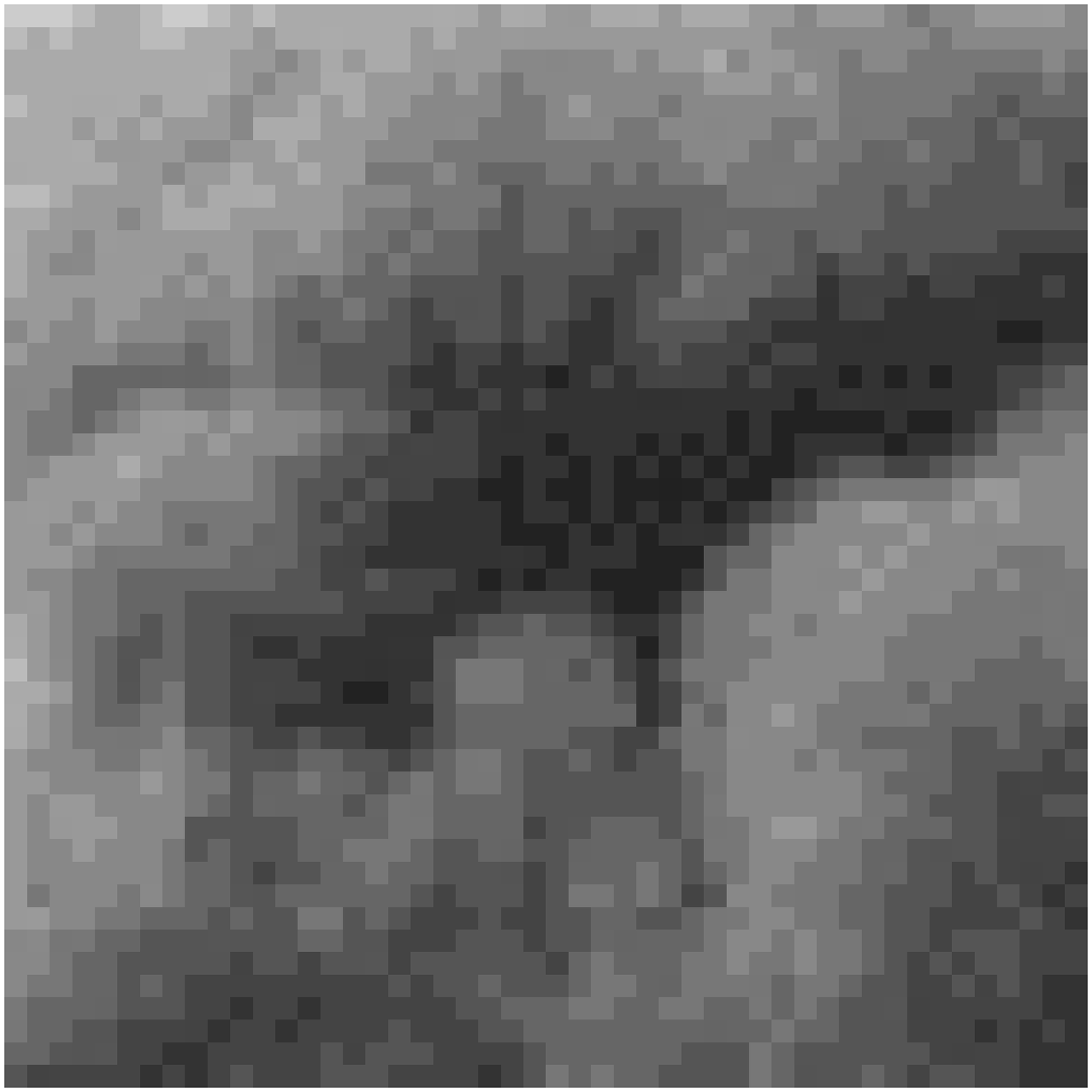}
      \label{fig_hep_dataset_3}
}
\subfigure[]
{
      \includegraphics[width=0.5in]{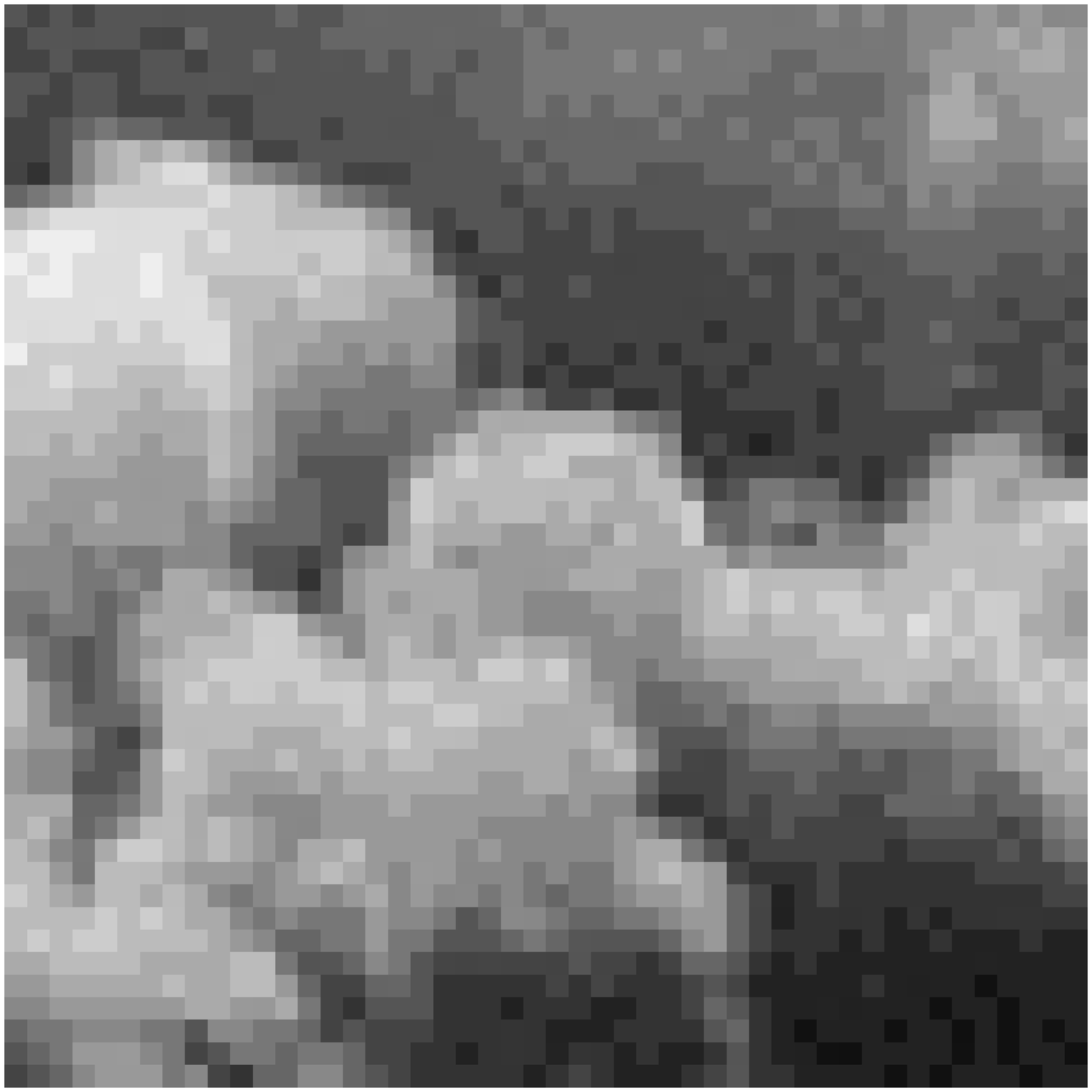}
      \label{fig_hep_dataset_4}
}
\subfigure[]
{
      \includegraphics[width=0.5in]{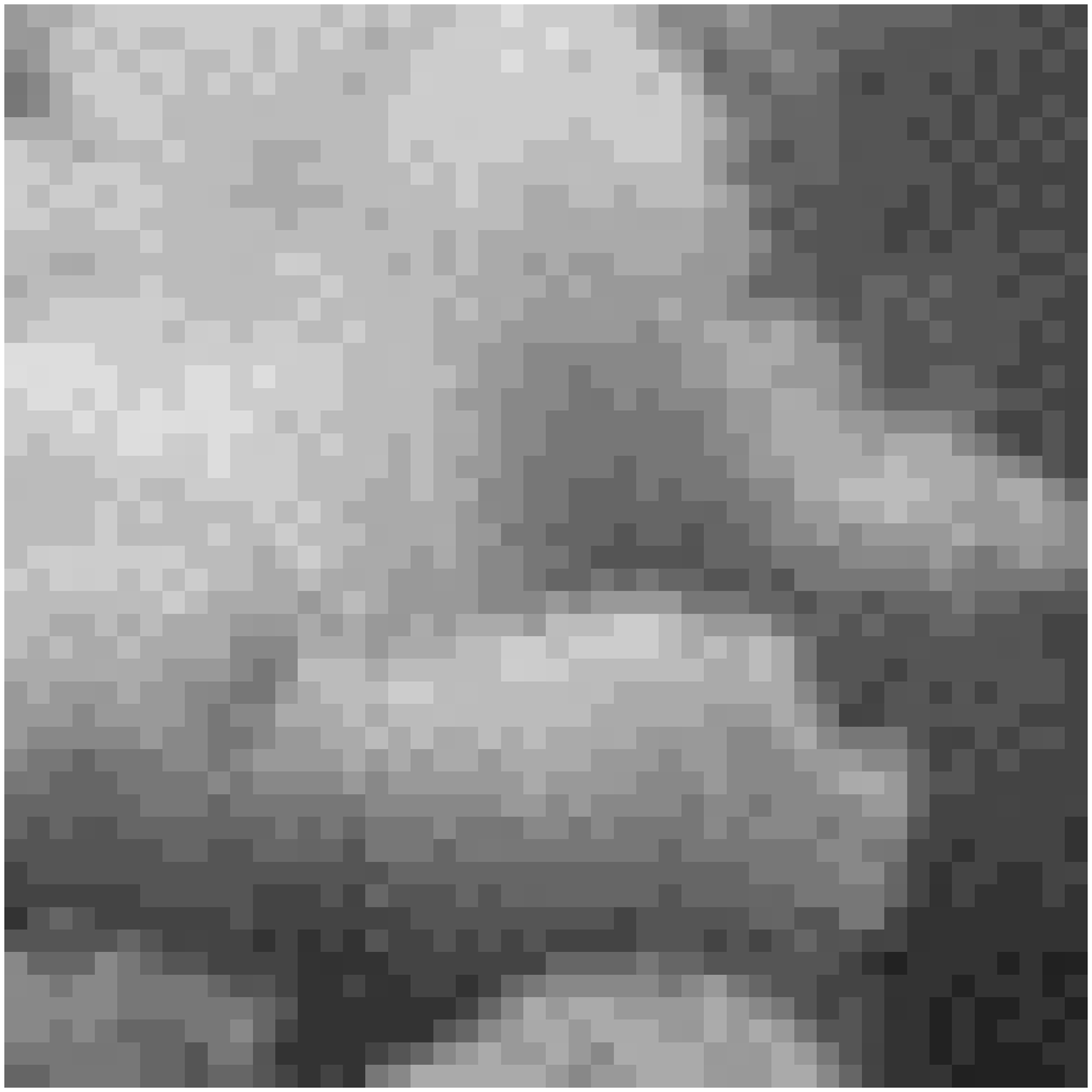}
      \label{fig_hep_dataset_5}
}
\subfigure[]
{
      \includegraphics[width=0.5in]{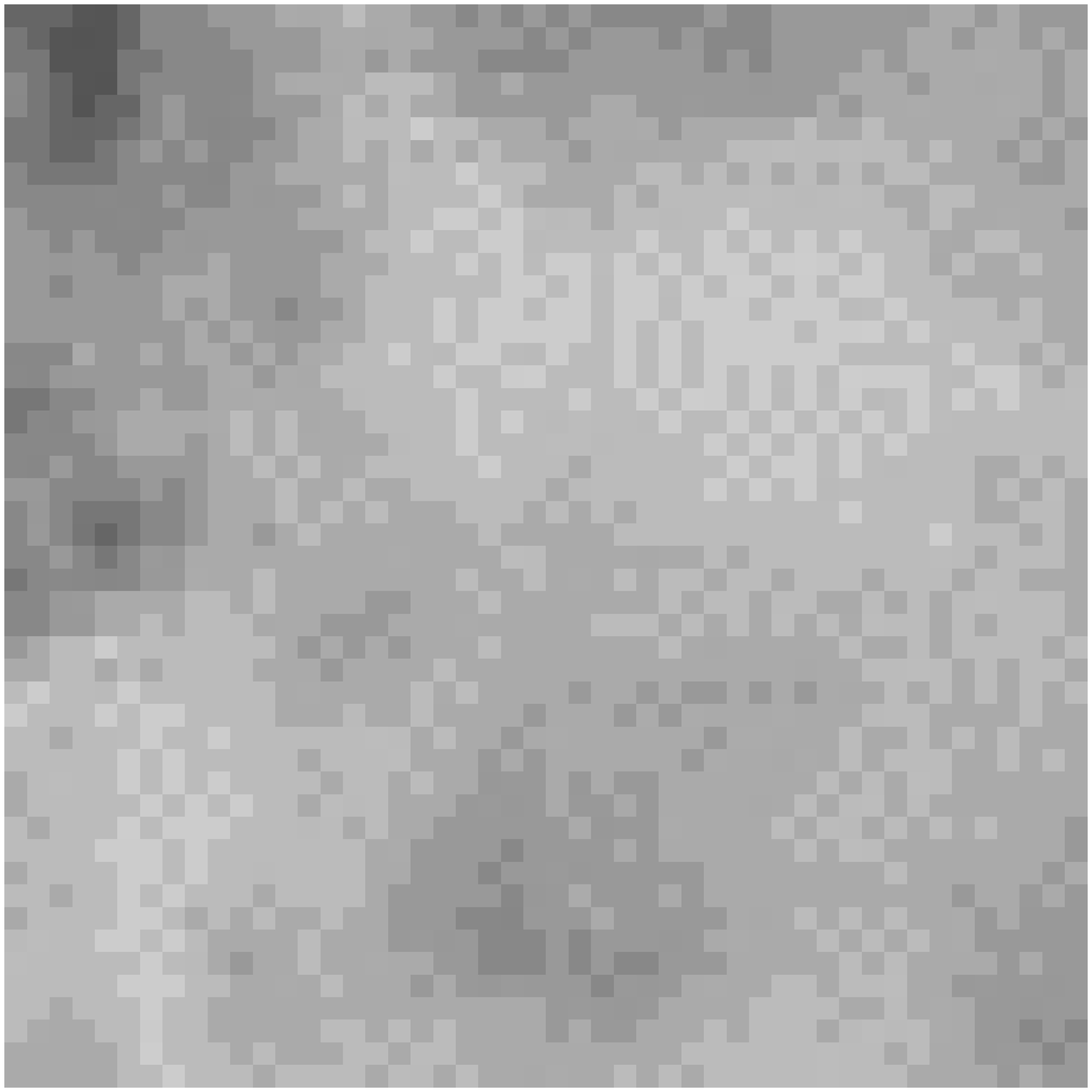}
      \label{fig_hep_dataset_6}
}
\subfigure[]
{
      \includegraphics[width=0.5in]{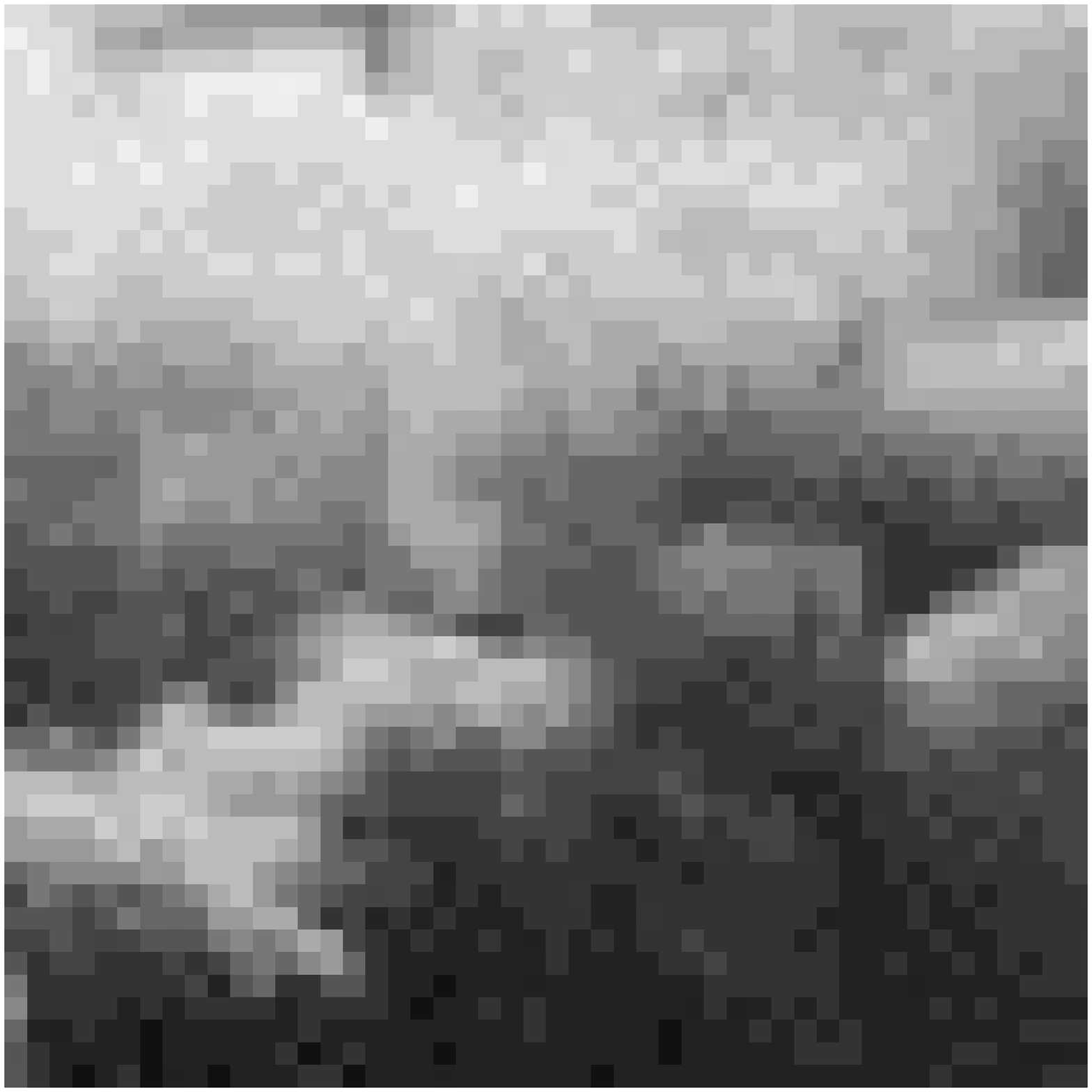}
      \label{fig_hep_dataset_7}
}
\subfigure[]
{
      \includegraphics[width=0.5in]{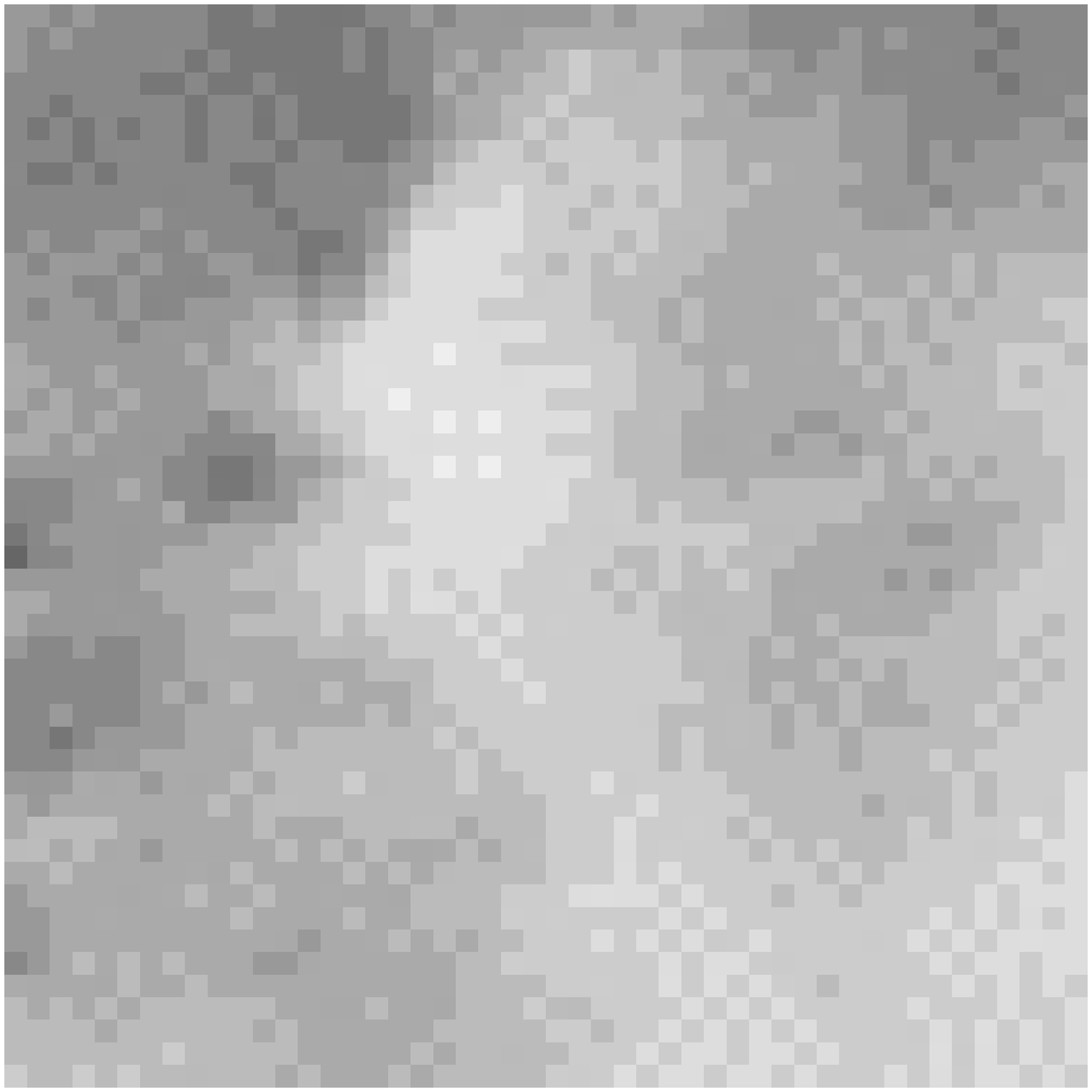}
      \label{fig_hep_dataset_8}
}
\subfigure[]
{
      \includegraphics[width=0.5in]{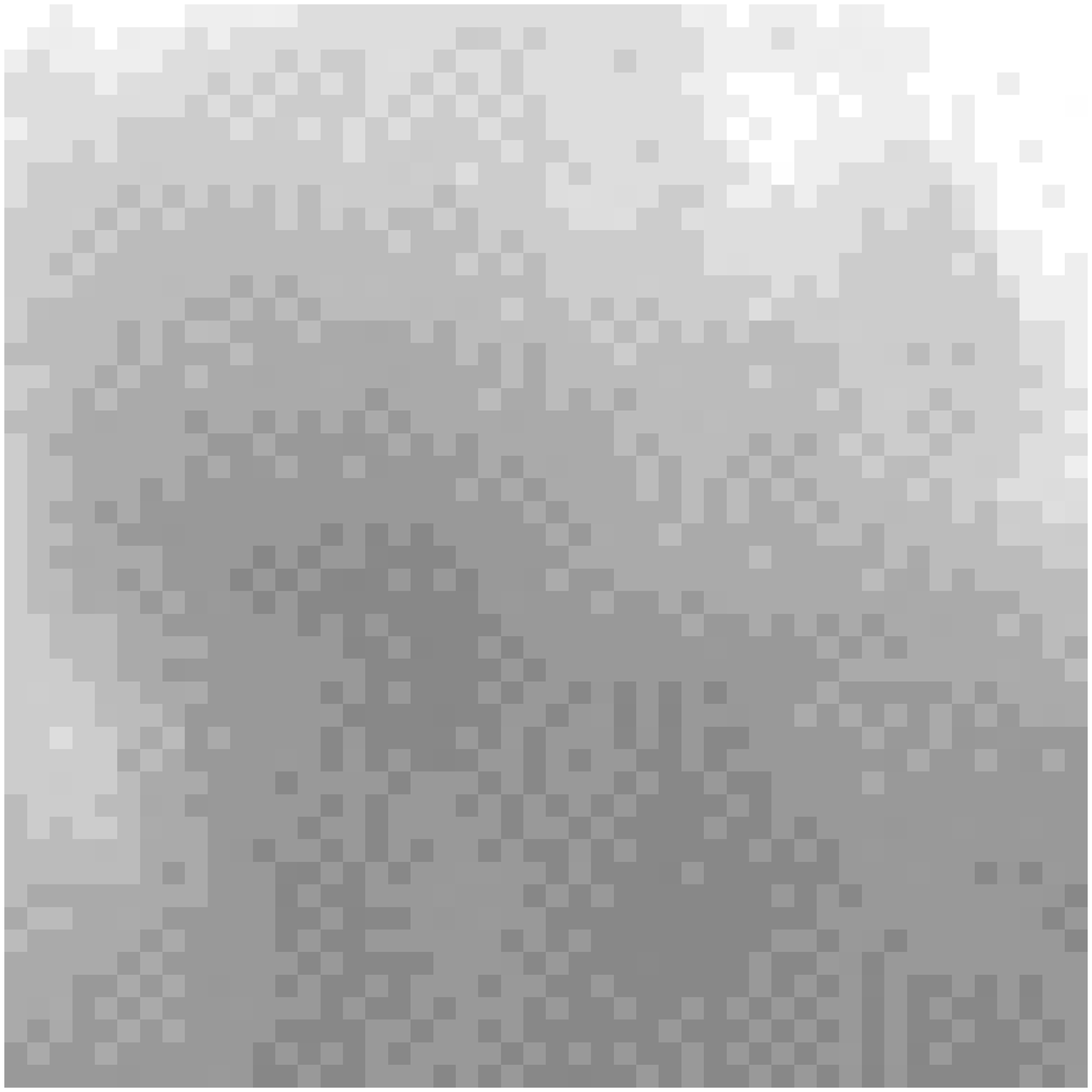}
      \label{fig_hep_dataset_9}
}
\subfigure[]
{
      \includegraphics[width=0.5in]{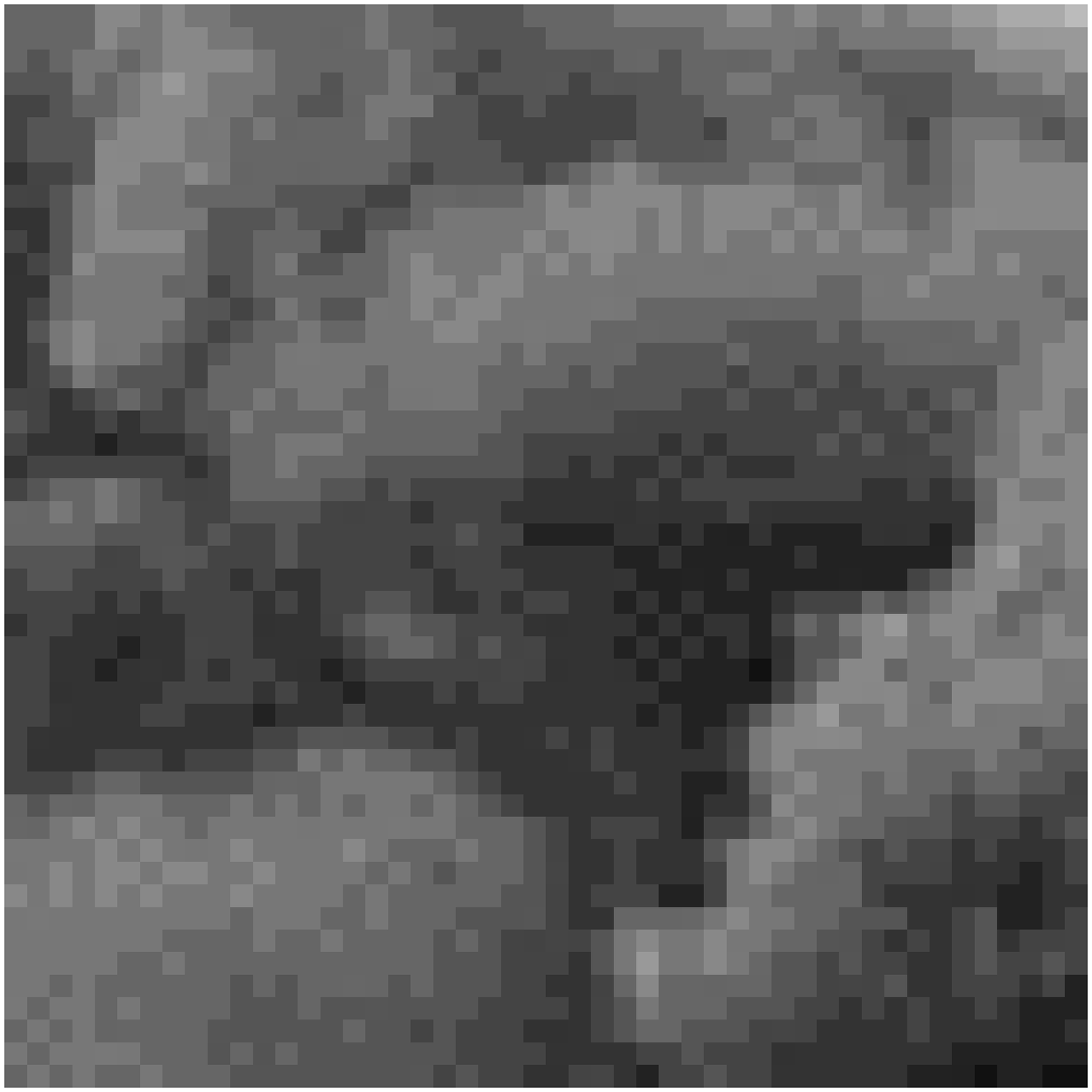}
      \label{fig_hep_dataset_10}
}
\subfigure[]
{
      \includegraphics[width=0.5in]{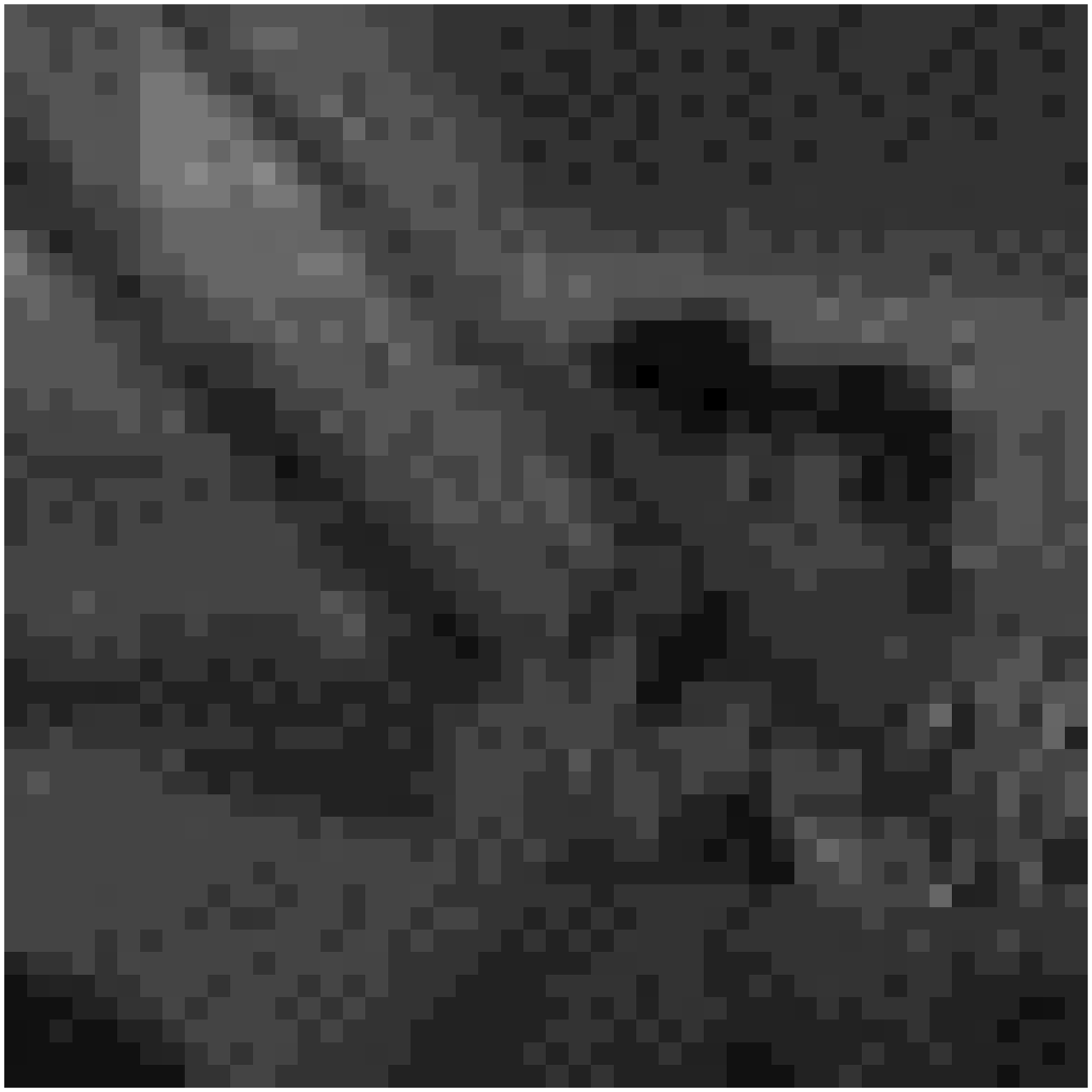}
      \label{fig_hep_dataset_11}
}
\subfigure[]
{
      \includegraphics[width=0.5in]{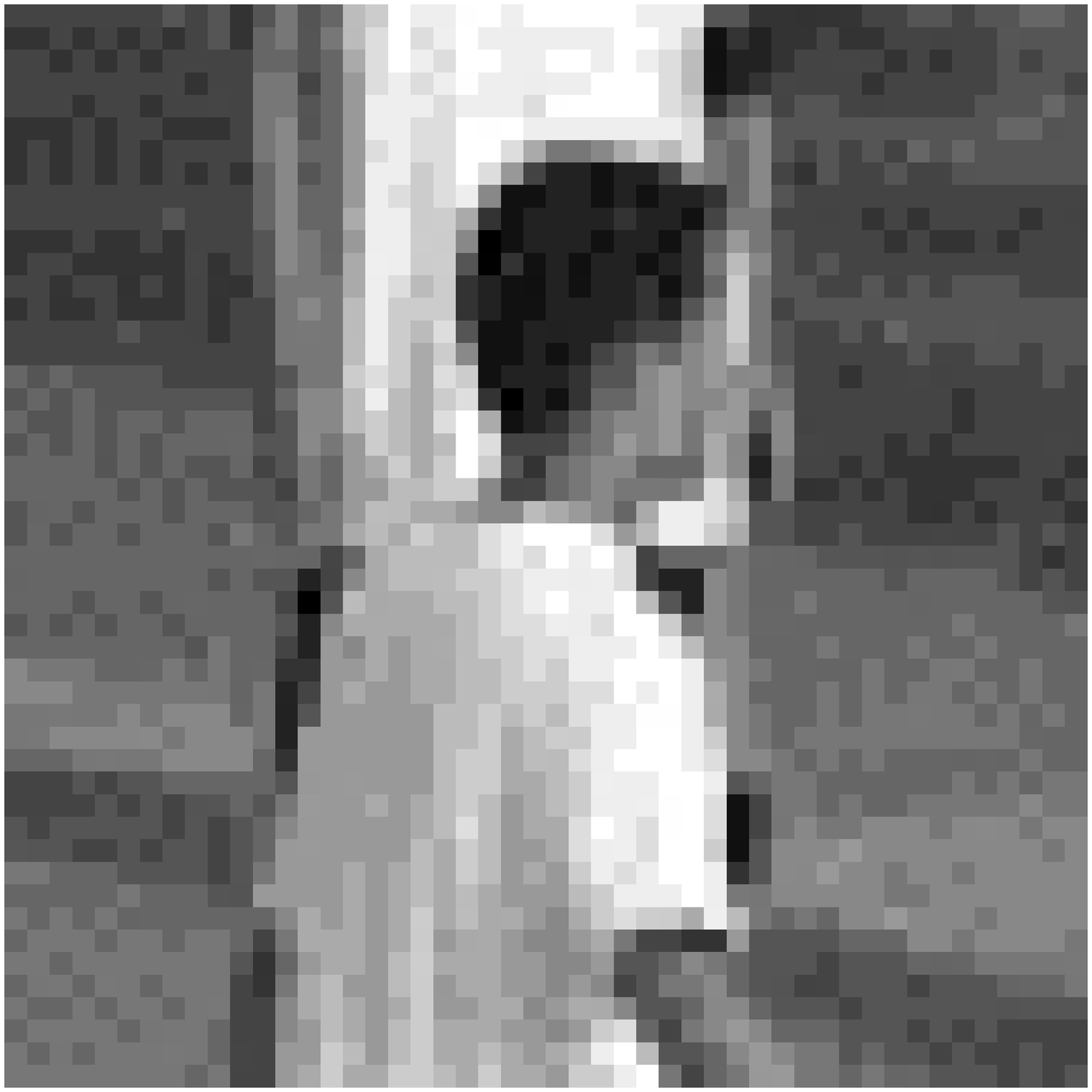}
      \label{fig_hep_dataset_12}
}
\subfigure[]
{
      \includegraphics[width=0.5in]{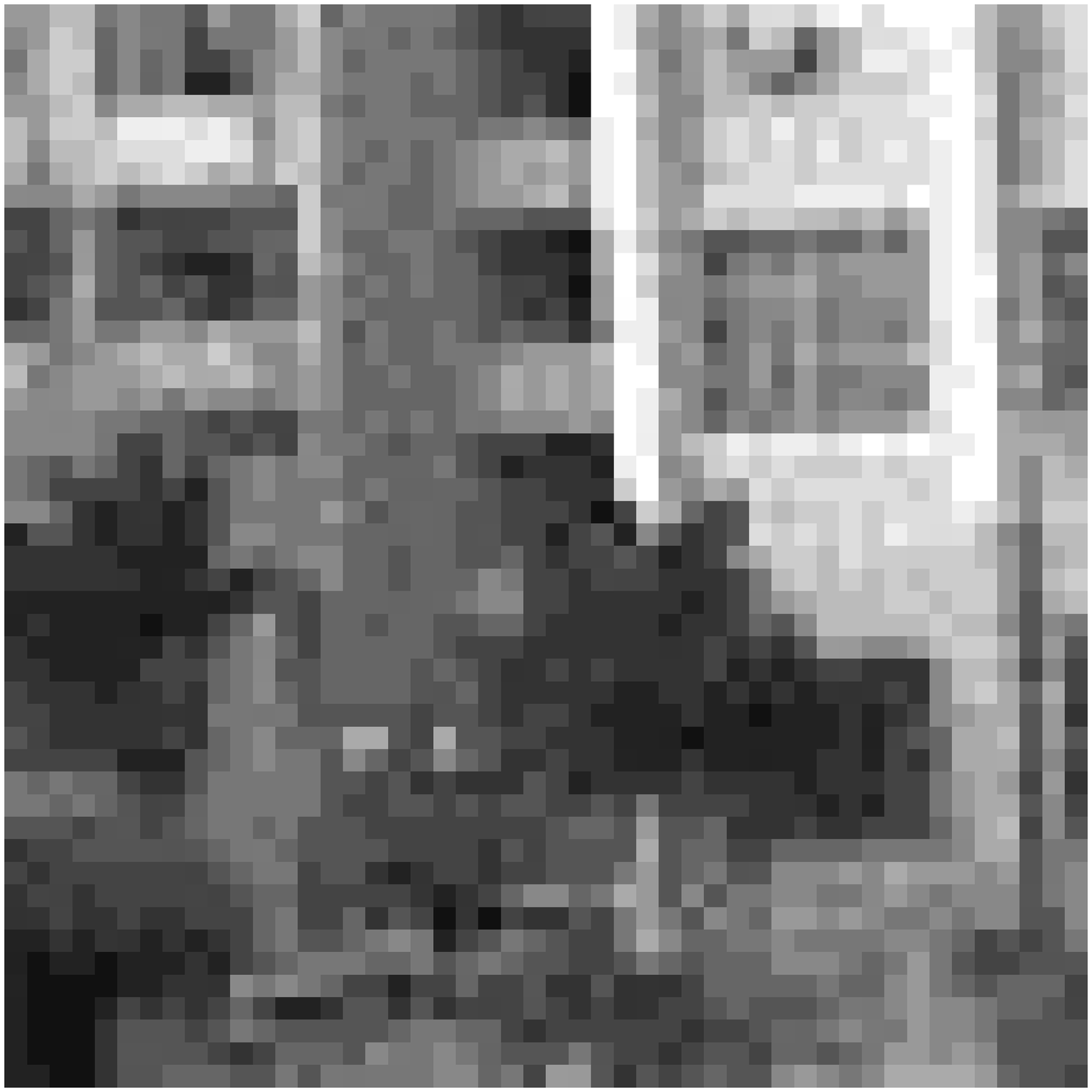}
      \label{fig_hep_dataset_13}
}
\subfigure[]
{
      \includegraphics[width=0.5in]{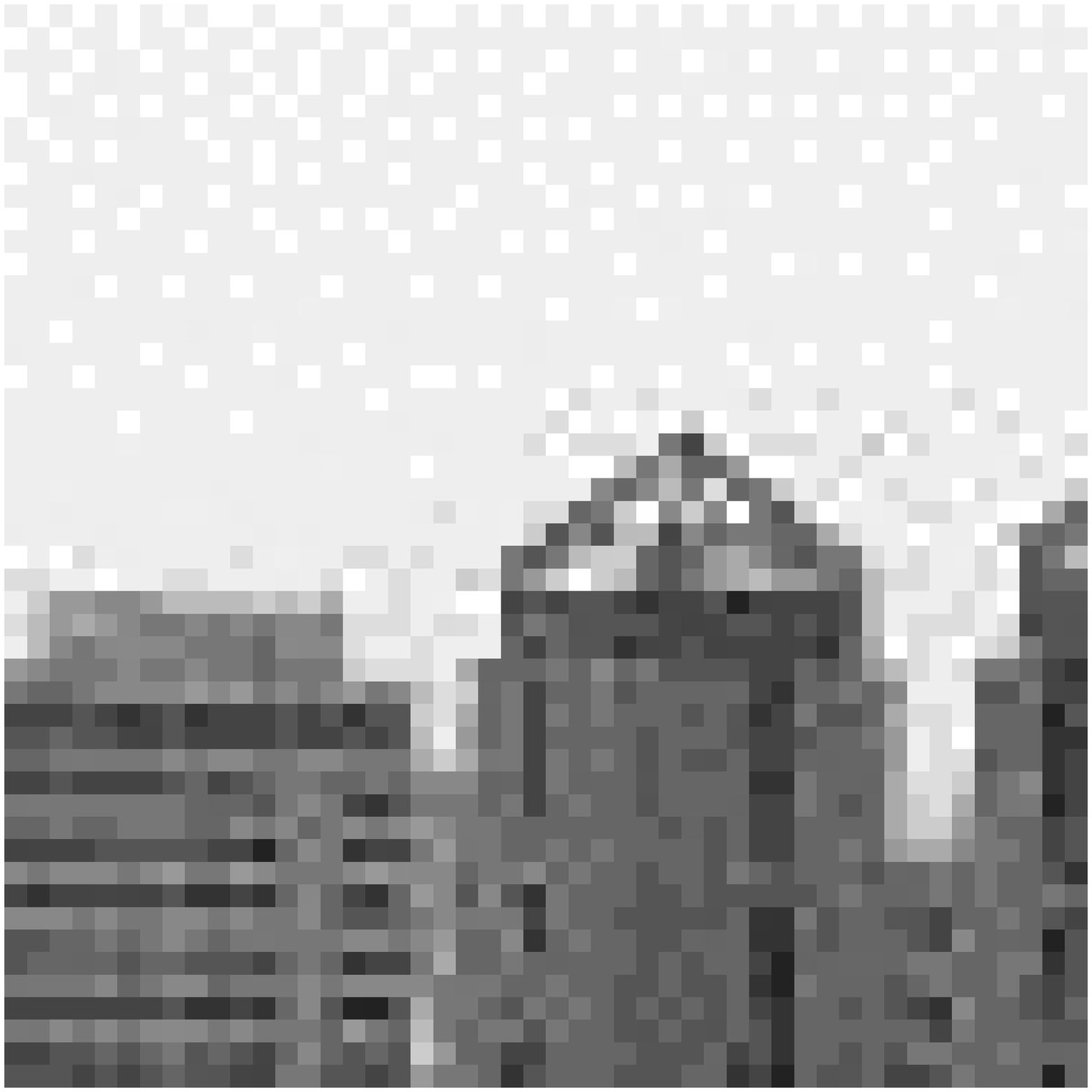}
      \label{fig_hep_dataset_14}
}
\subfigure[]
{
      \includegraphics[width=0.5in]{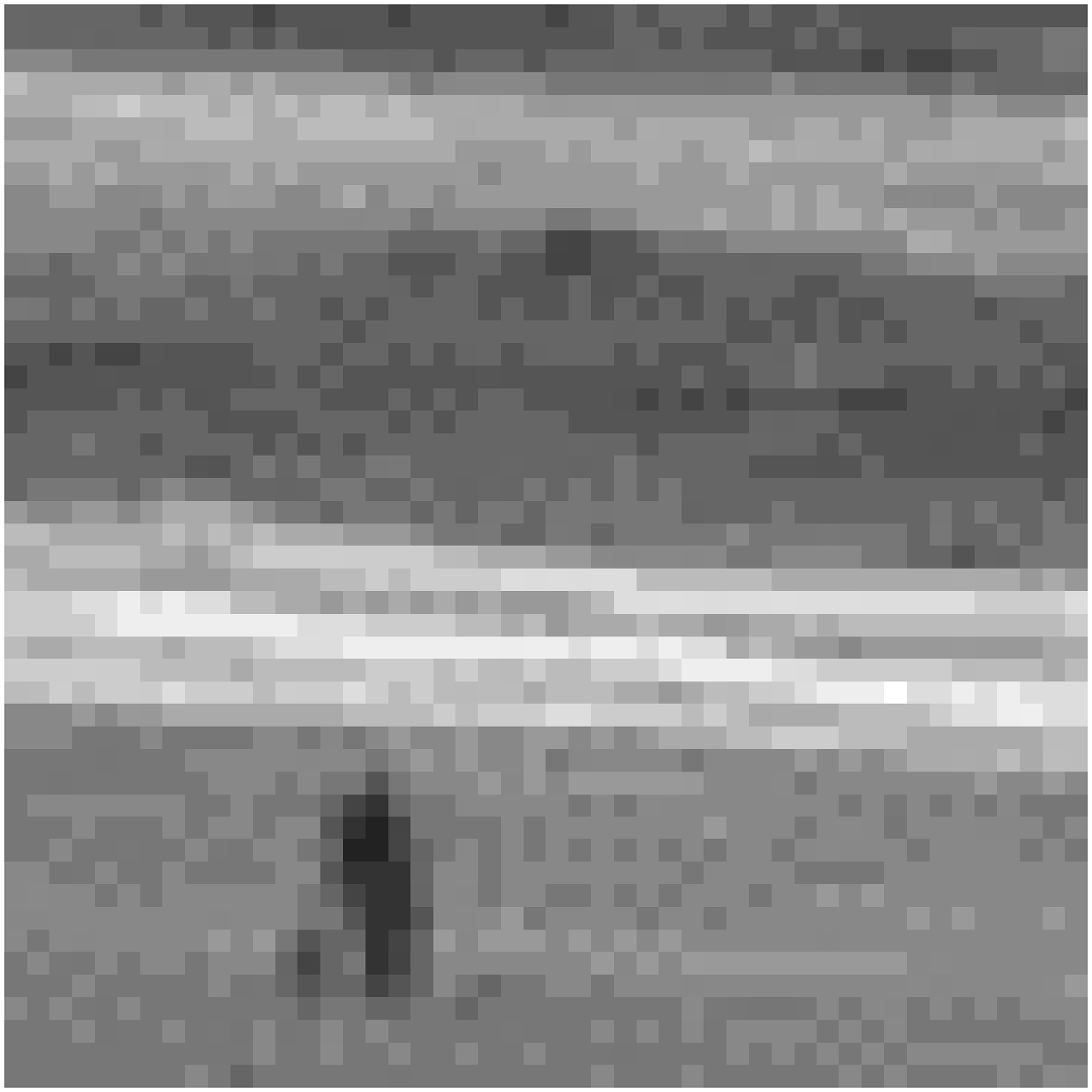}
      \label{fig_hep_dataset_15}
}
\subfigure[]
{
      \includegraphics[width=0.5in]{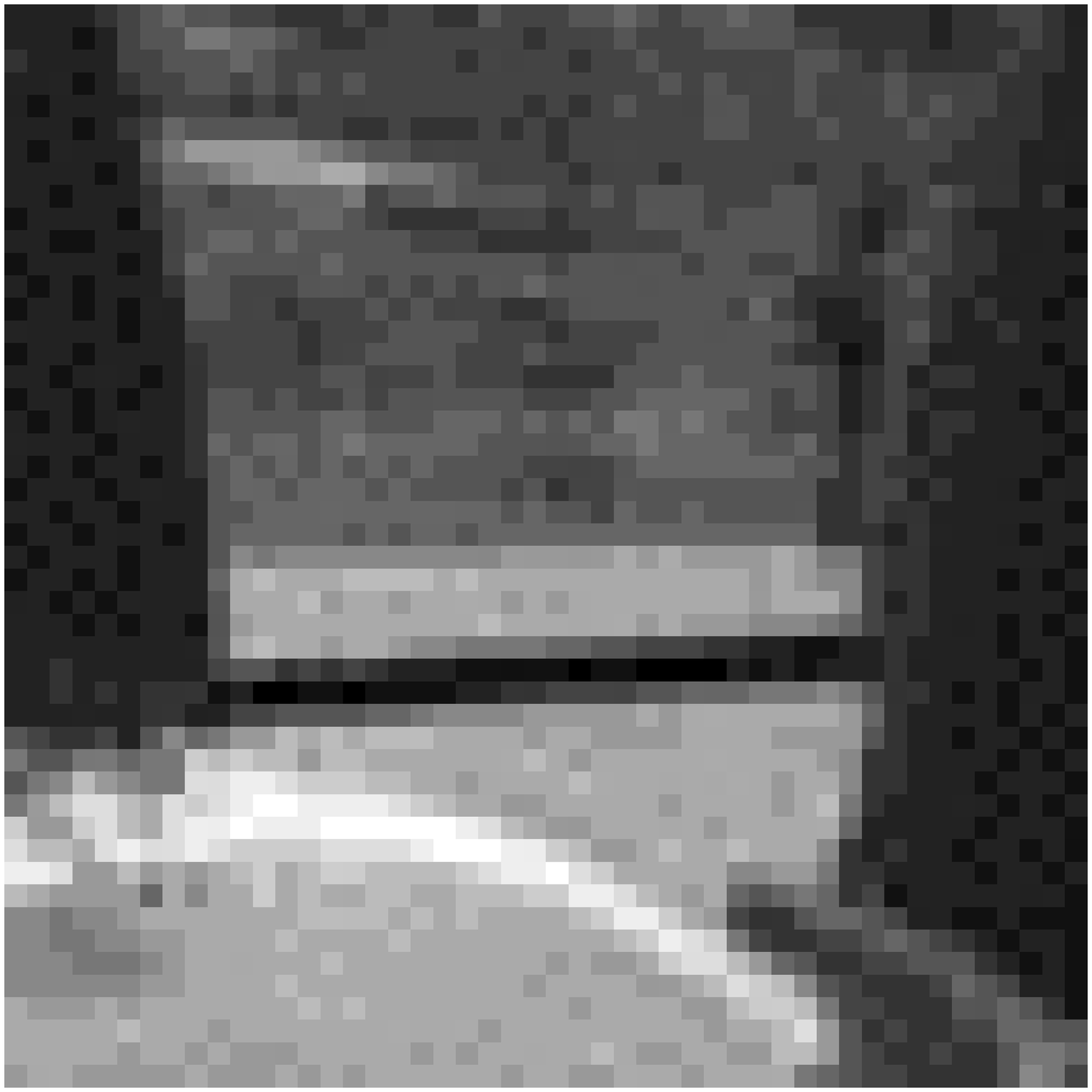}
      \label{fig_hep_dataset_16}
}
\subfigure[]
{
      \includegraphics[width=0.5in]{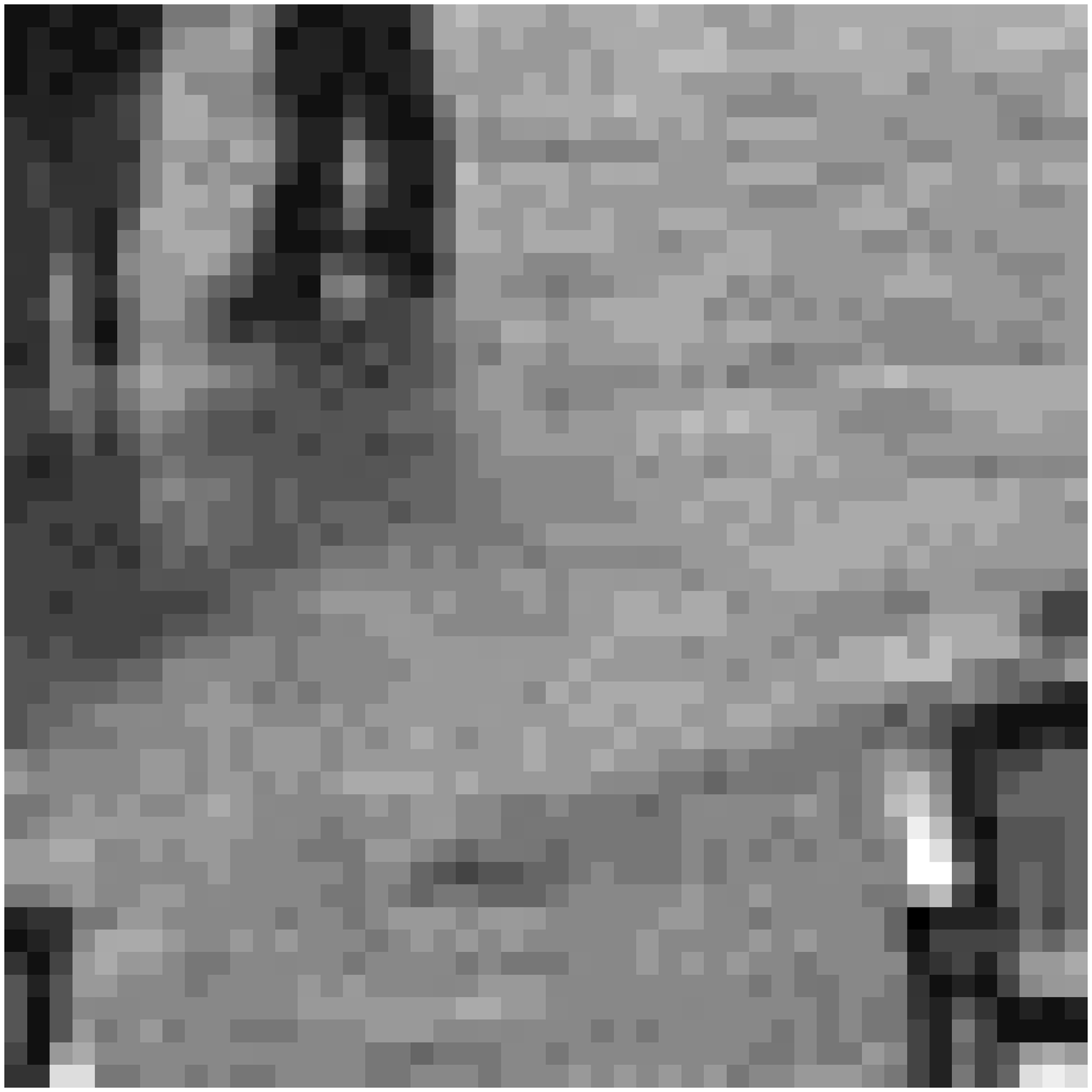}
      \label{fig_hep_dataset_17}
}
\subfigure[]
{
      \includegraphics[width=0.5in]{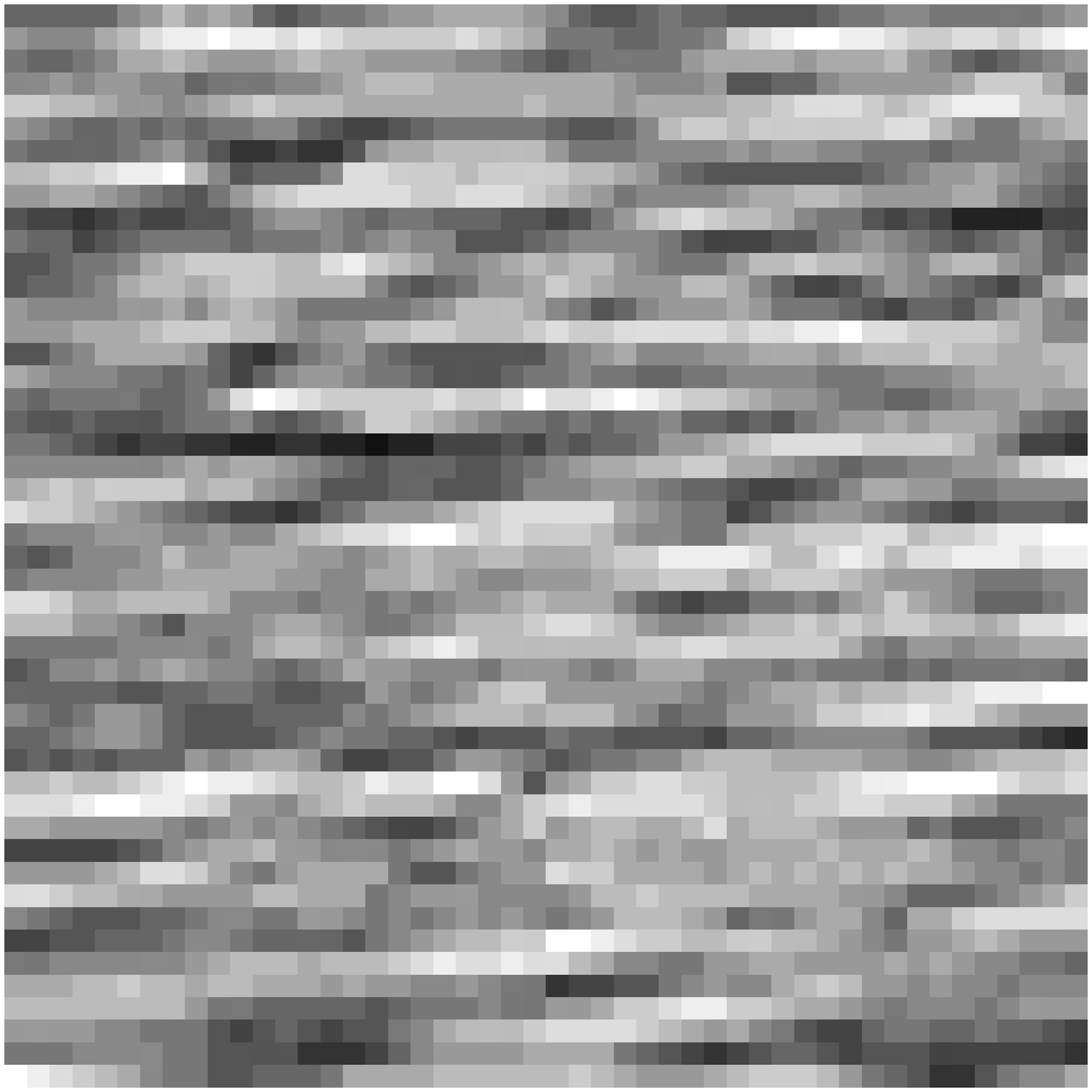}
      \label{fig_hep_dataset_18}
}
\subfigure[]
{
      \includegraphics[width=0.5in]{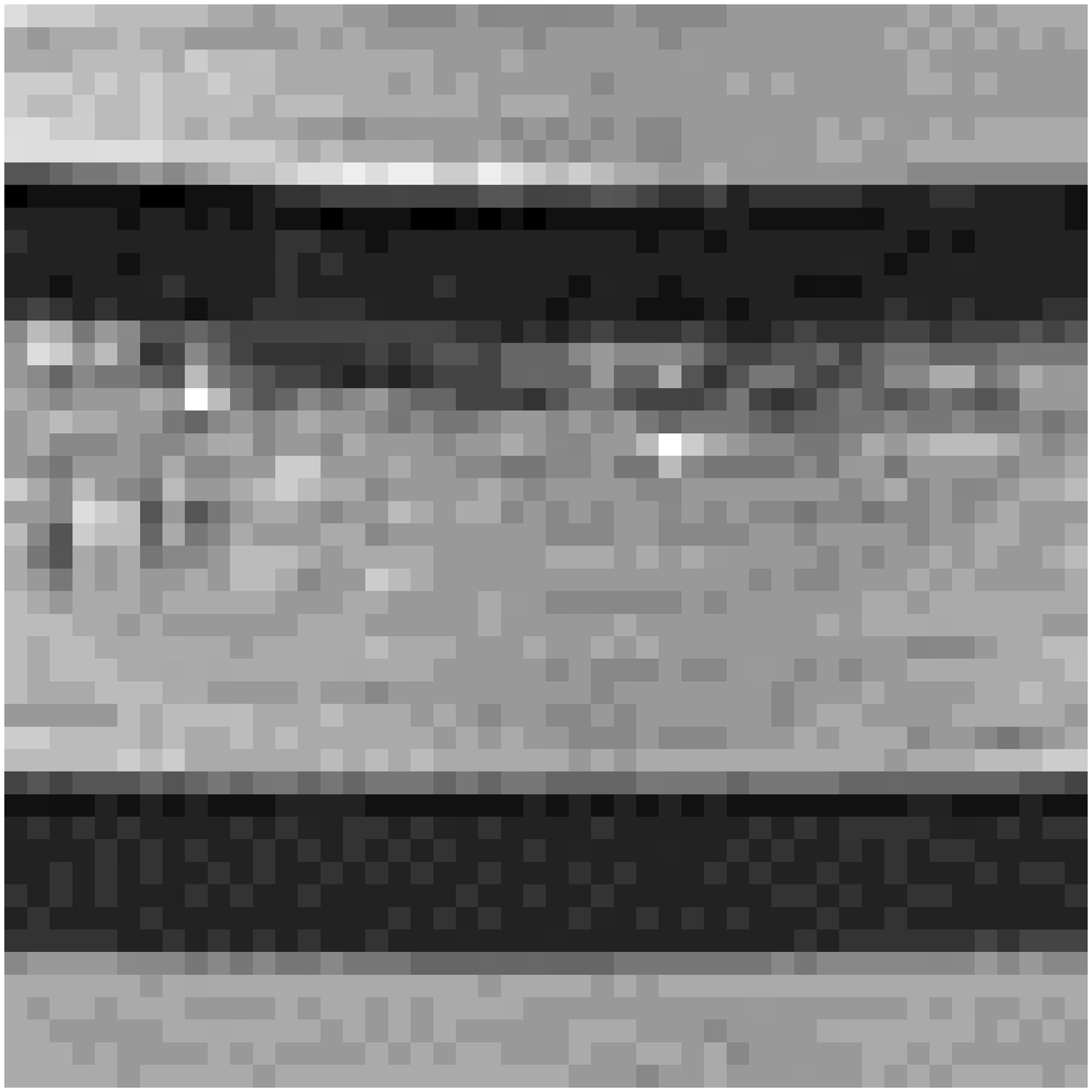}
      \label{fig_hep_dataset_19}
}
\subfigure[]
{
      \includegraphics[width=0.5in]{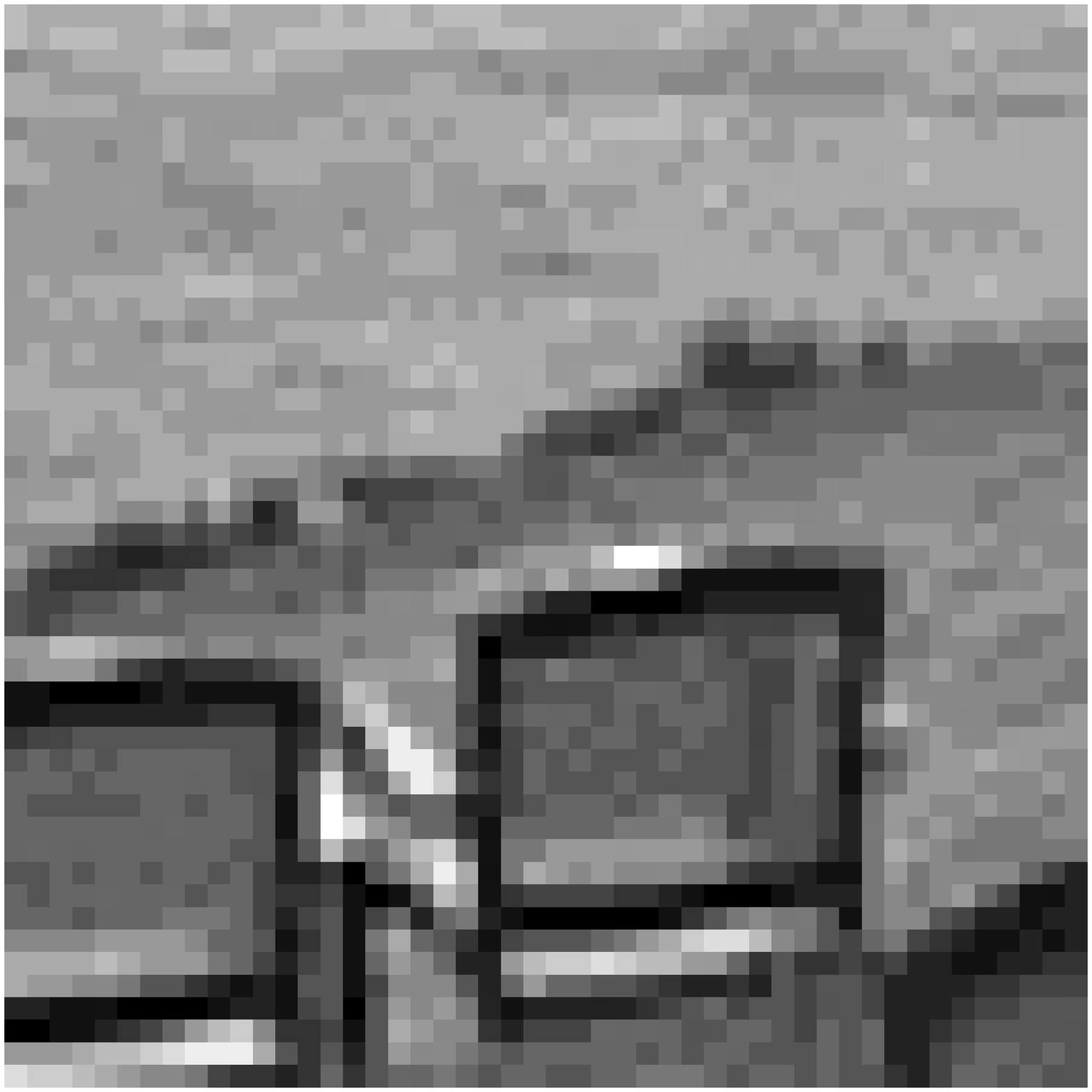}
      \label{fig_hep_dataset_20}
}
\caption{Examples of the image dataset, the first row are smoke images, others are non-smoke.}
\label{fig:hep_dataset}
\end{figure*}

As a unifying framework, a large number of  state-of-the-art texture descriptors have been implemented in HEP. For the sake of efficiency, we selected 22 descriptors with dimensions less than 1024. The details of these texture descriptors are shown in Table.\ref{tab:texture_desp_list}.
\begin{table}
\centering
\begin{center}
\caption{ Summary of texture descriptors belonging to the HEP ($G$ is the number of gray levels).}
\label{tab:texture_desp_list}
\begin{tabular}{|p{3.5cm}|p{0.8cm}|p{0.5cm}|p{0.5cm}|}
\hline \scriptsize Name &\scriptsize Acronym &\scriptsize Dims &\scriptsize Year\\
\hline \scriptsize Gray Level Differences &\scriptsize GLD &\scriptsize $G$ &\scriptsize 1976\\
\hline \scriptsize Rank Transform &\scriptsize RT &\scriptsize 9  &\scriptsize 1994\\
\hline \scriptsize Reduced Texture Unit &\scriptsize RTU &\scriptsize 45  &\scriptsize 1995\\
\hline \scriptsize Local Binary Patterns &\scriptsize LBP &\scriptsize 256  &\scriptsize 1996\\
\hline \scriptsize Simplified Texture Spectrum &\scriptsize STS &\scriptsize 81 &\scriptsize 2003\\
\hline \scriptsize Simplied texture units (+) &\scriptsize STU+ &\scriptsize 81 &\scriptsize 2003\\
\hline \scriptsize Simplied texture units (x) &\scriptsize STUx &\scriptsize 81  &\scriptsize 2003\\
\hline \scriptsize Modified Texture Spectrum &\scriptsize MTS &\scriptsize 16  &\scriptsize 2003\\
\hline \scriptsize Improved Local Binary Patterns &\scriptsize ILBP &\scriptsize 511  &\scriptsize 2004\\
\hline \scriptsize 3D Local Binary Patterns &\scriptsize 3DLBP &\scriptsize 1024  &\scriptsize 2006\\
\hline \scriptsize Center-symmetric Local Binary Pattern &\scriptsize CS-LBP &\scriptsize 16  &\scriptsize 2006\\
\hline \scriptsize Median Binary Patterns &\scriptsize MBP &\scriptsize 511  &\scriptsize 2007\\
\hline \scriptsize Local Ternary Patterns &\scriptsize LTP &\scriptsize 1024  &\scriptsize 2007\\
\hline \scriptsize Centralized Binary Patterns &\scriptsize CBP &\scriptsize 32  &\scriptsize 2008\\
\hline \scriptsize Improved Centre-Symmetric Local Binary Patterns(D) &\scriptsize D-LBP &\scriptsize 16  &\scriptsize 2009\\
\hline \scriptsize Improved Centre-Symmetric Local Binary Patterns(ID) &\scriptsize ID-LBP &\scriptsize 16  &\scriptsize 2009\\
\hline \scriptsize Improved Local Ternary Patterns &\scriptsize ILTP &\scriptsize 1024  &\scriptsize 2010\\
\hline \scriptsize Binary Gradient Contours (1) &\scriptsize BGC1 &\scriptsize 255 &\scriptsize 2011\\
\hline \scriptsize Binary Gradient Contours (2) &\scriptsize BGC2 &\scriptsize 225 &\scriptsize 2011\\
\hline \scriptsize Binary Gradient Contours (3) &\scriptsize BGC3 &\scriptsize 255 &\scriptsize 2011\\
\hline \scriptsize Center-symmetric Texture Spectrum &\scriptsize CS-TSDelta &\scriptsize 81 &\scriptsize 2011\\
\hline \scriptsize Gradient-based Local Binary Patterns &\scriptsize GLBP &\scriptsize 256 &\scriptsize 2011\\
\hline
\end{tabular}
\end{center}
\end{table}
The performances of the above descriptors are compared in regard of four aspects: recognition rate, extraction time  (the time of feature extraction for all the test images), feature dimensions and recognition time (the time of recognition for all the test images) using the YUAN's dataset (see Fig.\ref{fig:hep_dataset} for examples) which has 2592 smoke pictures and 1500 non-smoke pictures. We randomly select half of them for training and the remainders for testing.

\begin{figure}
\centering
{
  \includegraphics[width=2.5in]{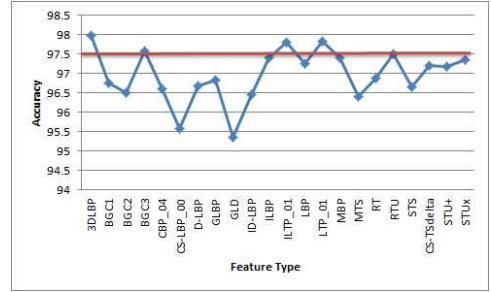}
  \label{fig_firstsub}
}
\caption{The Classification Accuracy of 22 Texture Descriptors.}
\label{fig:hep_accuracy}
\end{figure}

The SVM classifier with the five different pairs of parameters (discussed at the beginning of section \ref{sec:method}) are employed. The highest classification accuracy of each descriptor is presented in Fig.\ref{fig:hep_accuracy}.

\begin{figure}
\centering
{
  \includegraphics[width=2.5in]{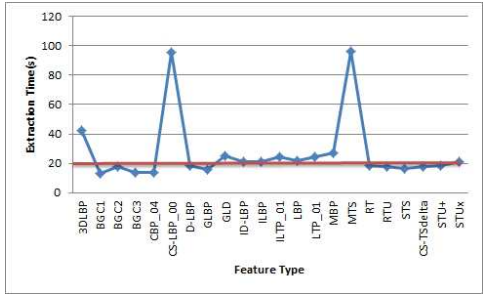}
  \label{fig_ext_time}
}
\caption{The Extraction Time of of 22 Texture Descriptors.}
\label{fig:hep_extraction_time}
\end{figure}

\begin{figure}
\centering
{
  \includegraphics[width=2.5in]{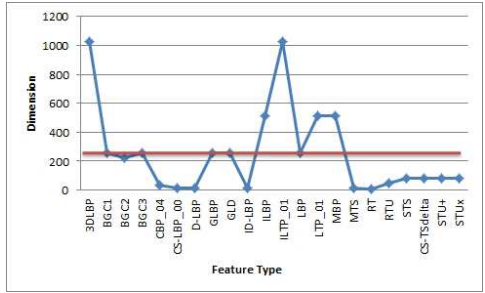}
  \label{fig_dims}
}
\caption{The Dims of 22 Texture Descriptors.}
\label{fig:hep_dims}
\end{figure}

\begin{figure*}
\centering
\subfigure[]
{
      \includegraphics[width=0.7in]{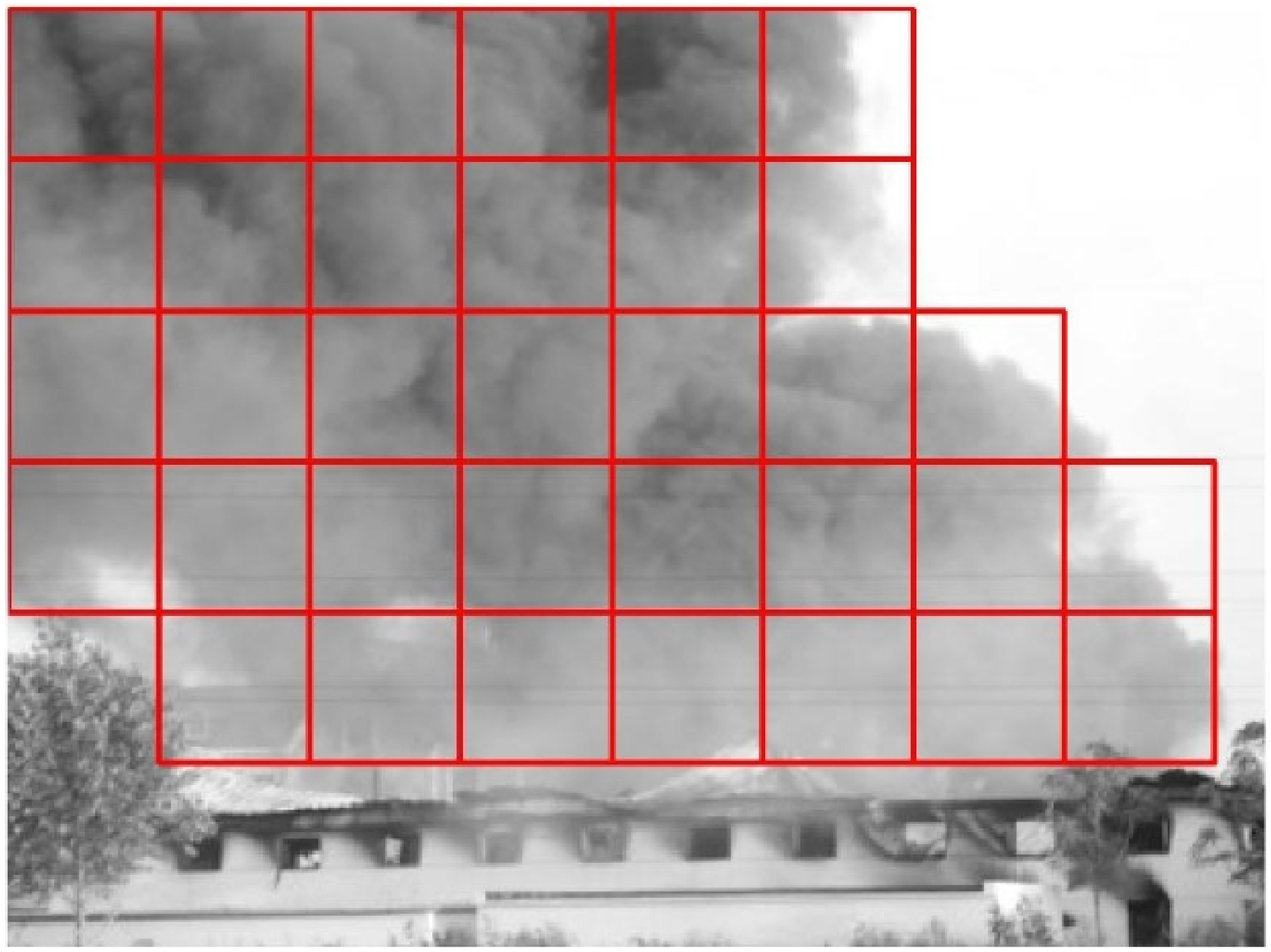}
      \label{fig_hep_dataset_1}
}
\subfigure[]
{
      \includegraphics[width=0.7in]{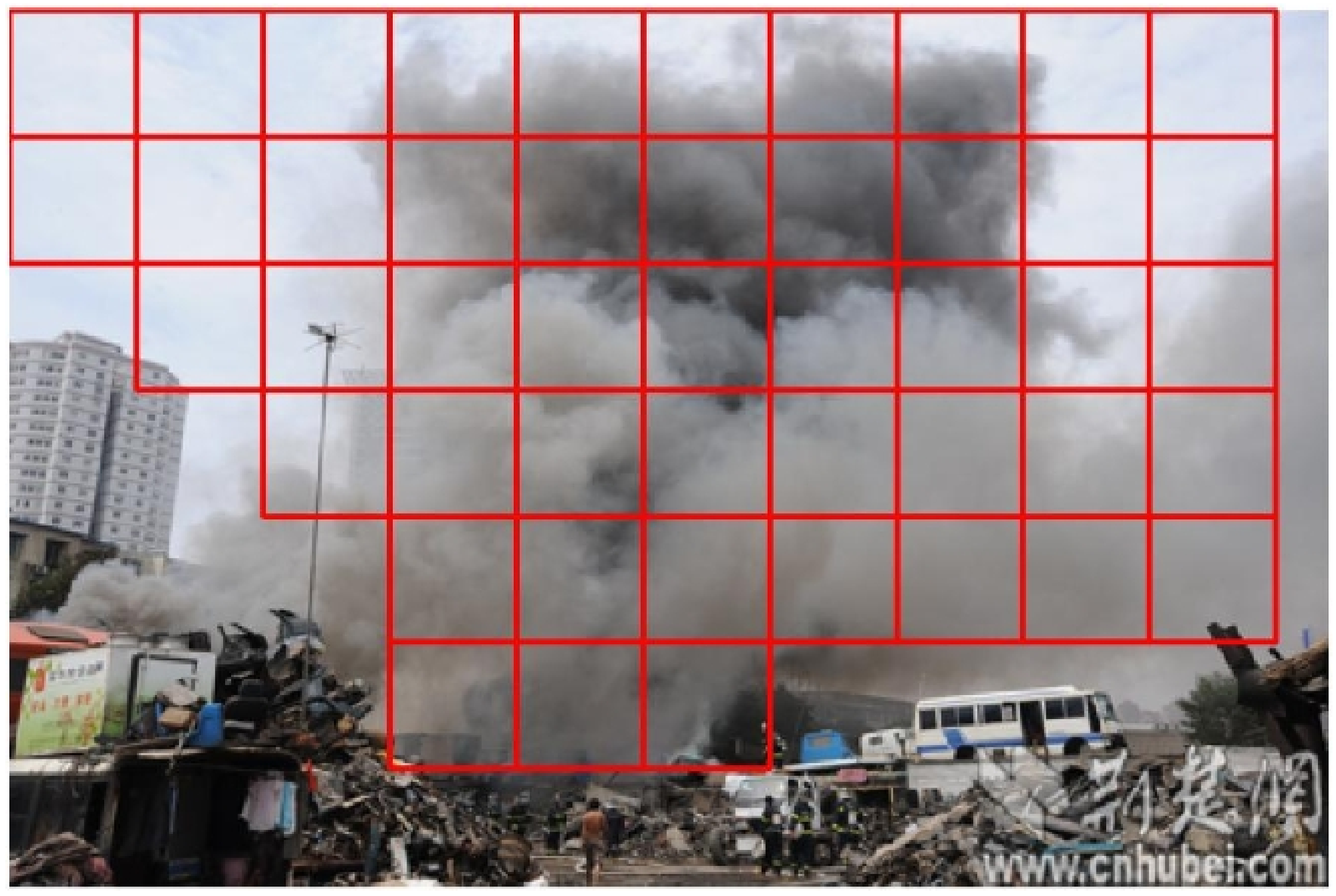}
      \label{fig_hep_dataset_2}
}
\subfigure[]
{
      \includegraphics[width=0.7in]{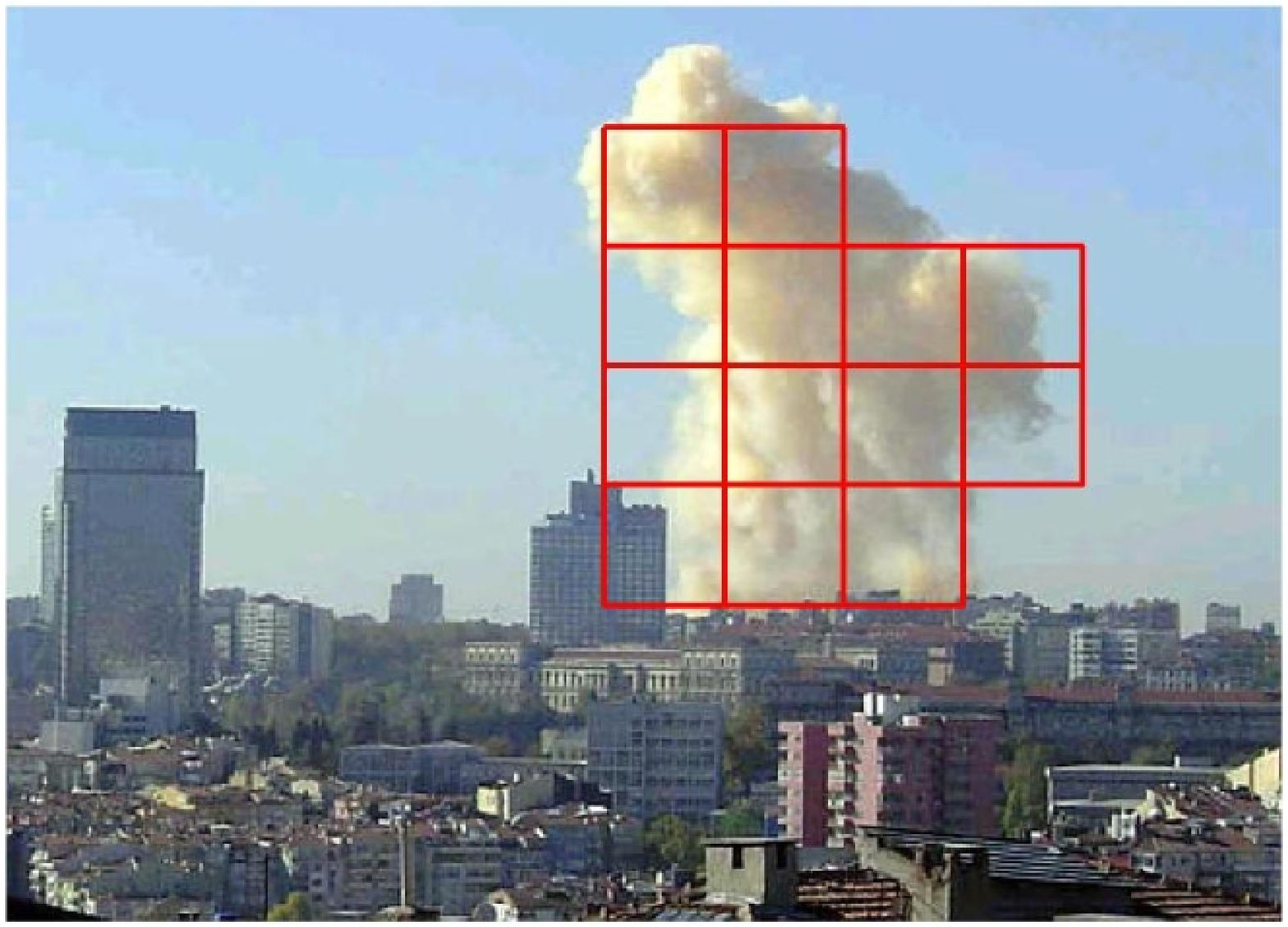}
      \label{fig_hep_dataset_3}
}
\subfigure[]
{
      \includegraphics[width=0.7in]{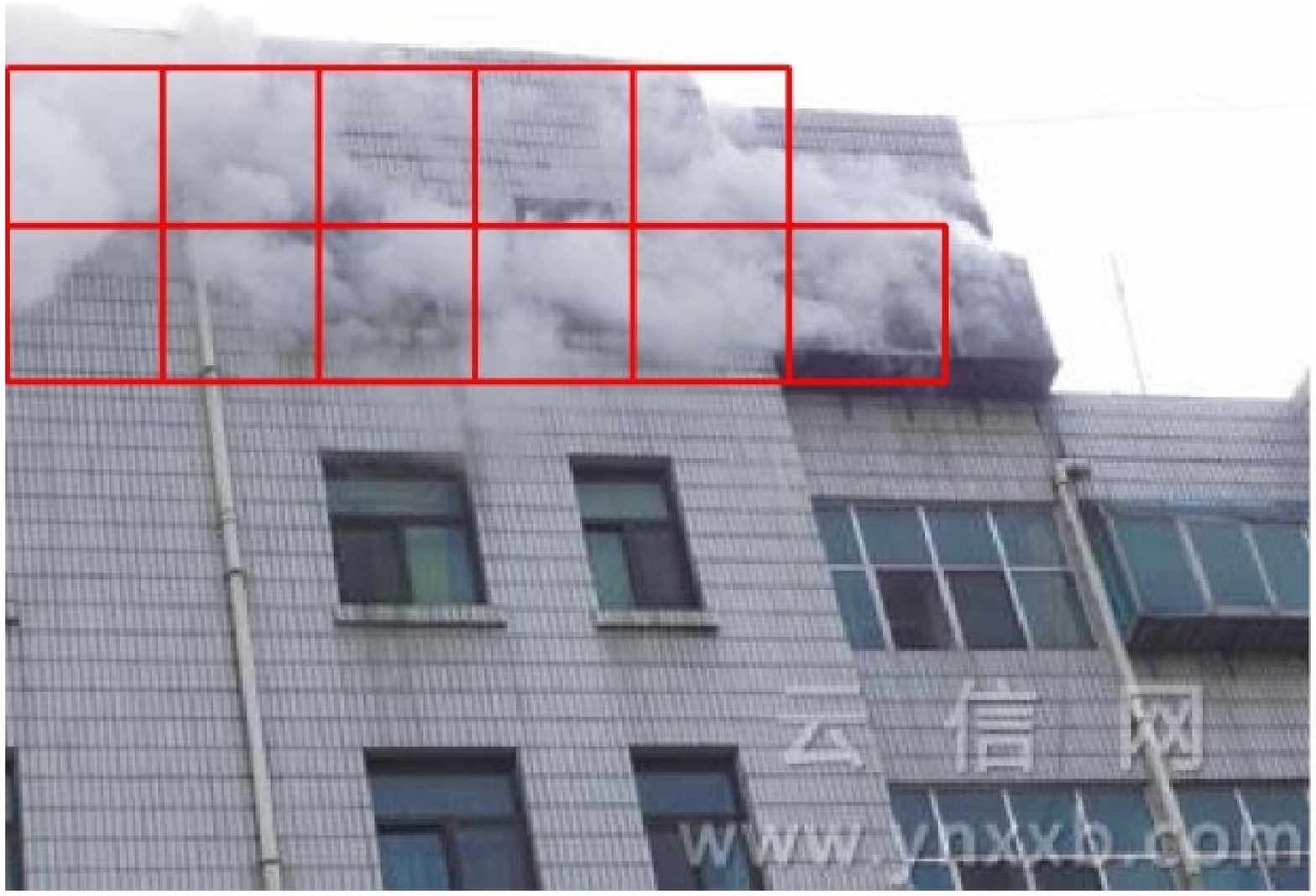}
      \label{fig_hep_dataset_4}
}
\subfigure[]
{
      \includegraphics[width=0.7in]{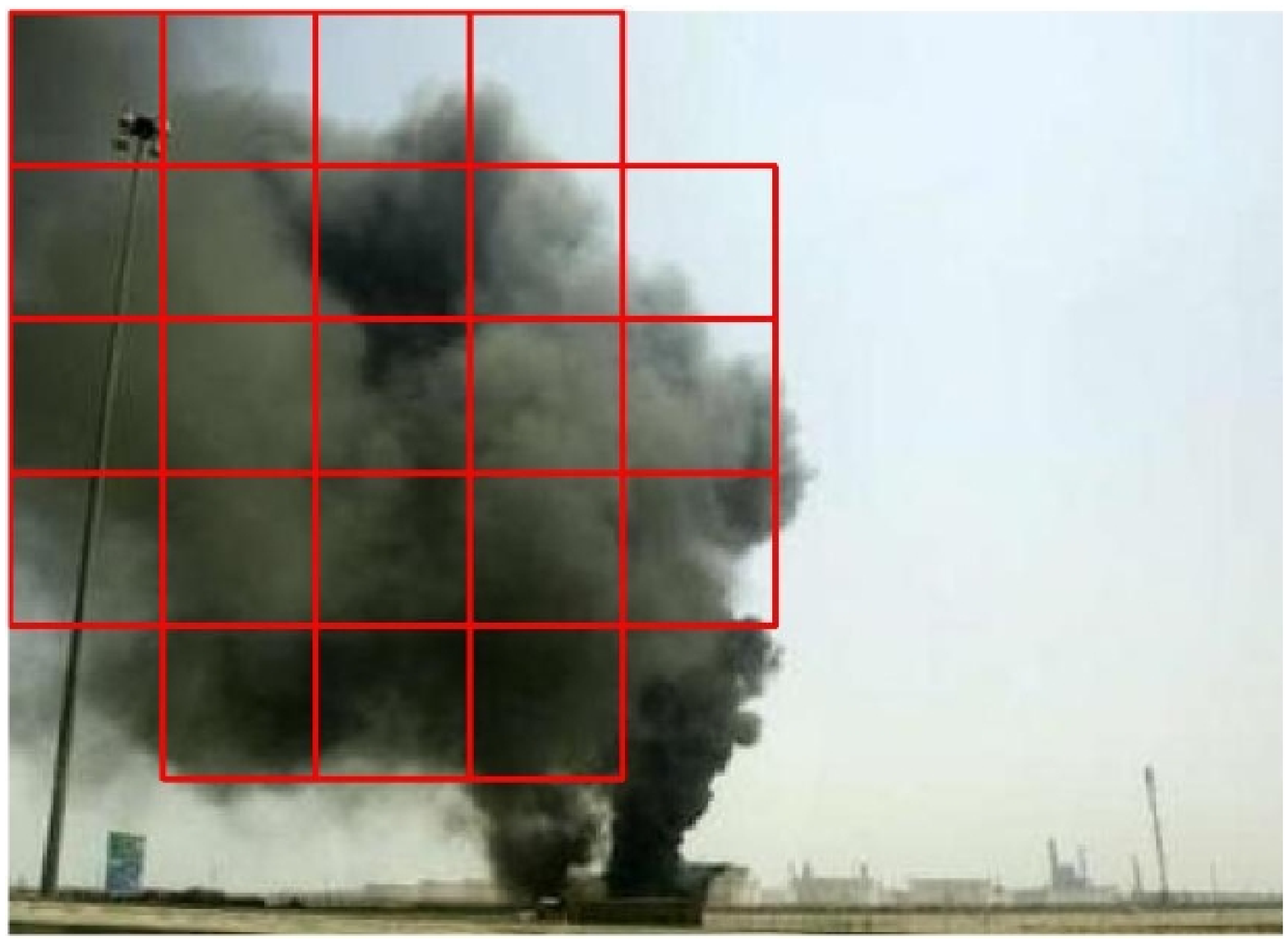}
      \label{fig_hep_dataset_5}
}
\subfigure[]
{
      \includegraphics[width=0.7in]{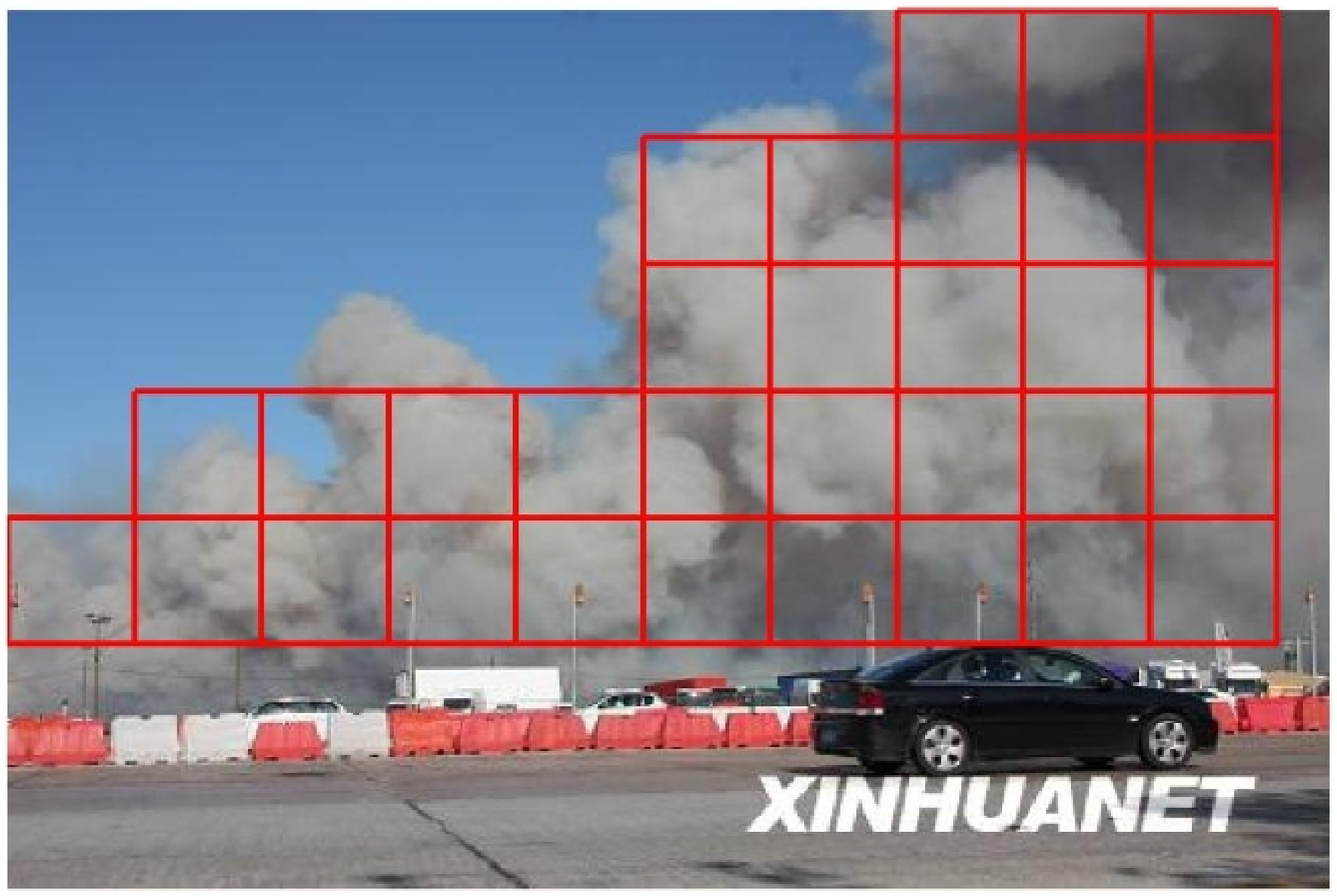}
      \label{fig_hep_dataset_6}
}
\subfigure[]
{
      \includegraphics[width=0.7in]{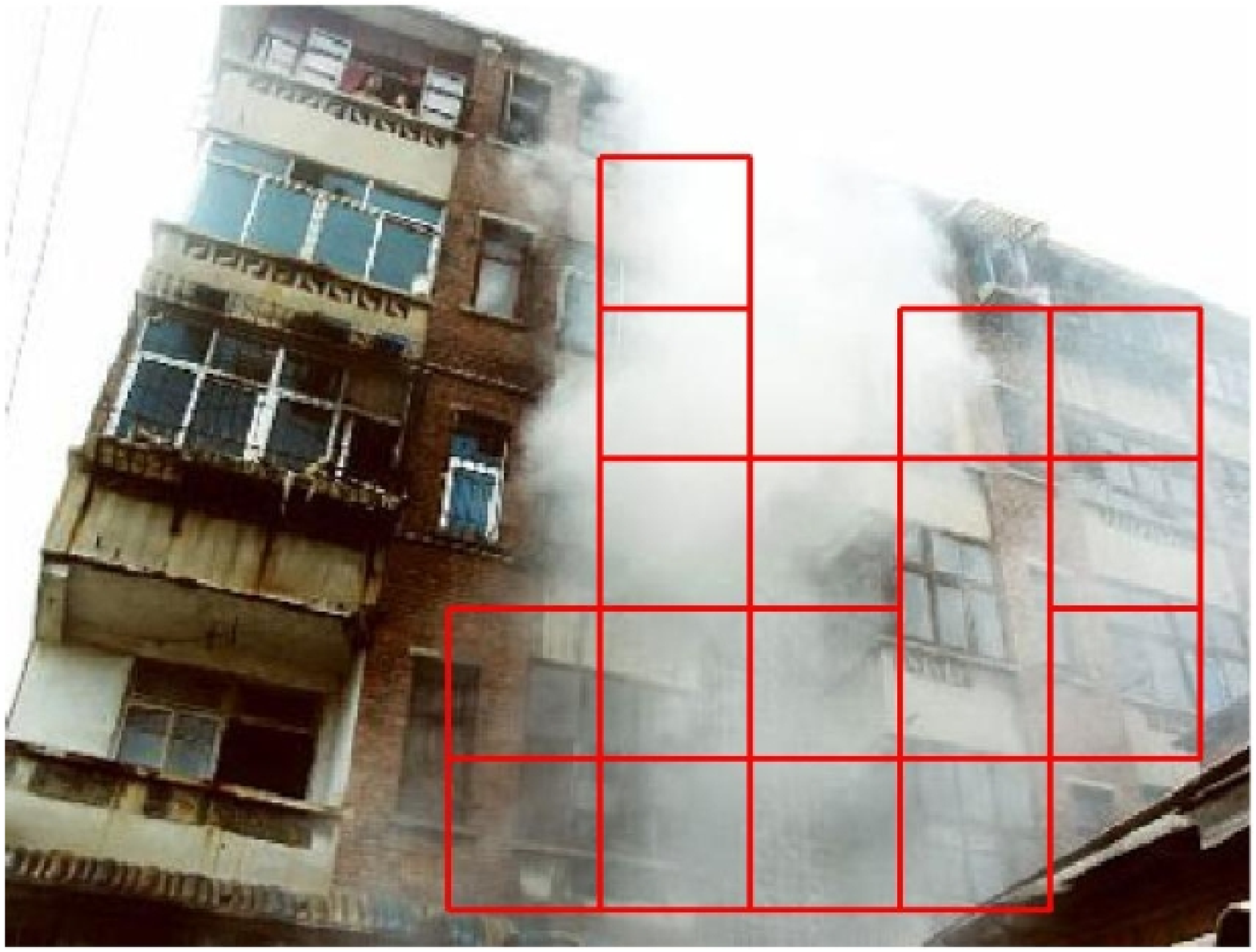}
      \label{fig_hep_dataset_7}
}
\subfigure[]
{
      \includegraphics[width=0.7in]{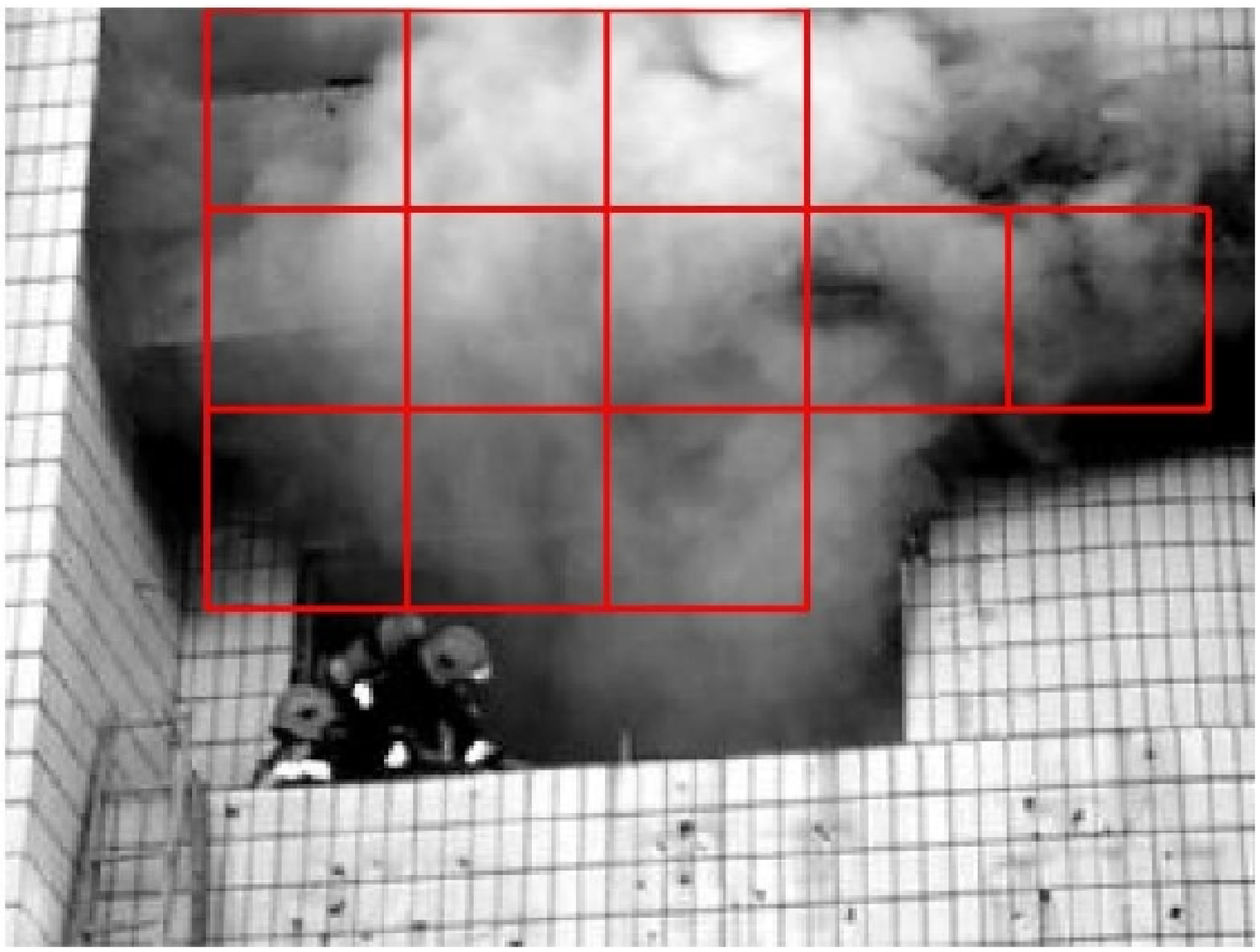}
      \label{fig_hep_dataset_8}
}
\subfigure[]
{
      \includegraphics[width=0.7in]{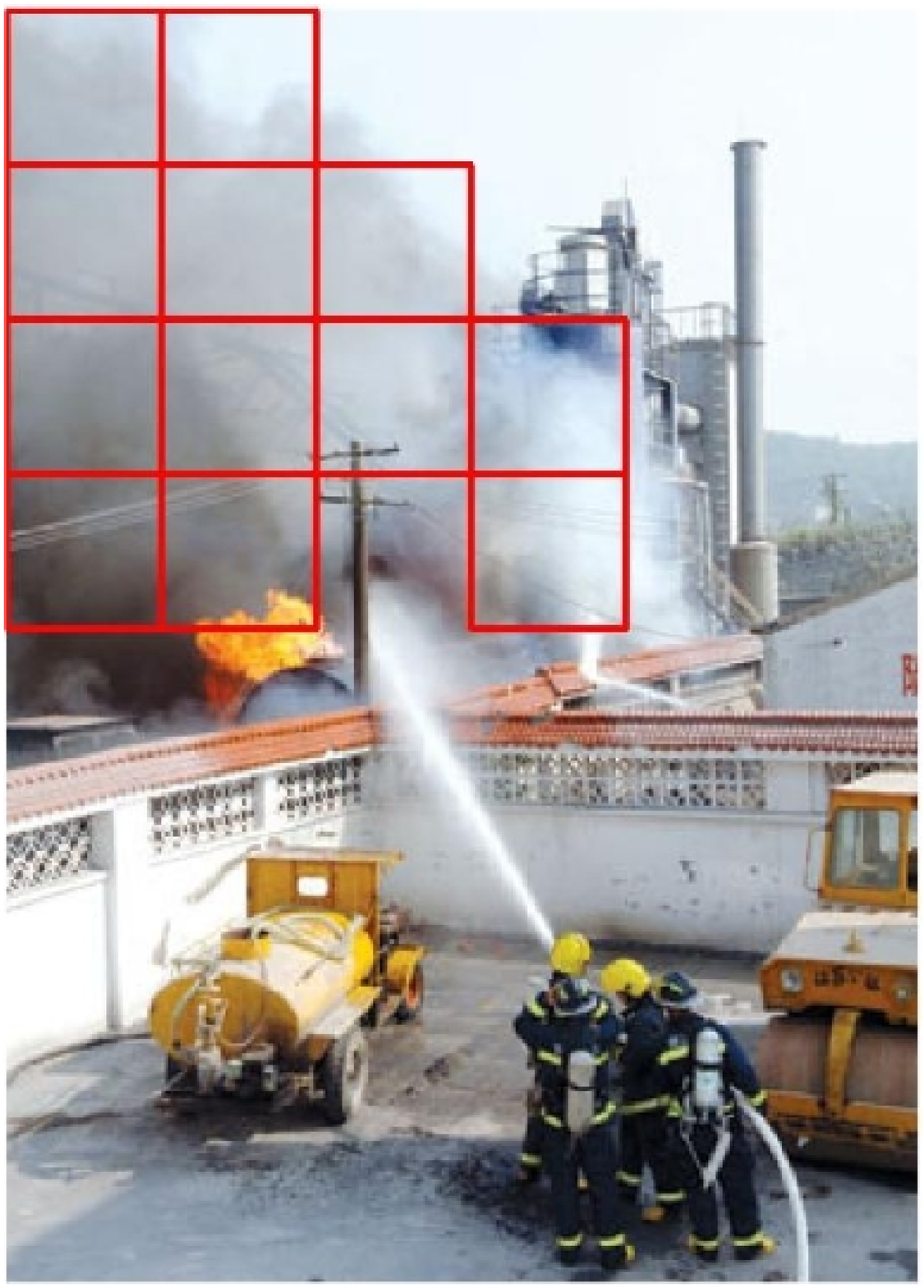}
      \label{fig_hep_dataset_9}
}
\subfigure[]
{
      \includegraphics[width=0.7in]{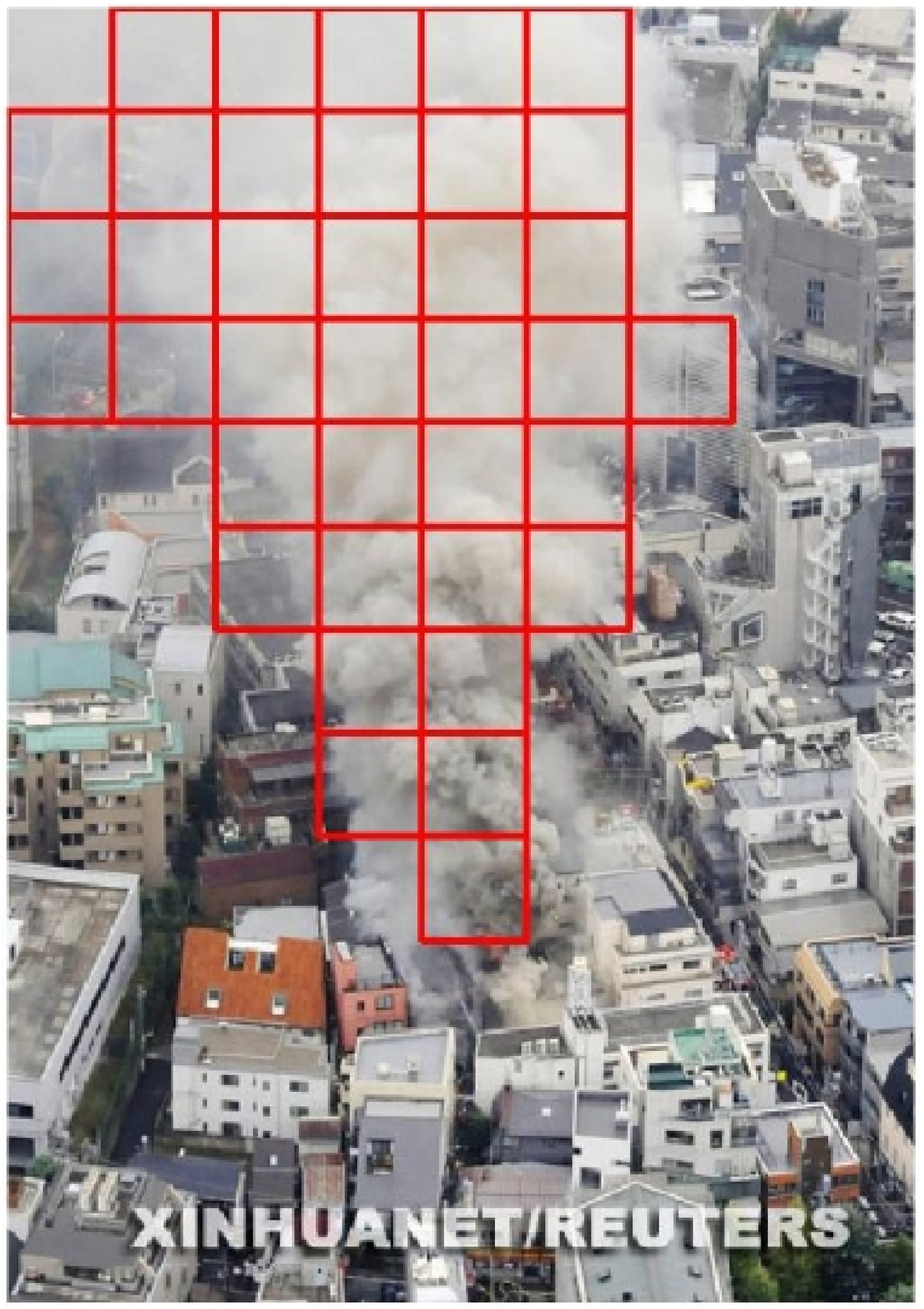}
      \label{fig_hep_dataset_10}
}
\subfigure[]
{
      \includegraphics[width=0.7in]{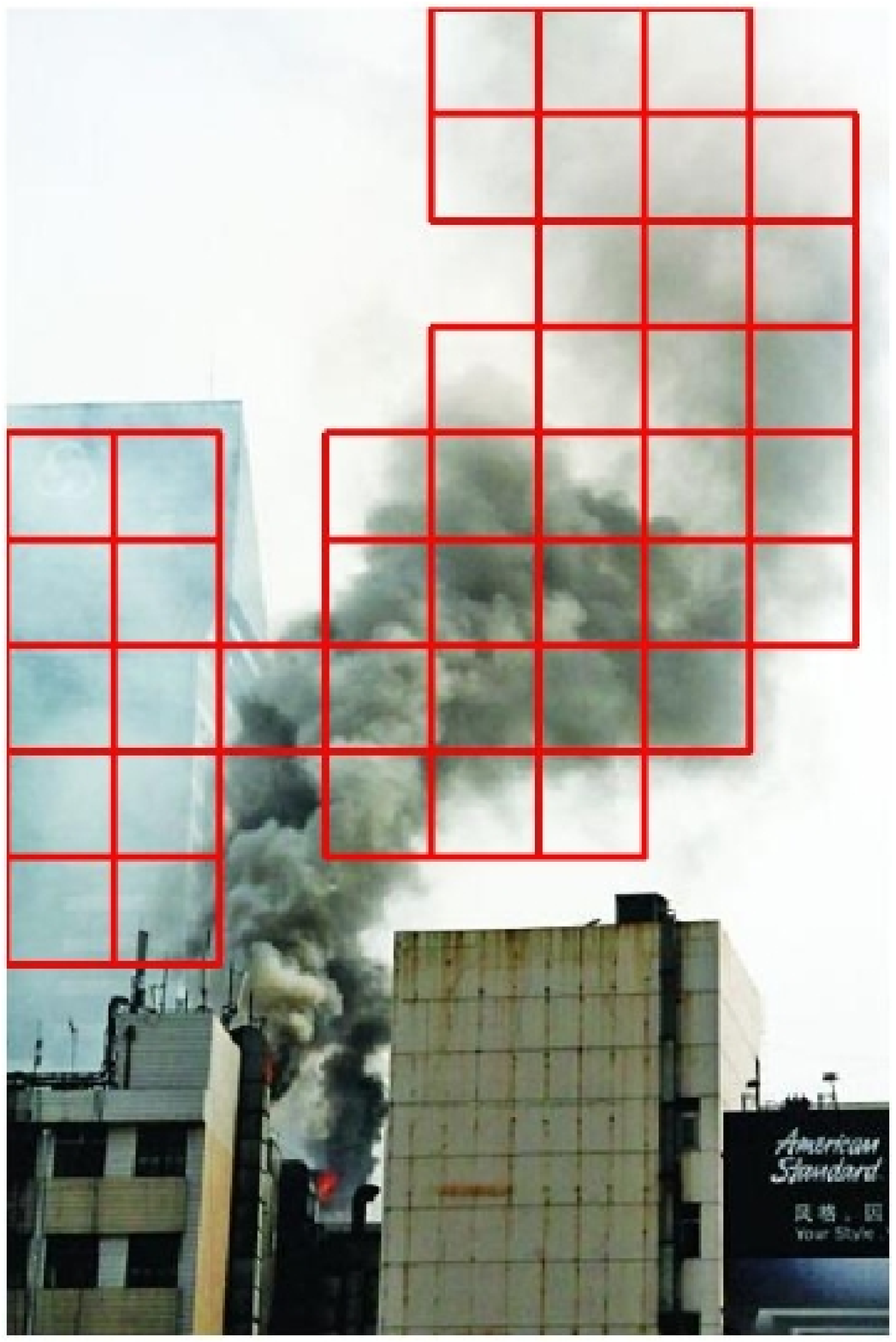}
      \label{fig_hep_dataset_11}
}
\subfigure[]
{
      \includegraphics[width=0.7in]{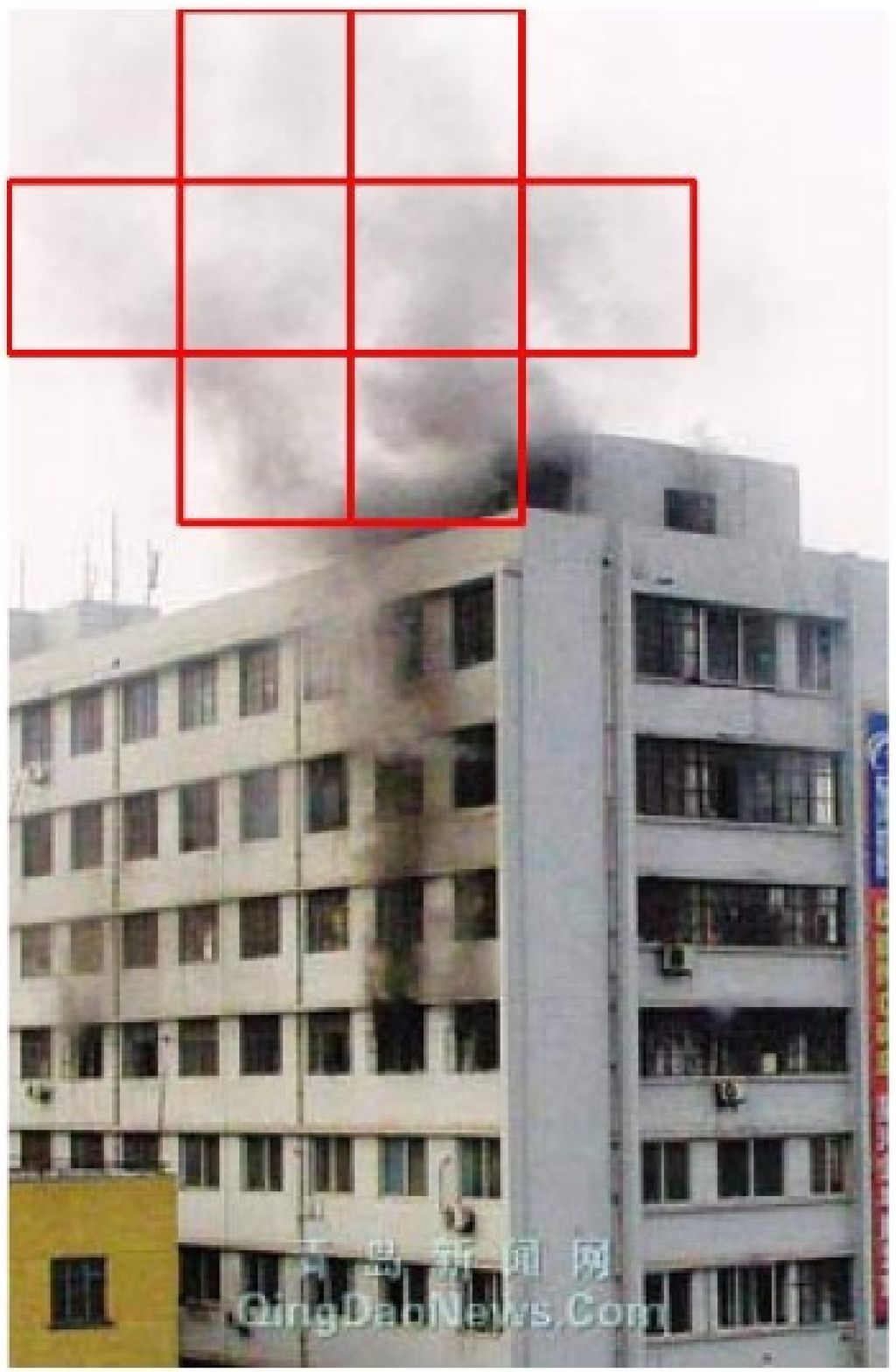}
      \label{fig_hep_dataset_12}
}
\caption{Examples of Smoke Detection Results Using BGC3.}
\label{fig:hep_result}
\end{figure*}

It can be seen that all the 22 descriptors perform well ($\ge$ 95\%) in regard of classification accuracy. The highest classification accuracy (97.98\%) is obtained by using 3DLBP. The extraction time and dims of these descriptors are illustrated in Fig.\ref{fig:hep_extraction_time} and Fig.\ref{fig:hep_dims}. Since there is no descriptor can perform the best in all the aspects, three criterion (Accuracy $\geq$ 97.5\%, Extraction time $\leq$ 20s and dims $\leq$ 256. Indicated by the red line in each corresponding figure) are utilized to find suitable descriptors. Only two descriptors, BGC3 and RTU, left. Table \ref{tab:cmp_rtu_bgc3} show details of them for comparison. Since BGC3 is faster and more accurate than RTU, it is chose for the smoke block verification in our algorithm. Some examples of smoke detection results using BGC3 are shown in Fig.\ref{fig:hep_result}.

\begin{table}
\begin{center}
\caption{ The Comparison between RTU and BGC3.}
\label{tab:cmp_rtu_bgc3}
\begin{tabular}{|c|c|c|c|c|}
\hline \scriptsize Name &\scriptsize Accuracy &\scriptsize Extraction Time &\scriptsize Dims &\scriptsize Recognition Time\\
\hline \scriptsize BGC3 &\scriptsize 97.58 &\scriptsize 13.78    &\scriptsize 255 &\scriptsize 0.32\\
\hline \scriptsize RTU &\scriptsize 97.50 &\scriptsize 17.72 &\scriptsize 45 &\scriptsize 0.02\\
\hline
\end{tabular}
\end{center}
\end{table}

\subsection{Space-time Feature Analysis}
To avoid false alarm, spatial appearance and temporal dynamics is combined for further verification of the suspected smoke blocks.
\subsubsection{Block Based Inter-Frame Difference} \label{sec:bbifd}
\begin{figure*} \centering
{
  \includegraphics[width=9cm]{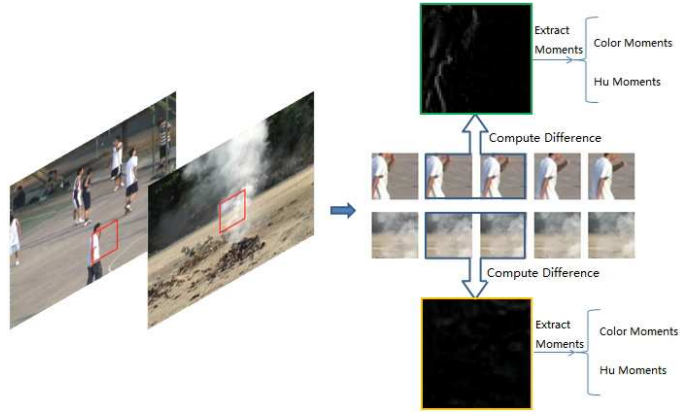}
  \label{fig_firstsub}
}
\caption{The process of BIFD feature extraction.}
\label{fig:BIFD_Extraction}
\end{figure*}

In \cite{chen2013dynamic}, Chen et al. proposed space-time feature to analyze the inter-frame variations of the candidate smoke blocks. A brief description is illustrated in Fig. \ref{fig:BIFD_Extraction}. Let $\mathds{B}_{p,k}$ be the $k$-th block estimated as smoke candidate block in $p$-th frame $\mathds{F}_{p}$. The differences of the corresponding blocks between adjacent frames from $\mathds{F} _{p-q+1}$ to $\mathds{F}_{p}$ are calculated as

\begin{equation}\label{eq:BDA}
\begin{split}
    \textit{BDA}_{p,q,k} = \{\textit{DIF}_{p-q+1,p-q+2,k}, \\\textit{DIF}_{p-q+2,p-q+3,k},...,\textit{DIF}_{p-1,p,k}\}
\end{split}
\end{equation}

\begin{equation}\label{eq:DIF}
    \textit{DIF}_{i,j,k} = |\mathds{B}_{i,k}-\mathds{B}_{j,k}|
\end{equation}
where $\textit{DIF}_{i,j,k}$ is the difference between the corresponding blocks of the $\mathds{F}_{i}$ and $\mathds{F}_{j}$.

\begin{figure} \centering
{
  \includegraphics[width=6.5cm]{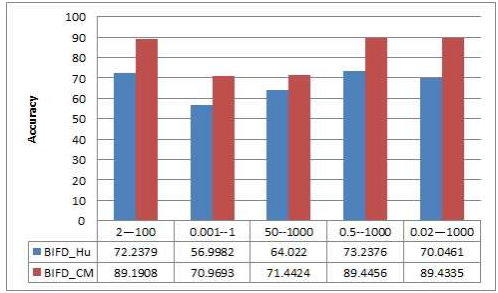}
  \label{fig_firstsub}
}
\caption{The accuracy comparison between $\textit{BIFD}_{Hu}$ and $\textit{BIFD}_{CM}$.}
\label{fig:BIFD_Hu_CM}
\end{figure}

Hu moment \cite{hu1962visual} and Color moment \cite{stricker1995similarity} are implemented to construct dynamic feature of the candidate smoke blocks as described in equation (\ref{eq:BIFD_Hu}) and (\ref{eq:BIFD_CM}) respectively. In which, $h_{i,k}$ is $i$-th Hu moment of the $k$-th $\textit{DIF}$ of the $\textit{BDA}$; $\mu_{k}$, $\sigma_{k}$ and  $s_{k}$ are the mean, variance and skewness of $k$-th $\textit{DIF}$ of the $\textit{BDA}$ respectively.

\begin{equation}\label{eq:BIFD_Hu}
\begin{split}
\textit{BIFD}_{Hu} = \{h_{1,1}, h_{2,1}, h_{3,1}, h_{4,1}, h_{5,1}, h_{6,1}, h_{7,1}, \cdots, \\
h_{1,q-1}, h_{2,q-1}, h_{3,q-1}, h_{4,q-1}, h_{5,q-1}, h_{6,q-1}, h_{7,q-1}\}
\end{split}
\end{equation}
\begin{equation}\label{eq:BIFD_CM}
\begin{split}
\textit{BIFD}_{CM} = \{\mu_1, \sigma_1, s_1, \mu_2, \sigma_2, s_2, \\
\cdots, \mu_{q-1}, \sigma_{q-1}, s_{q-1}\}
\end{split}
\end{equation}
\begin{table}
\centering
\caption{Videos of training data}
\label{tab:Traindatalist}
\begin{tabular}{|c|c|c|c|c|}
\hline \scriptsize Video &\scriptsize Total videos &\scriptsize Total frames used &\scriptsize Total number of \textit{BDA}s \\
\hline \scriptsize Smoke &\scriptsize 9 &\scriptsize 320 &\scriptsize 4697\\
\hline \scriptsize Non-smoke &\scriptsize 8 &\scriptsize 480 &\scriptsize 11788\\
\hline
\end{tabular}
\end{table}
The performance of smoke block classification of $\textit{BIFD}_{Hu}$ and $\textit{BIFD}_{CM}$ are evaluated on a dataset (see Table \ref{tab:Traindatalist}) by SVM with different values of $gamma$ and $C$. The corresponding results are presented in Fig. \ref{fig:BIFD_Hu_CM}. As $\textit{BIFD}_{CM}$ achieves higher classification accuracy in all the configurations, it is selected as the feature to describe the inter-frame variation.

\subsubsection{Descriptors on Three Orthogonal Planes}
\begin{figure*}
\centering
{
      \includegraphics[width=10cm]{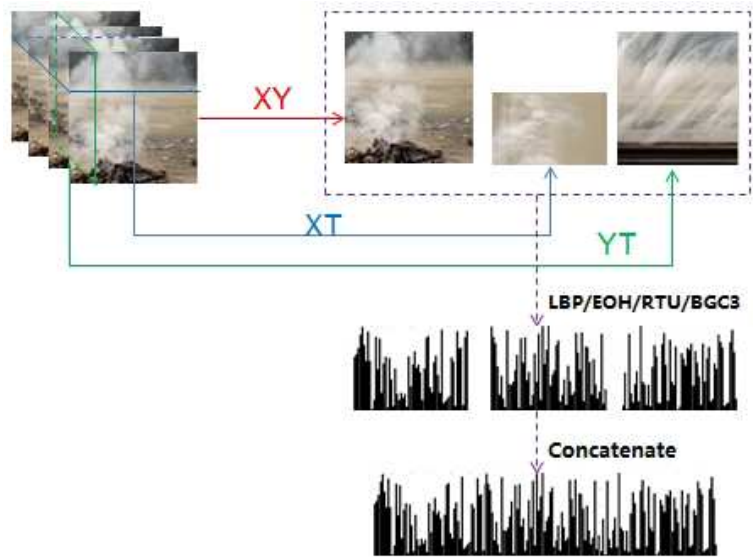}
      \label{fig_firstsub}
}
\caption{The construction of LBP-TOP, EOH-TOP, RTU-TOP and BGC3-TOP.}
\label{fig:LBP-TOP_Extraction}
\end{figure*}

The LBP-TOP \cite{Zhao2007lbp_top} was proposed to calculate the Local Binary Pattern (LBP) from Three Orthogonal Planes (TOP), denoted as XY, XT and YT. The operator is defined as
\begin{equation}\label{eq:lbp-top}
\mathbf{LBP\text{--}TOP}_{P_{XY},P_{XT},P_{YT},R_X,R_Y,R_T}
\end{equation}
where the notation ($P_{XY},P_{XT},P_{YT},R_X,R_Y,R_T$) denotes a neighborhood of $P$ points equally sampled on a circle of radius $R$ on XY, XT and YT planes respectively. Fig. \ref{fig:LBP-TOP_Extraction} illustrates the procedure of the LBP-TOP descriptor extraction. In such a scheme, LBP encodes appearance and motion in three directions, incorporating spatial information in XY and spatial temporal co-occurrence statistics in XT and YT (see equation \ref{eq:LBPXY}, \ref{eq:LBPXT} and \ref{eq:LBPYT}). The three statistics are concatenated into a single histogram as equation (\ref{eq:LBP-TOP}).
\begin{equation}\label{eq:LBPXY}
\textit{LBP}_{XY}=\textit{LBP}_{uniform}(XY)
\end{equation}
\begin{equation}\label{eq:LBPXT}
\textit{LBP}_{XT}=\textit{LBP}_{uniform}(XT)
\end{equation}
\begin{equation}\label{eq:LBPYT}
\textit{LBP}_{YT} = \textit{LBP}_{uniform}(YT)
\end{equation}
\begin{equation}\label{eq:LBP-TOP}
\mathrm{LBP\text{--}TOP}=\{\textit{LBP}_{XY}, \textit{LBP}_{XT}, \textit{LBP}_{YT}\}
\end{equation}

\begin{figure}
\centering
{
      \includegraphics[width=6.5cm]{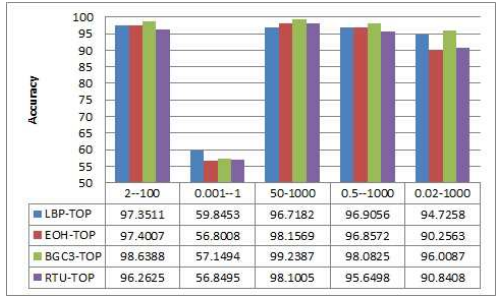}
      \label{fig_firstsub}
}
\caption{The detection accuracy of different TOP descriptors.}
\label{fig:LBP-TOP_Accuracy}
\end{figure}

From the inspiration of LBP-TOP, we construct new TOP descriptors by substituting the LBP of equation (\ref{eq:LBP-TOP}) for another descriptor $\mathfrak{X}$. According to the evaluation in section \ref{sec:hep}, the EOH, RTU and BGC3 are employed. Fig. \ref{fig:LBP-TOP_Extraction} illustrates the process of feature extraction. Fig. \ref{fig:LBP-TOP_Accuracy} shows the corresponding classification results using SVM with different values of $gamma$ and $C$ on the dataset introduced in section \ref{sec:bbifd}. A more detailed comparison can be found in Table \ref{tab:tops_comparison}. In most cases, the propose new TOP descriptors obtain higher degree of classification accuracy. Since EOH-TOP has less dims and computation cost, it is utilized in our smoke detection system.

\begin{table}
\centering
\caption{The comparison between LBP-TOPs on dims, extraction time, recognition time and sum time}
\label{tab:tops_comparison}
\begin{tabular}{|c|c|c|c|c|}
\hline \scriptsize Feature Type &\scriptsize LBP-TOP &\scriptsize EOH-TOP &\scriptsize BGC3-TOP &\scriptsize RTU-TOP\\
\hline \scriptsize Dims &\scriptsize 177 &\scriptsize 48 &\scriptsize 765 &\scriptsize 135\\
\hline \scriptsize Extraction Time &\scriptsize 0.018 &\scriptsize 0.005 &\scriptsize 0.007 &\scriptsize 0.007\\
\hline \scriptsize Recognition Time &\scriptsize 0.0006 &\scriptsize 0.0007 &\scriptsize 0.012 &\scriptsize 0.001\\
\hline \scriptsize Total Time &\scriptsize 0.0186 &\scriptsize 0.0057 &\scriptsize 0.019 &\scriptsize 0.008\\
\hline
\end{tabular}
\end{table}

\subsection{Block-based Smoke History Image}
At the final step of our smoke detection method, we proposed an algorithm which computes a confidence value using the current and historical classification results of candidate smoke blocks to reduce the false alarm. Inspired by Motion History Image (MHI) \cite{19,20}, we construct a static image template in which the pixel intensity is a function of the recency classification results, namely Smoke History Image (SHI). The brighter values indicate the corresponding blocks are more recently been classified as smoke candidates. SHI have a max value $T$, if the $i$-th block is detected as smoke $Det(i)=1$, then the SHI value of block is set to $T$, otherwise the SHI value of block minus one.
\begin{equation}\label{eq:SHIU}
\begin{split}
     SHI(i) = &\left\{\begin{aligned}
     T & & Det(i)=1 \\
     SHI(i)-1 & & Otherwise \\
                    \end{aligned}\right.
\end{split}
\end{equation}

If $Det(i)=1$ and the $SHI(i)\geq {TH}$, the $i$-th block is finally detected as smoke, otherwise non-smoke.
\begin{equation}\label{eq:SHID}
\begin{split}
     Final(i) = &\left\{\begin{aligned}
     1 & & Det(i)=1 \&\& SHI(i)\geq {TH} \\
     0 & & Otherwise \\
                    \end{aligned}\right.
\end{split}
\end{equation}

\begin{figure}[h]
\centering
\subfigure[Detected false positive]
{
      \includegraphics[width=1.3in]{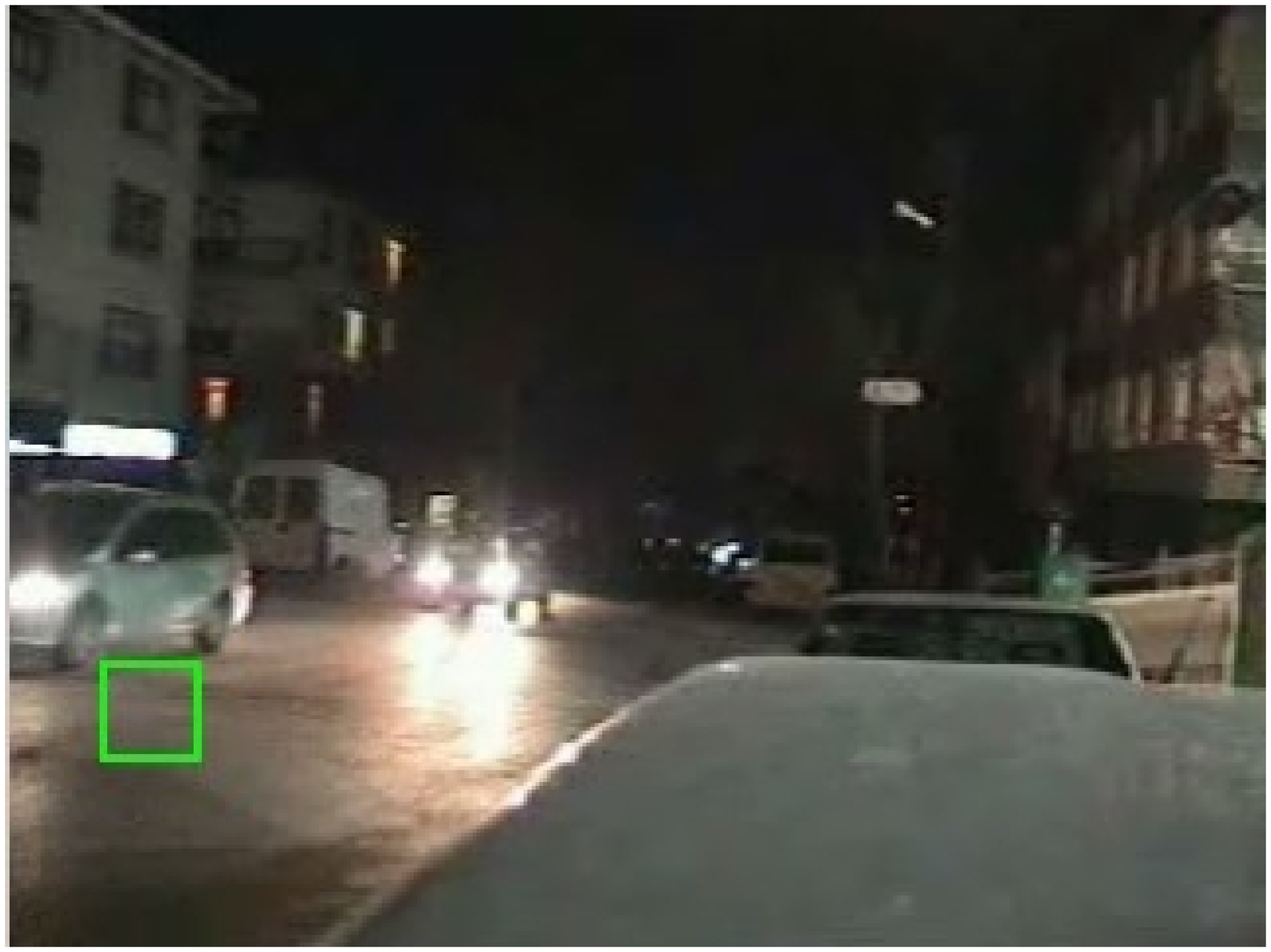}
      \label{fig:SHI_a}
}
\subfigure[Current SHI]
{
      \includegraphics[width=1.3in]{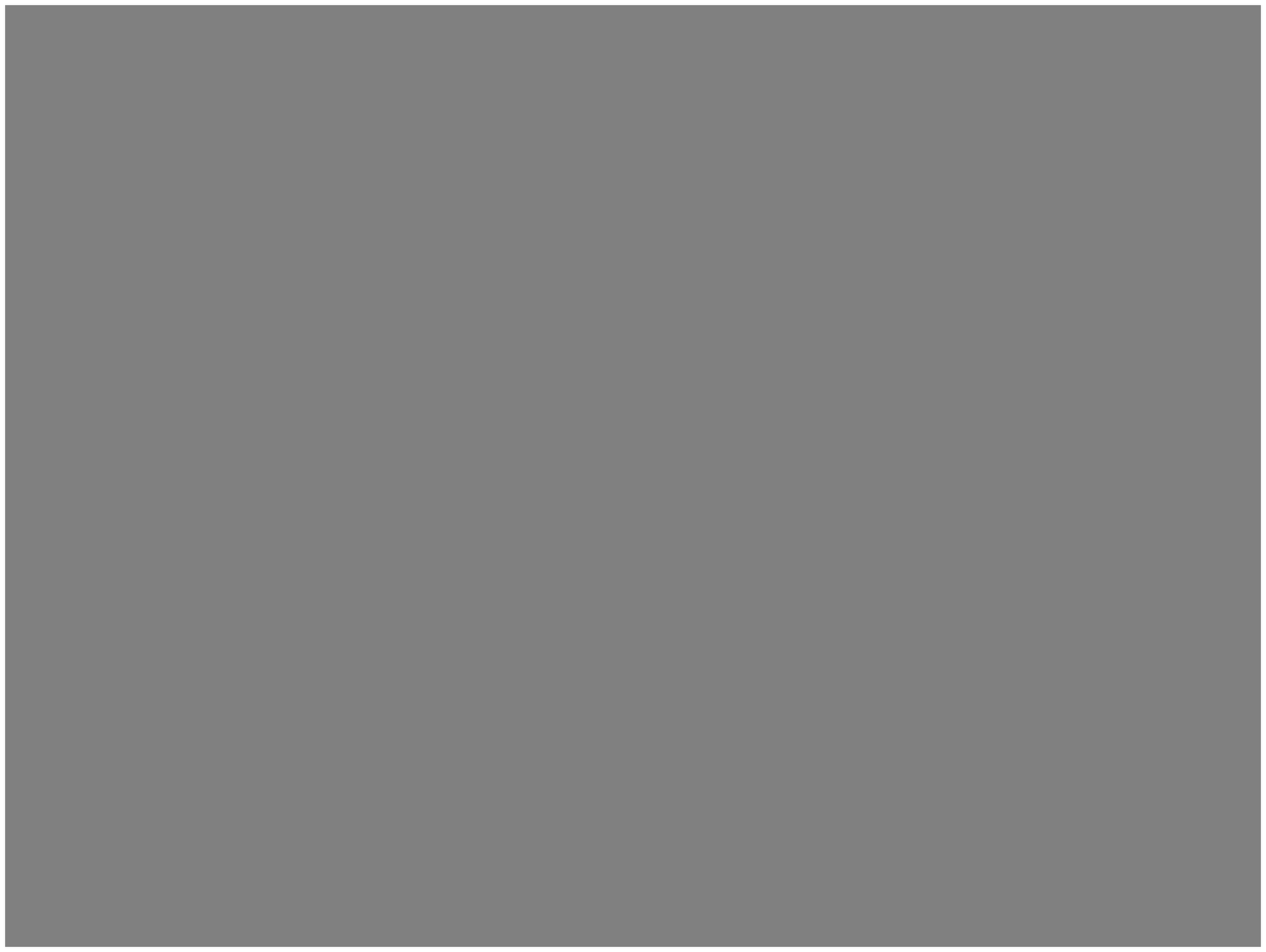}
      \label{fig:SHI_b}
}
\subfigure[Final result]
{
      \includegraphics[width=1.3in]{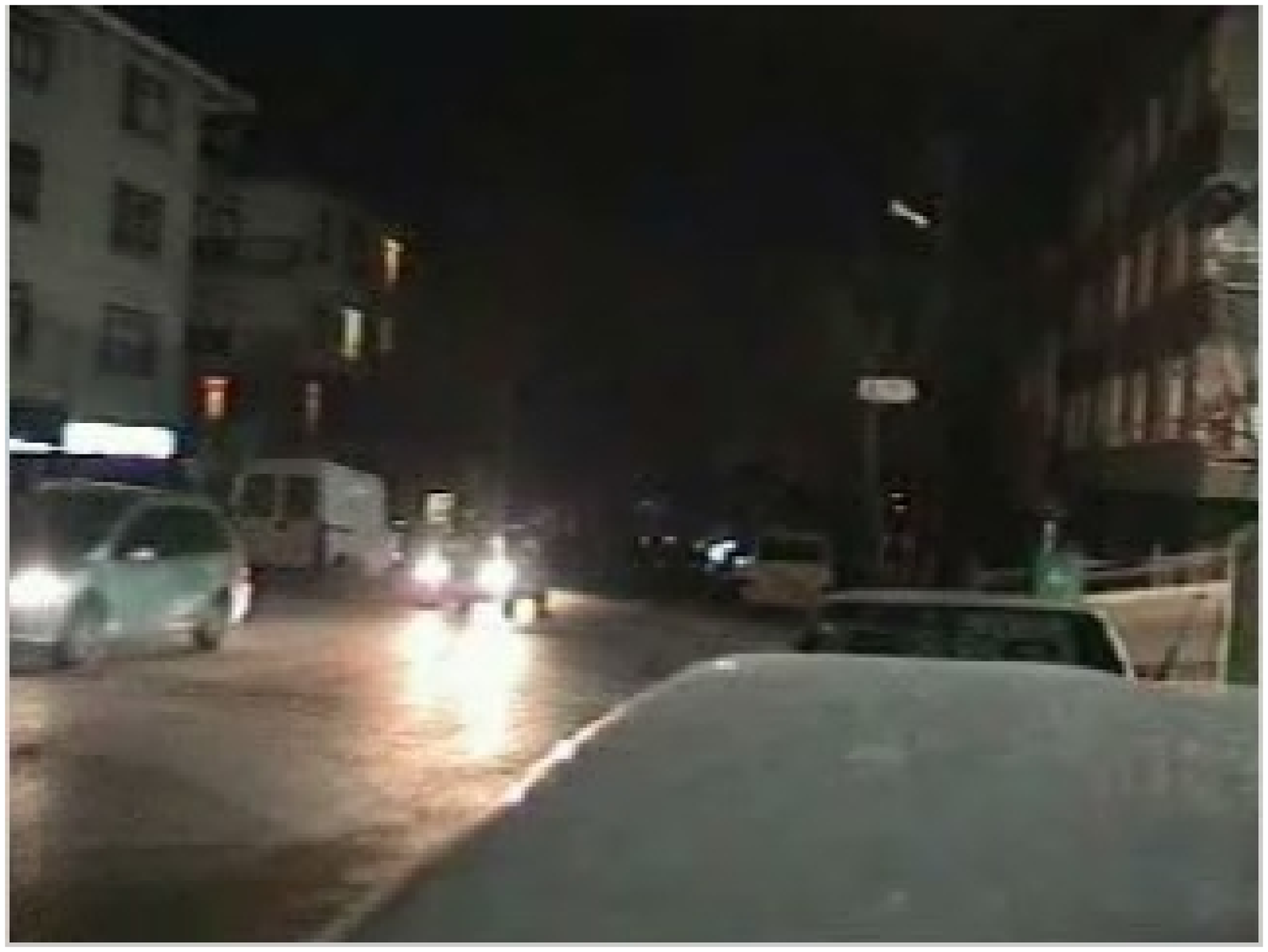}
      \label{fig:SHI_c}
}
\subfigure[Updated SHI]
{
      \includegraphics[width=1.3in]{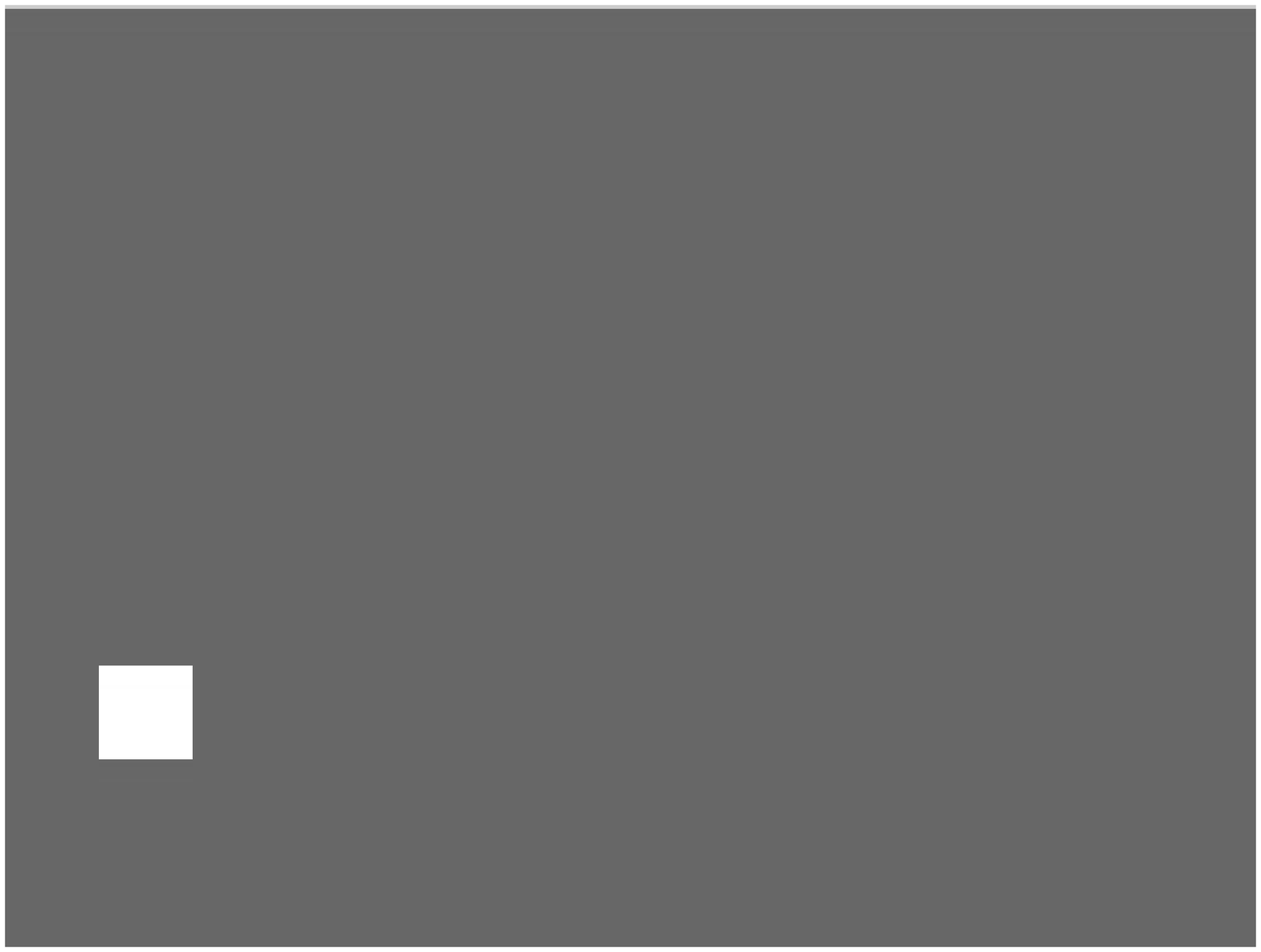}
      \label{fig:SHI_d}
}
\caption{An example of removing false positive by using SHI. }
\label{fig:SHI}
\end{figure}

Fig. \ref{fig:SHI} (a) and (b) show a detected smoke candidate (indicated by the green rectangle) and the corresponding SHI image. Finally, the smoke candidate is classified as non-smoke since the corresponding value of SHI is less than the predefined threshold. After the final decision of each frame, the SHI is updated (see Fig. \ref{fig:SHI} (d)). It can be see that the false positive is removed by using SHI.

\section{Result} \label{sec:result}
\subsection{Data Set}
\begin{figure*}\label{fig:exam1}
\centering
\subfigure[]
{
      \includegraphics[width=0.7in]{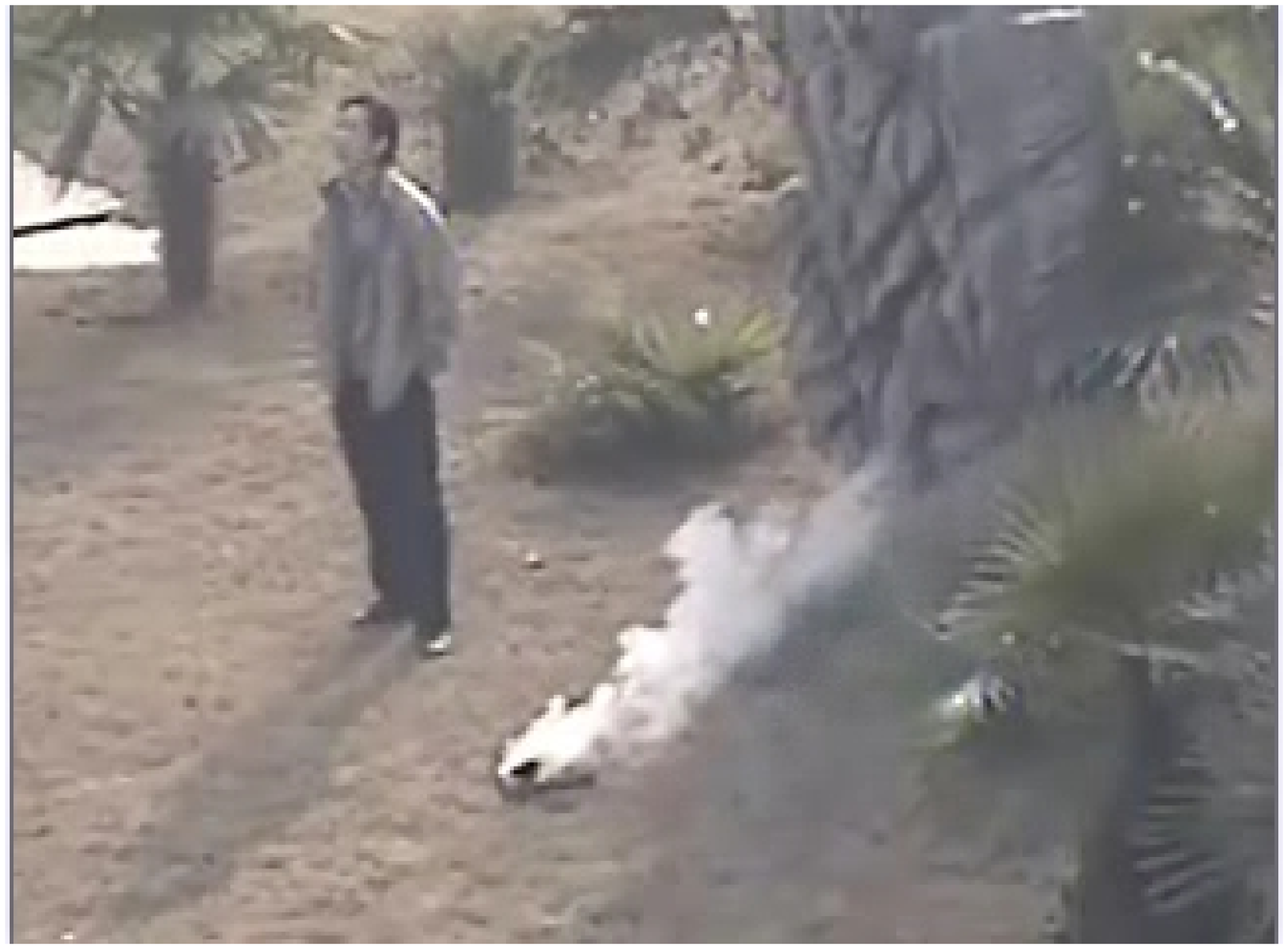}
      \label{fig_TestVideo1}
}
\subfigure[]
{
      \includegraphics[width=0.7in]{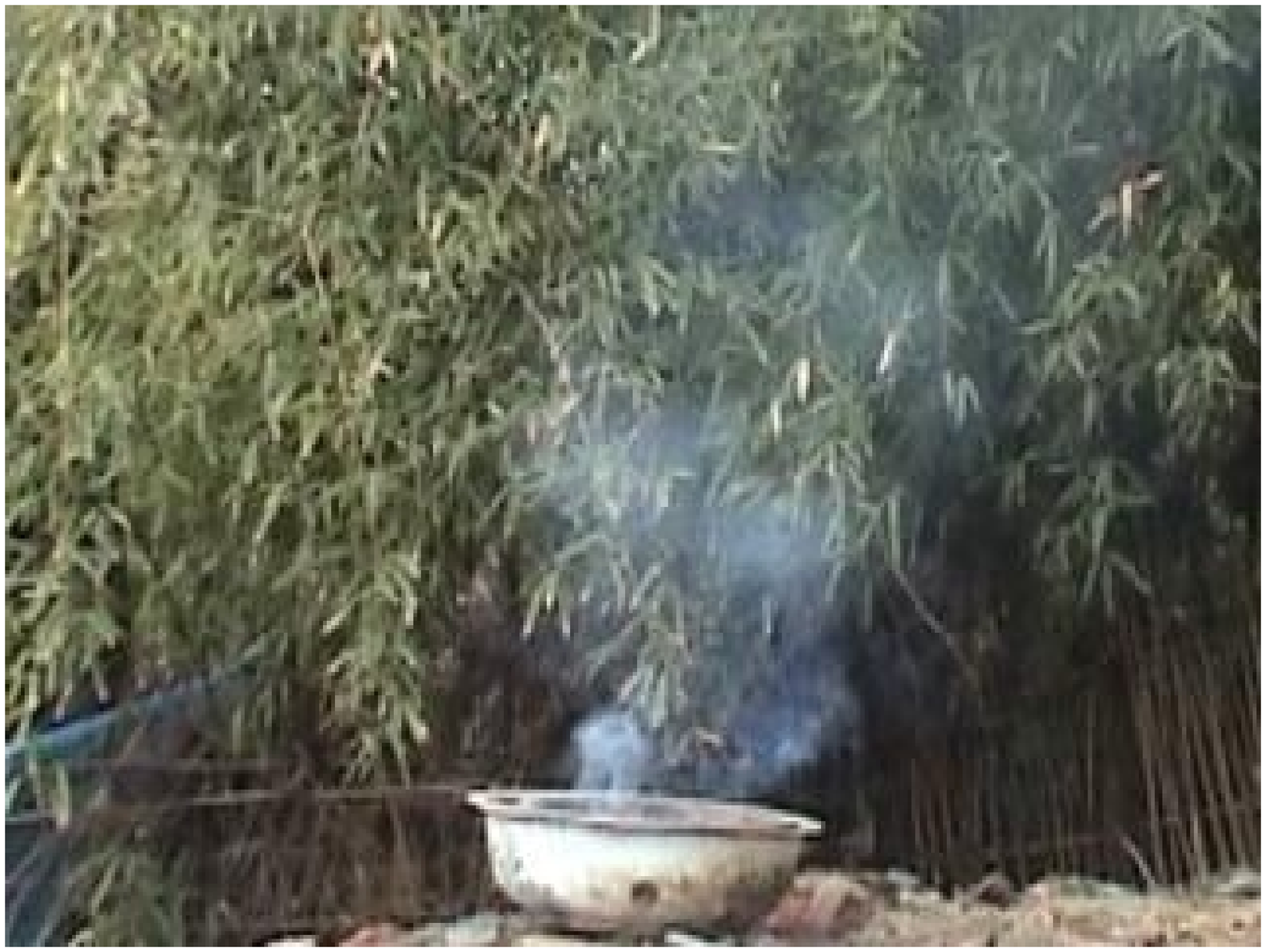}
      \label{fig_TestVideo2}
}
\subfigure[]
{
      \includegraphics[width=0.7in]{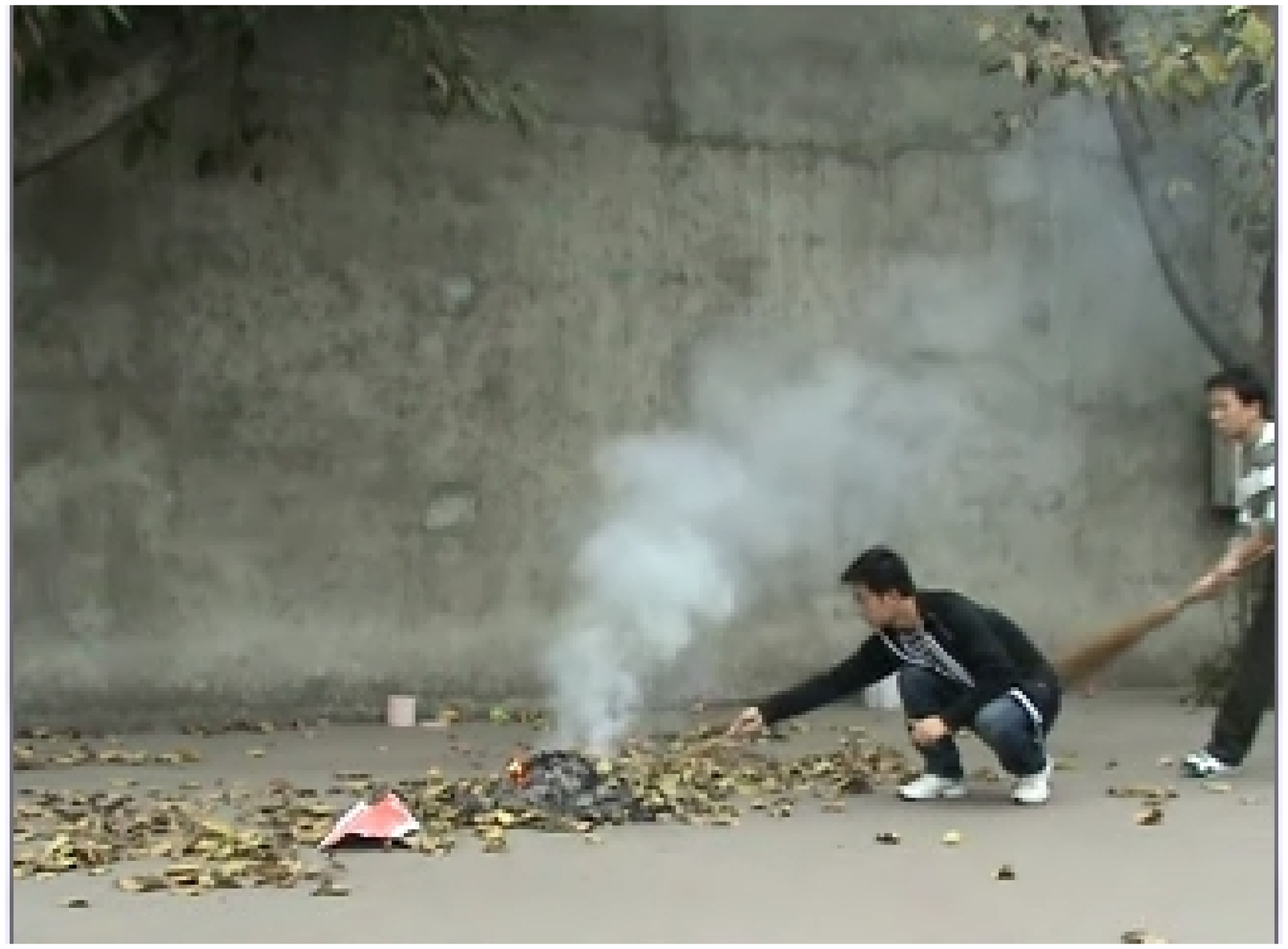}
      \label{fig_TestVideo3}
}
\subfigure[]
{
      \includegraphics[width=0.7in]{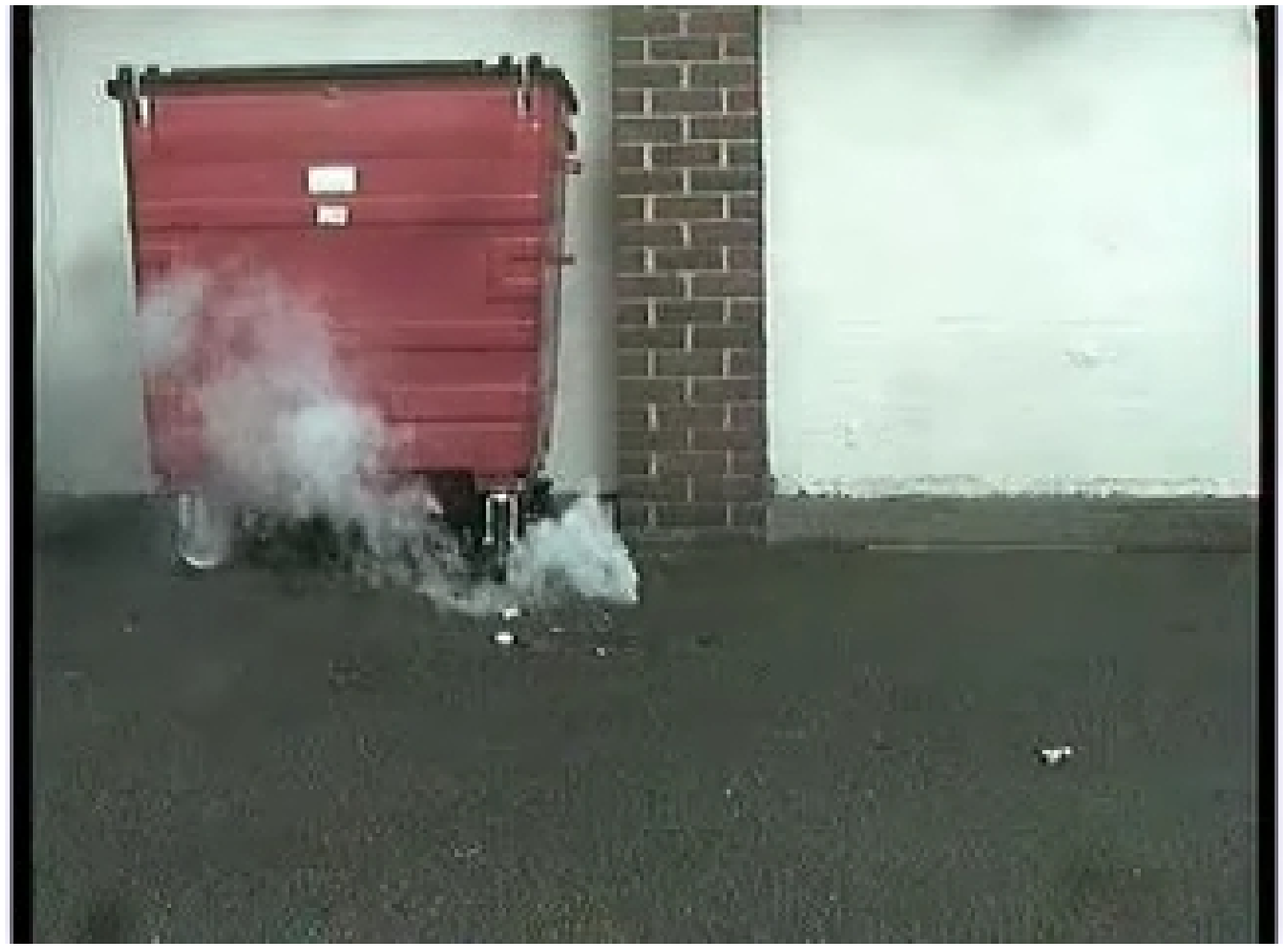}
      \label{fig_TestVideo4}
}
\subfigure[]
{
      \includegraphics[width=0.7in]{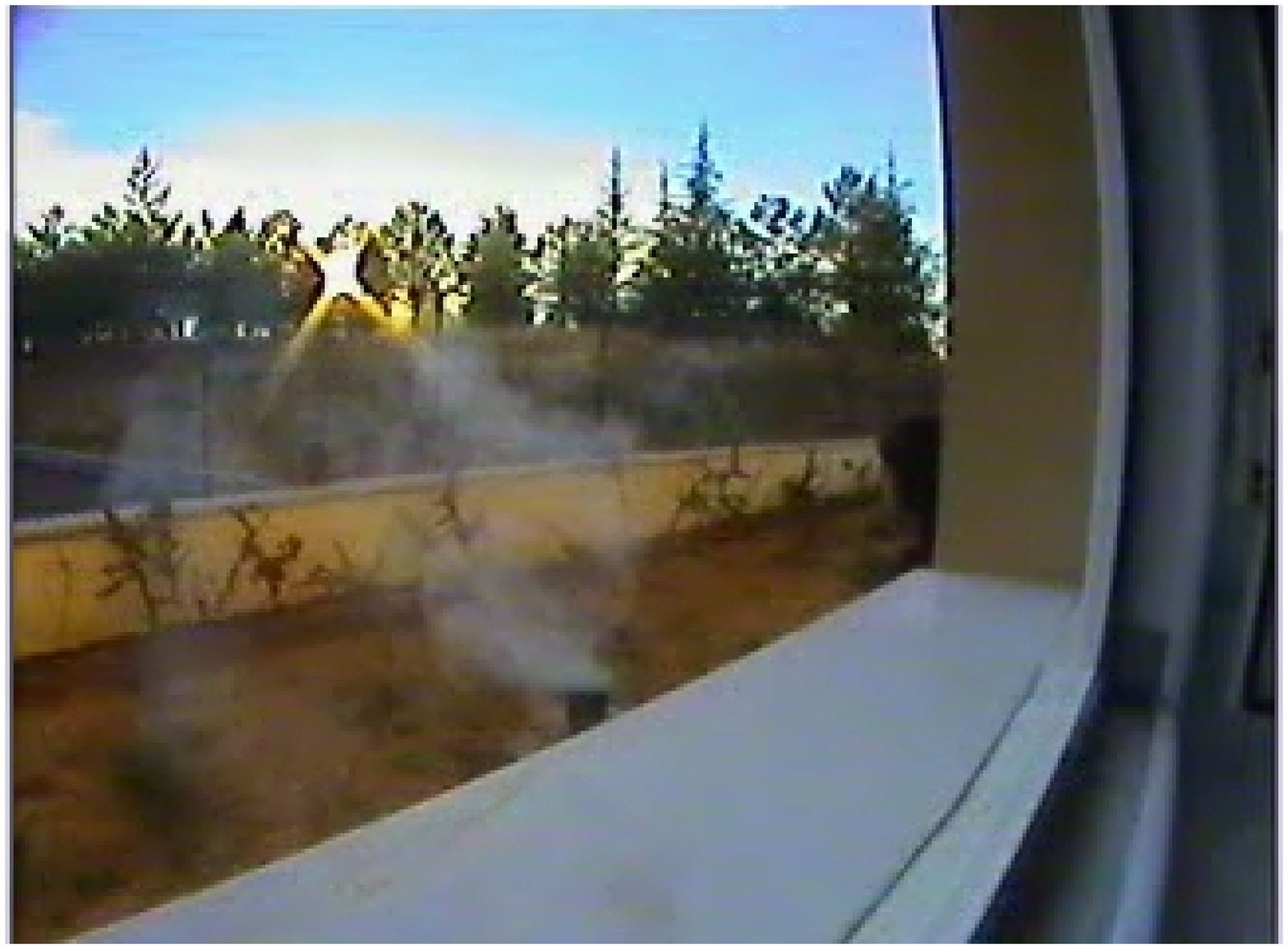}
      \label{fig_TestVideo5}
}
\subfigure[]
{
      \includegraphics[width=0.7in]{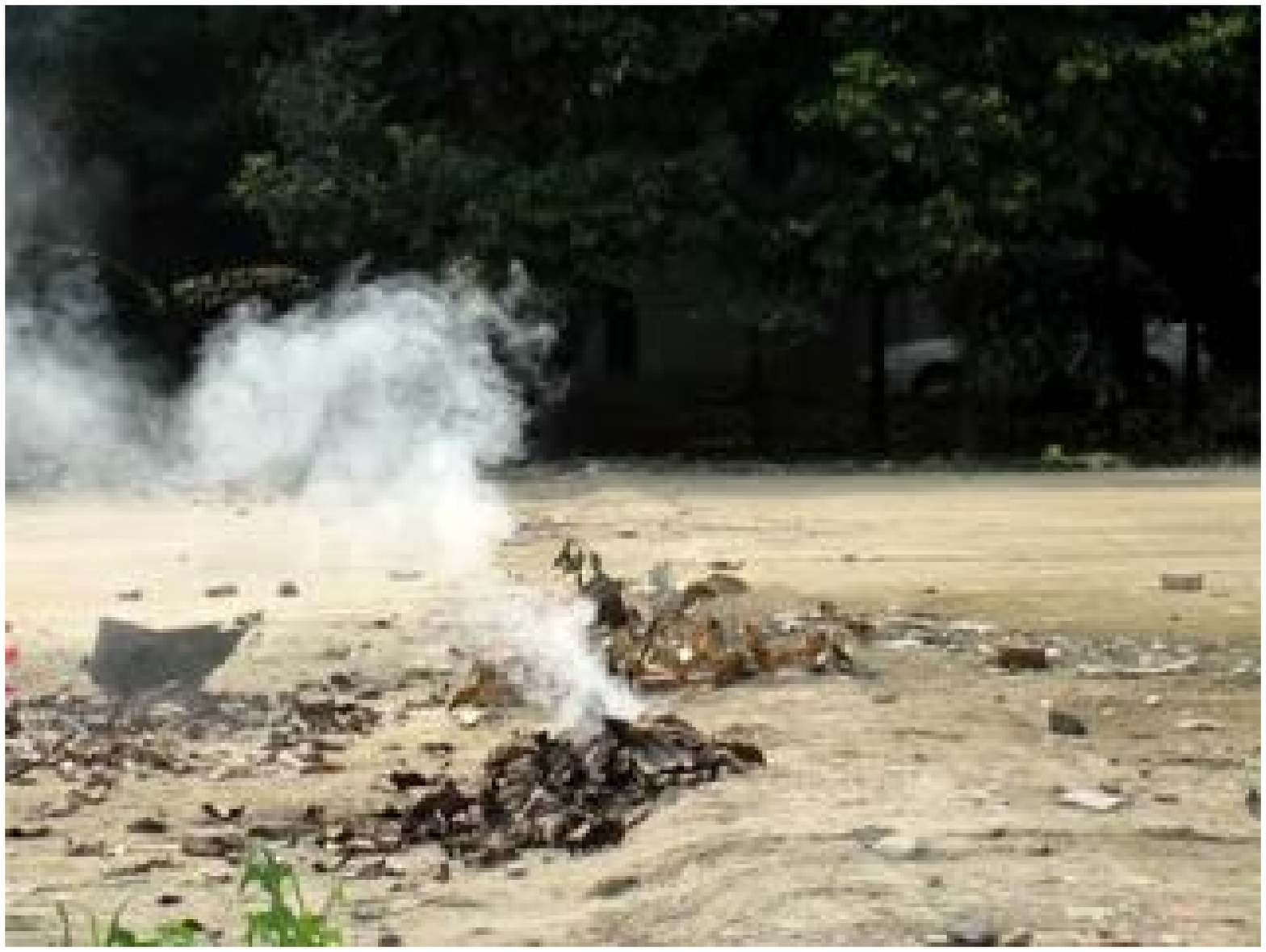}
      \label{fig_TestVideo6}
}
\subfigure[]
{
      \includegraphics[width=0.7in]{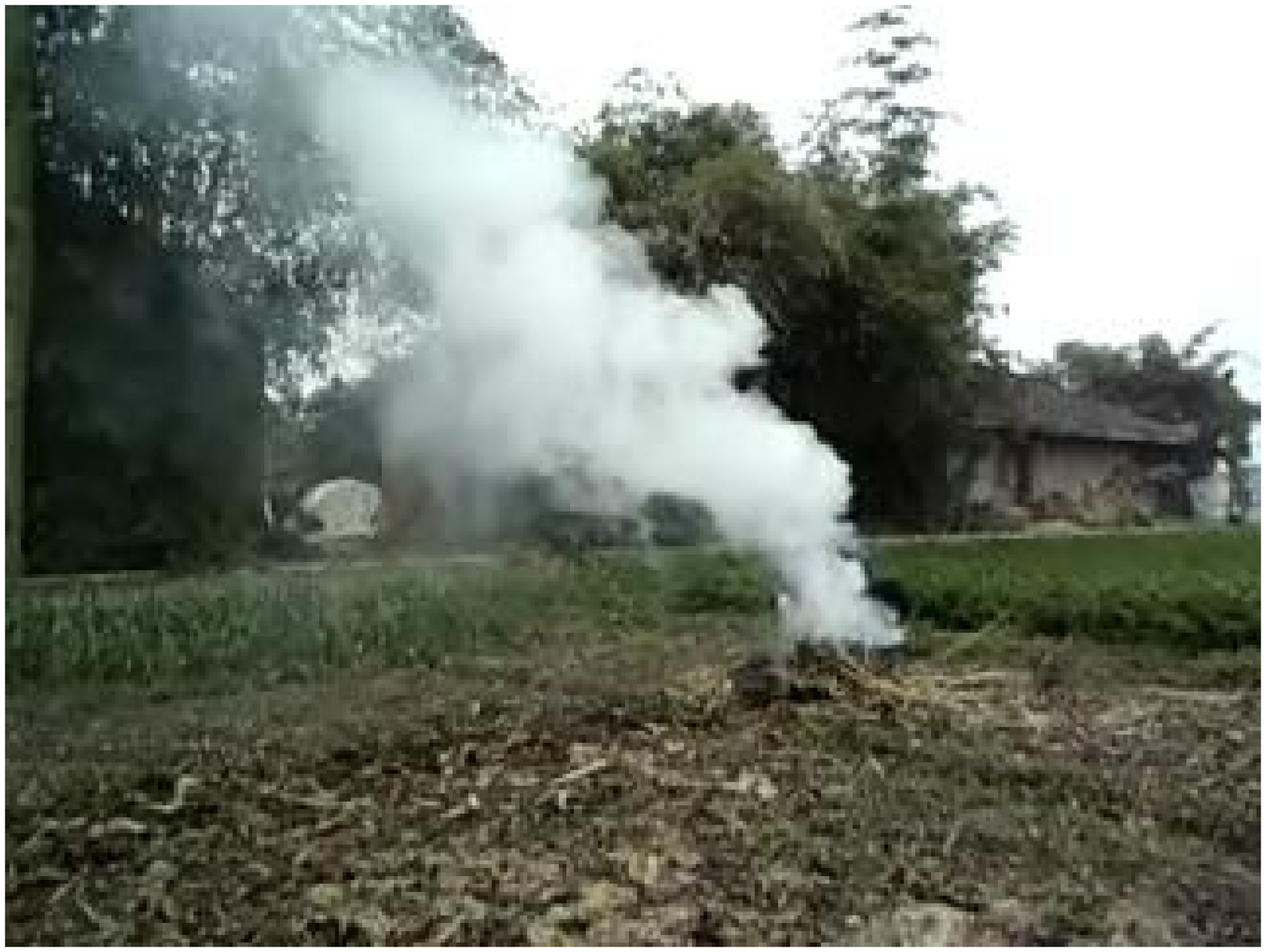}
      \label{fig_TestVideo7}
}
\subfigure[]
{
      \includegraphics[width=0.7in]{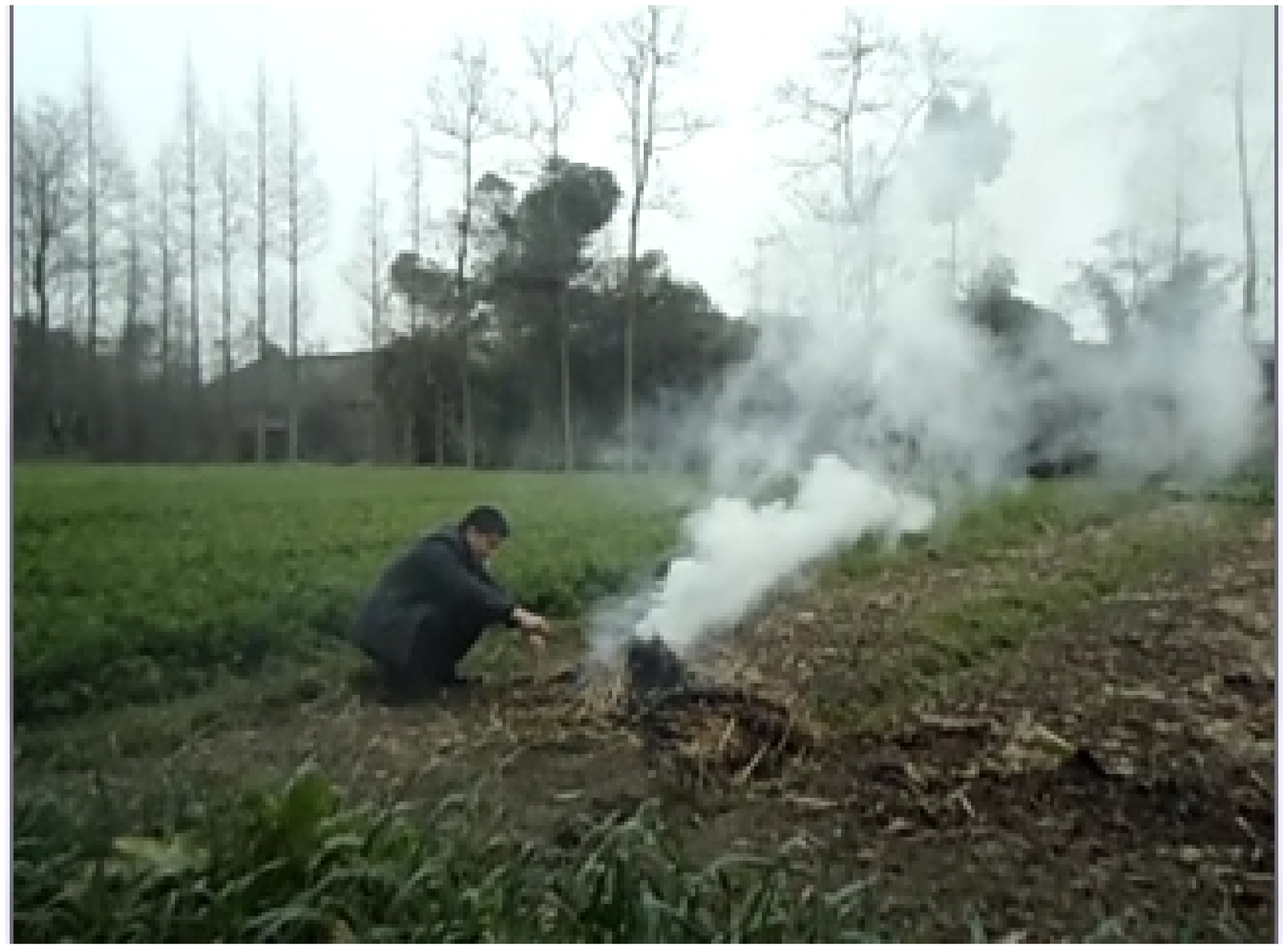}
      \label{fig_TestVideo8}
}
\subfigure[]
{
      \includegraphics[width=0.7in]{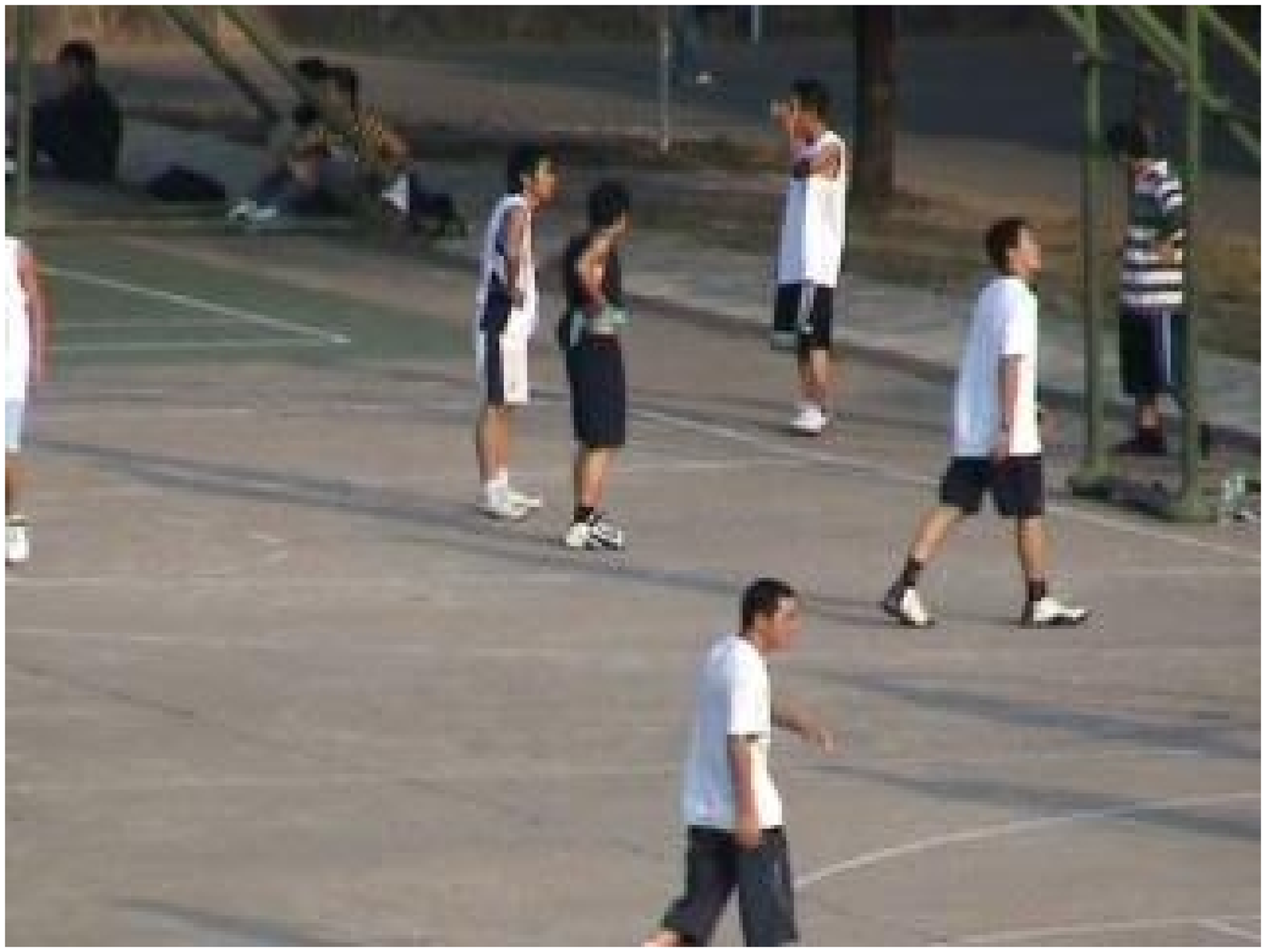}
      \label{fig_TestVideo9}
}
\subfigure[]
{
      \includegraphics[width=0.7in]{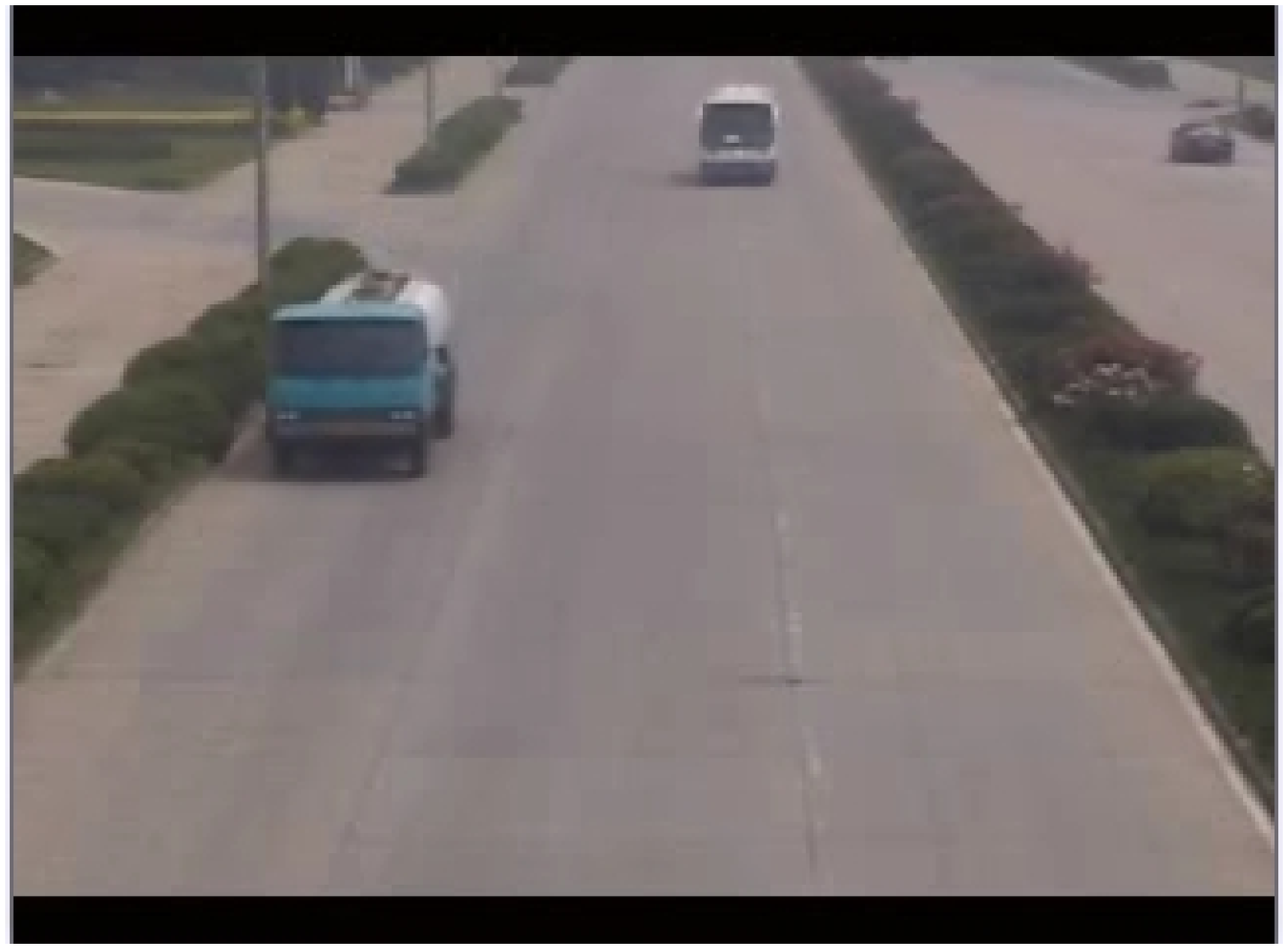}
      \label{fig_TestVideo10}
}
\subfigure[]
{
      \includegraphics[width=0.7in]{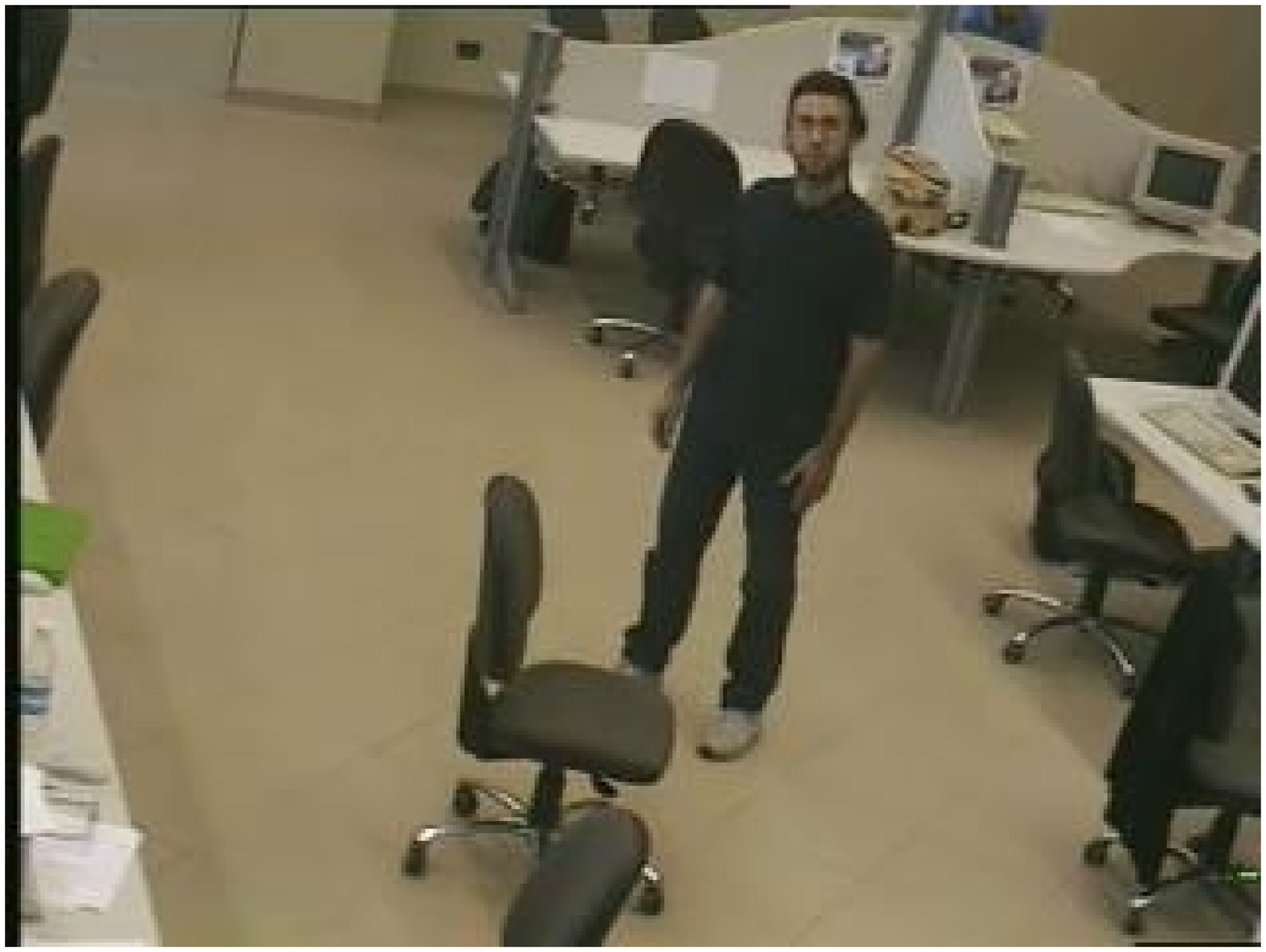}
      \label{fig_TestVideo11}
}
\subfigure[]
{
      \includegraphics[width=0.7in]{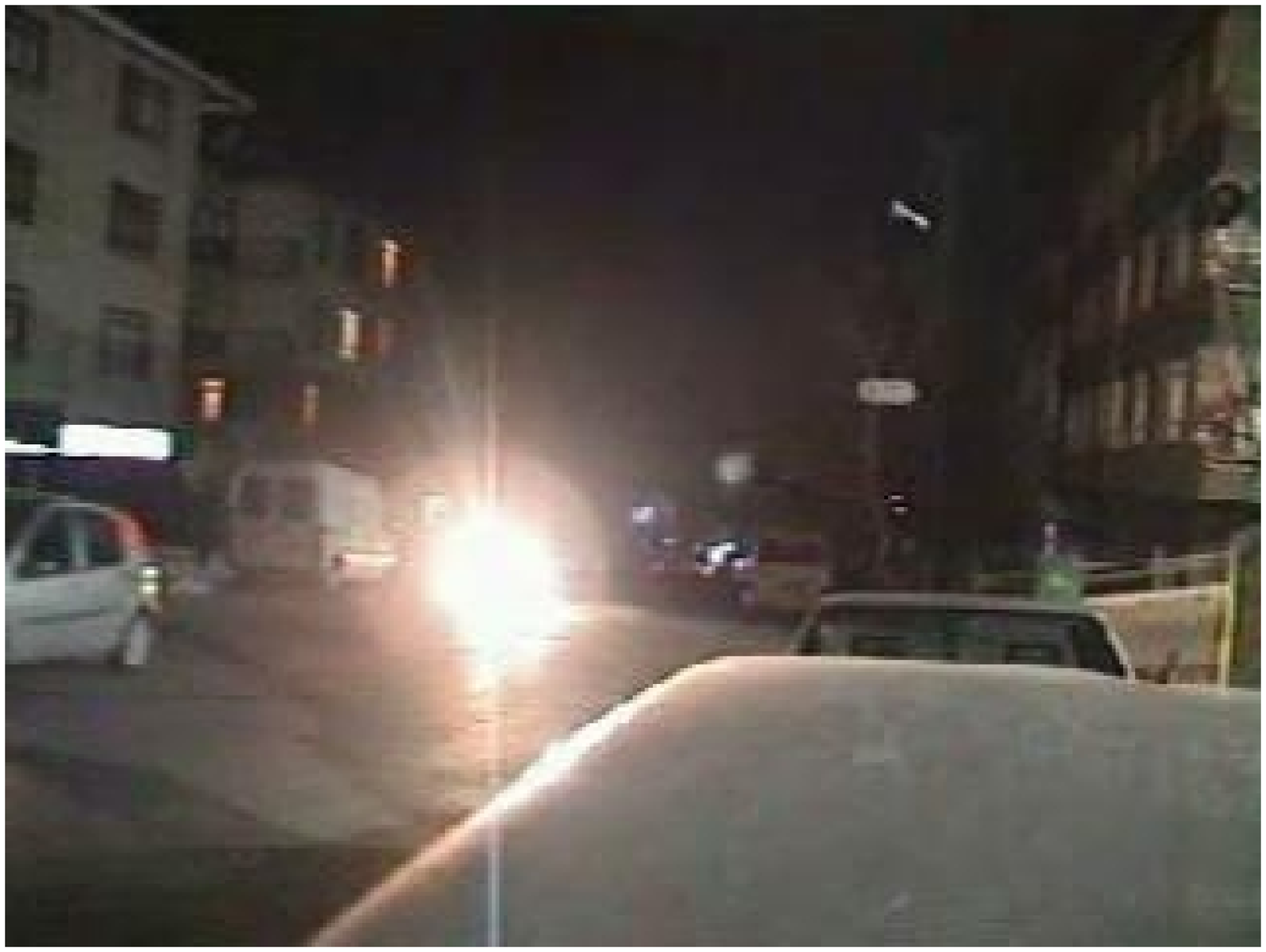}
      \label{fig_TestVideo12}
}
\subfigure[]
{
      \includegraphics[width=0.7in]{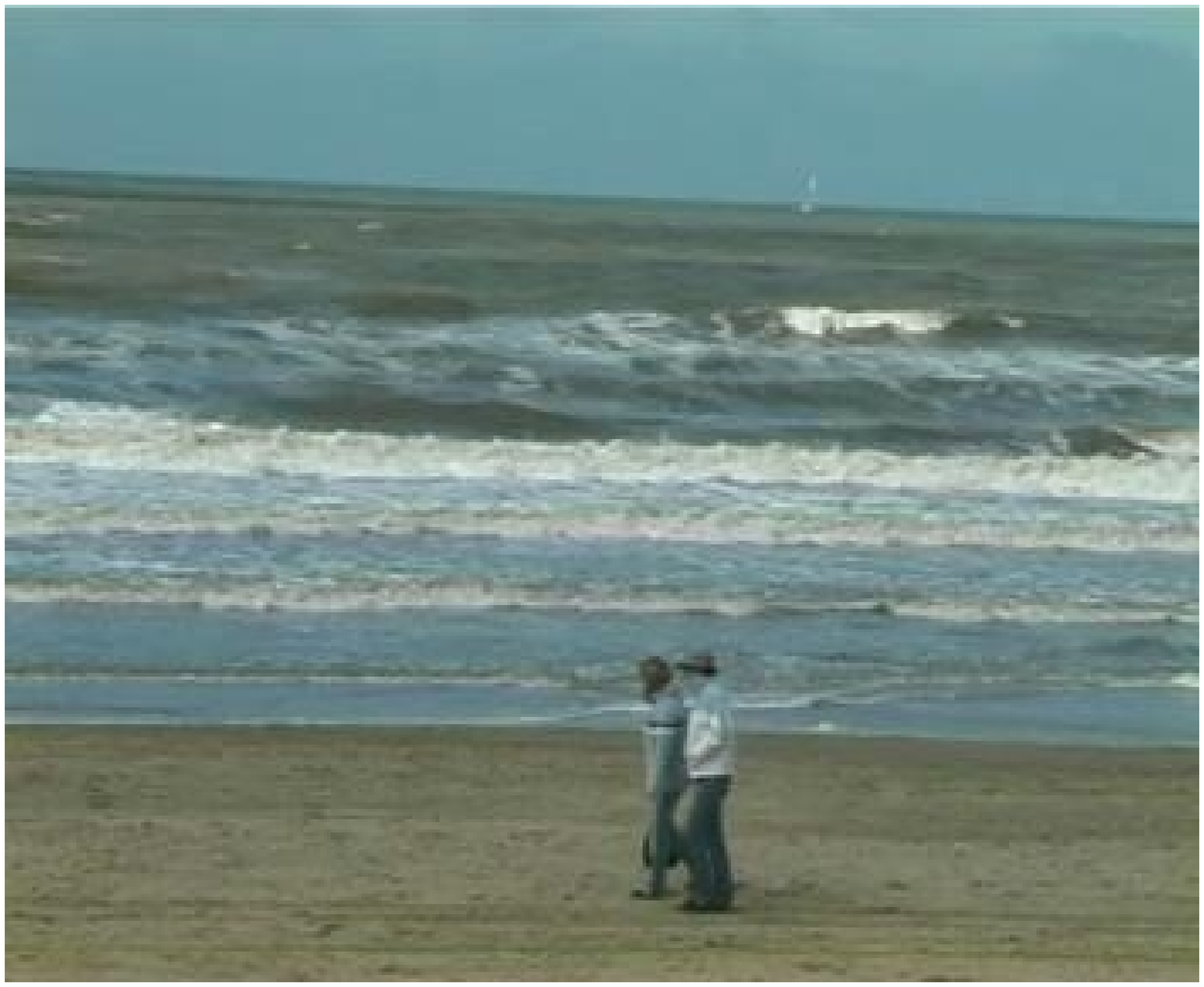}
      \label{fig_TestVideo13}
}
\subfigure[]
{
      \includegraphics[width=0.7in]{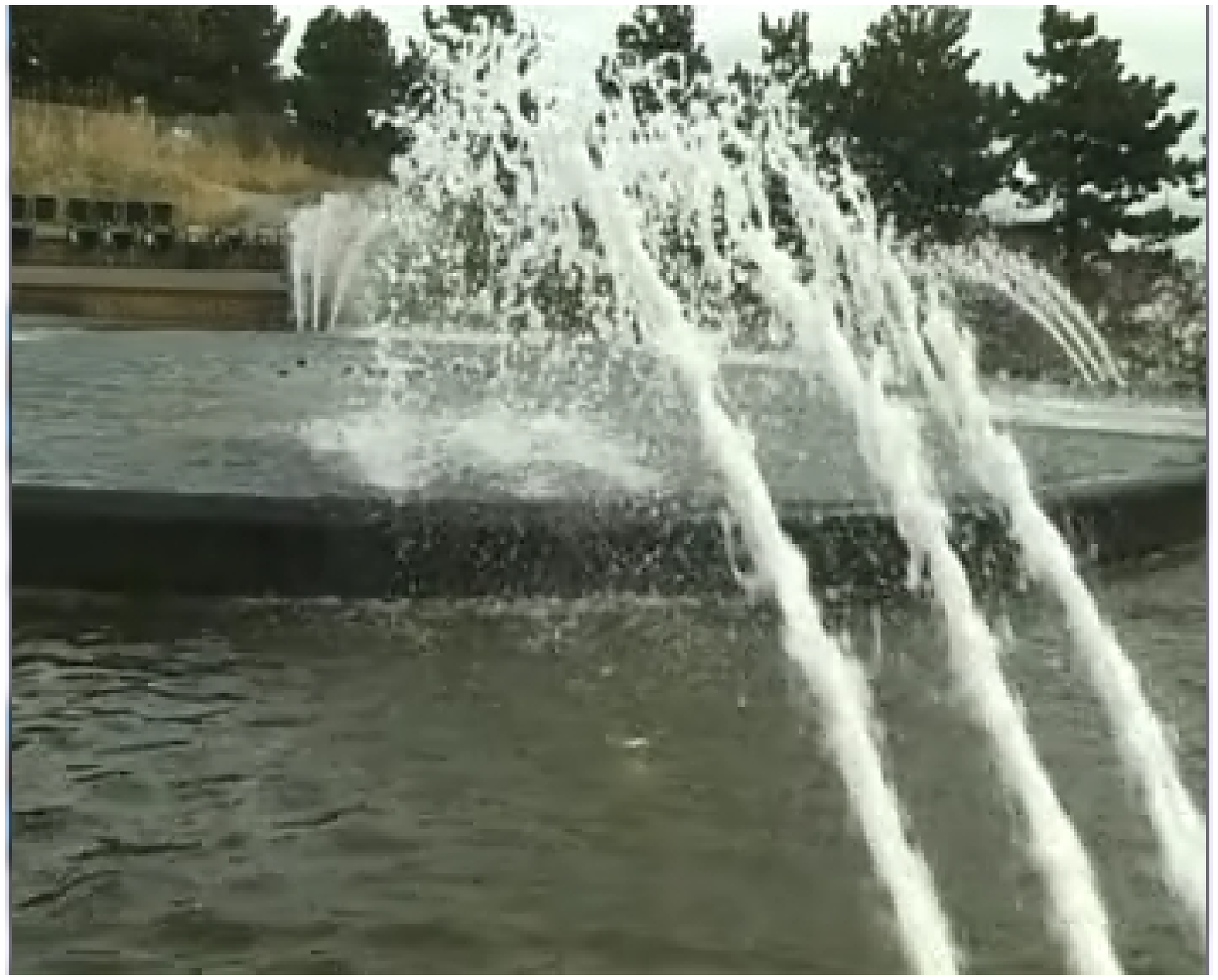}
      \label{fig_TestVideo14}
}
\subfigure[]
{
      \includegraphics[width=0.7in]{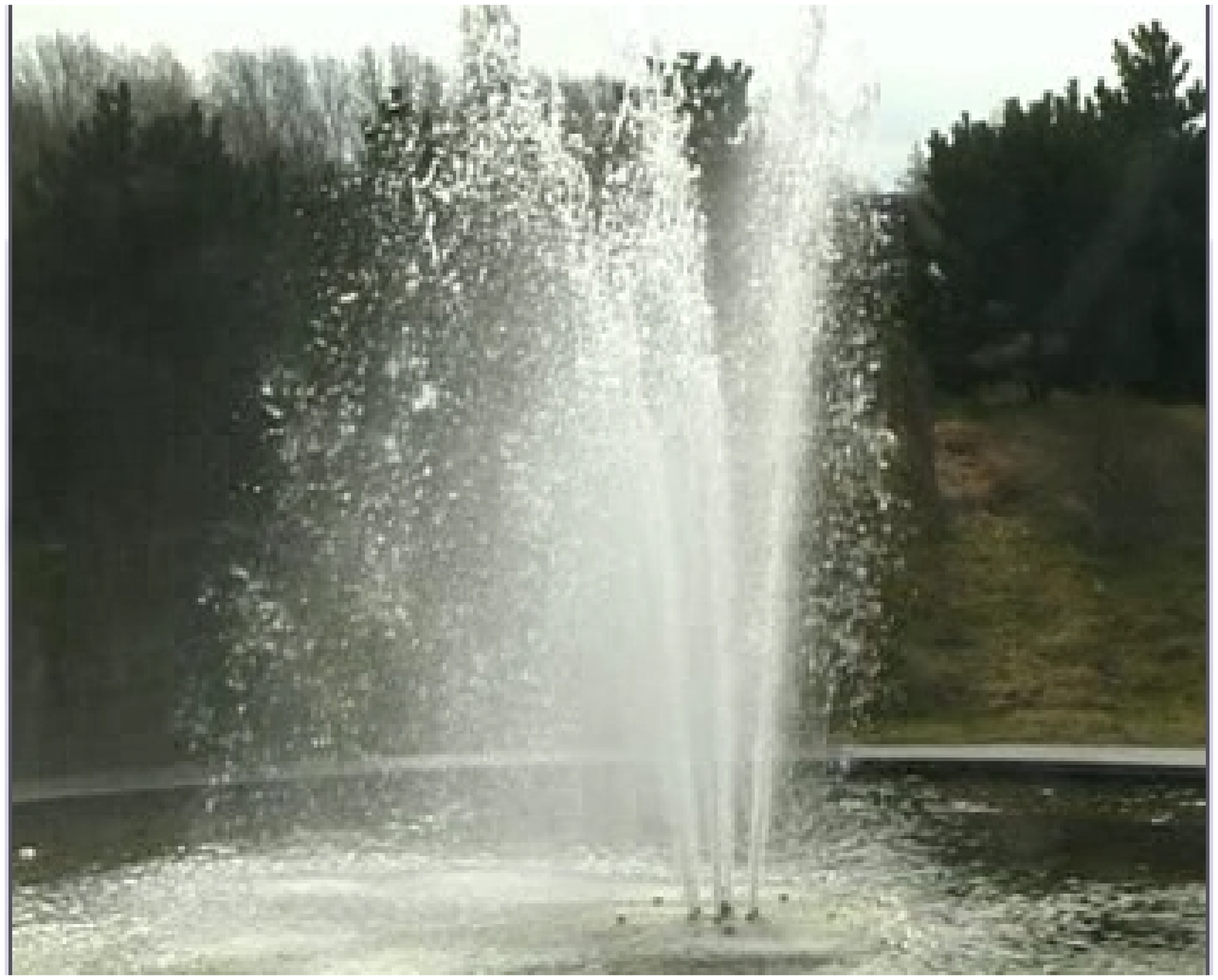}
      \label{fig_TestVideo15}
}
\subfigure[]
{
      \includegraphics[width=0.7in]{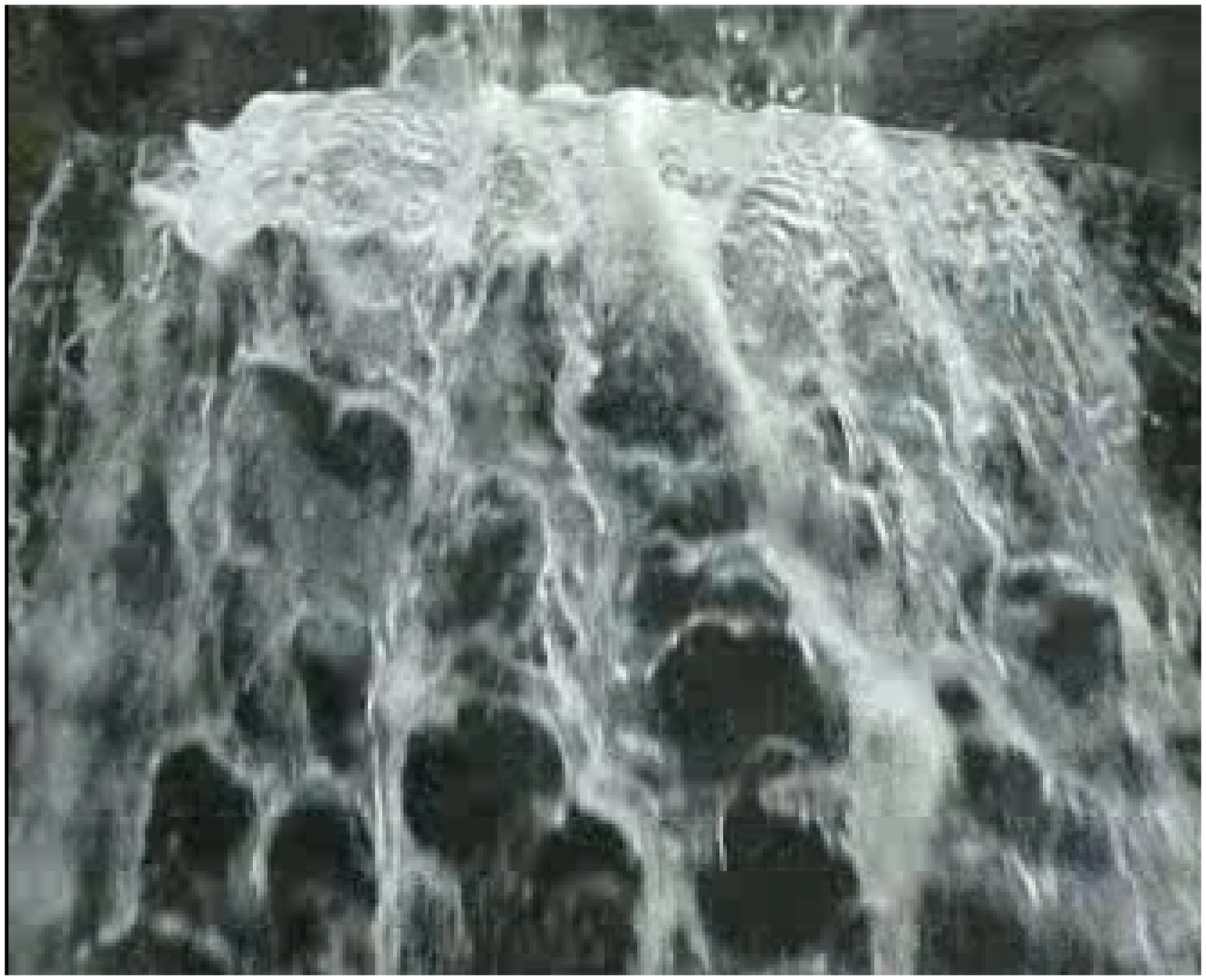}
      \label{fig_TestVideo16}
}
\caption{Examples of the video data set, the first two rows are smoke videos, others are non-smoke.}
\label{fig:TestVideos}
\end{figure*}

\begin{figure*}
\centering
\subfigure[]
{
      \includegraphics[width=0.7in]{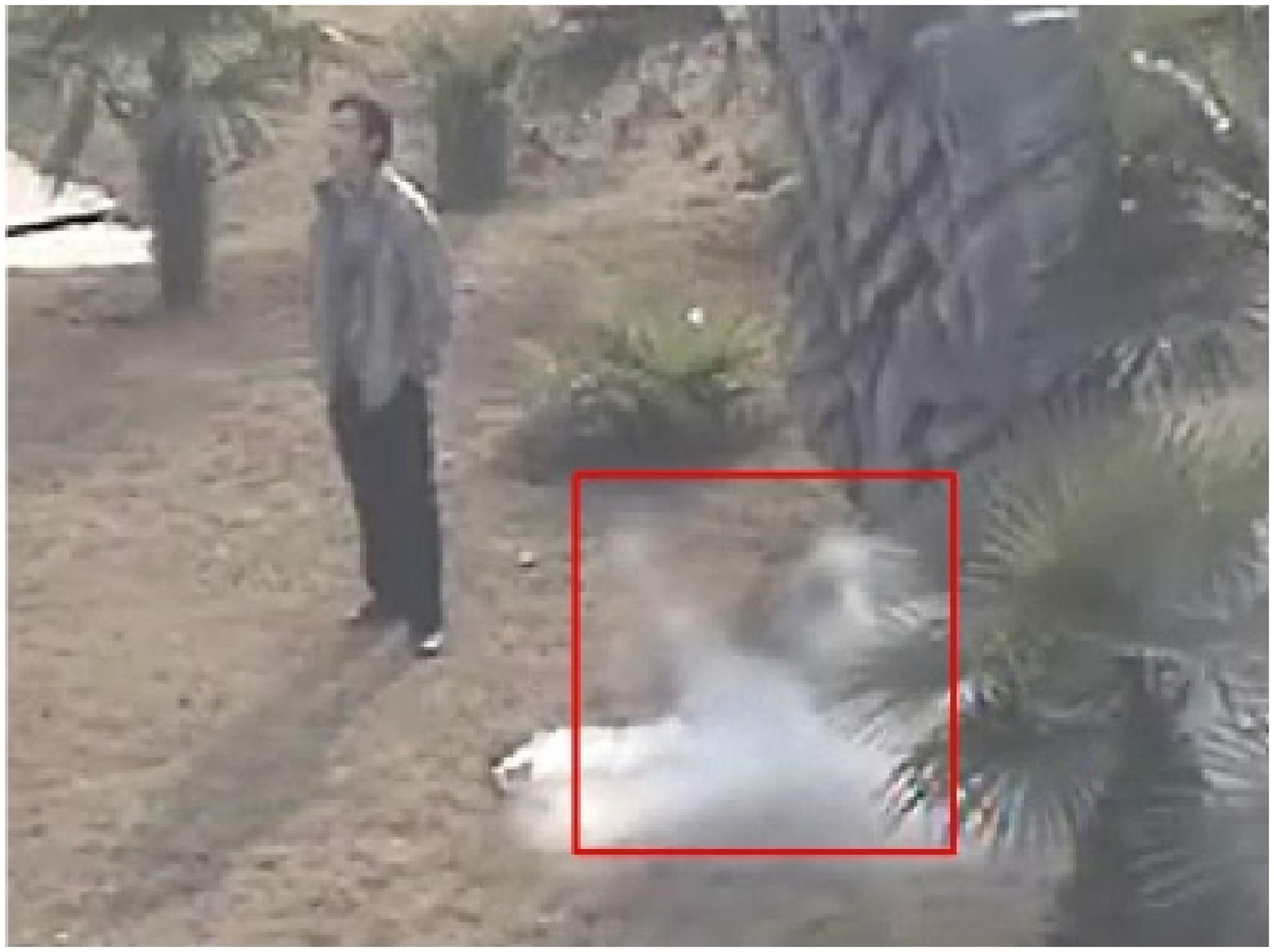}
}
\subfigure[]
{
      \includegraphics[width=0.7in]{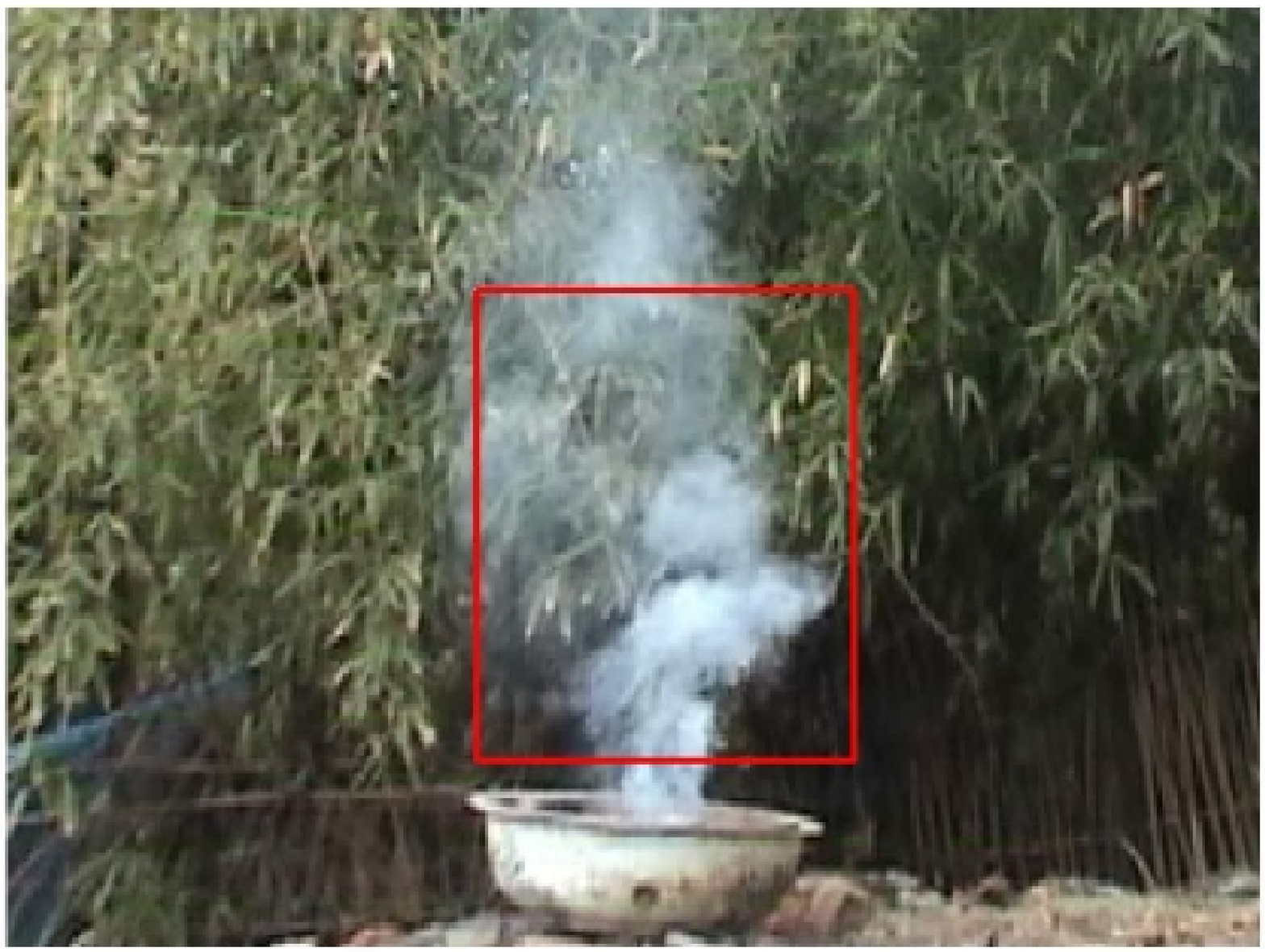}
}
\subfigure[]
{
      \includegraphics[width=0.7in]{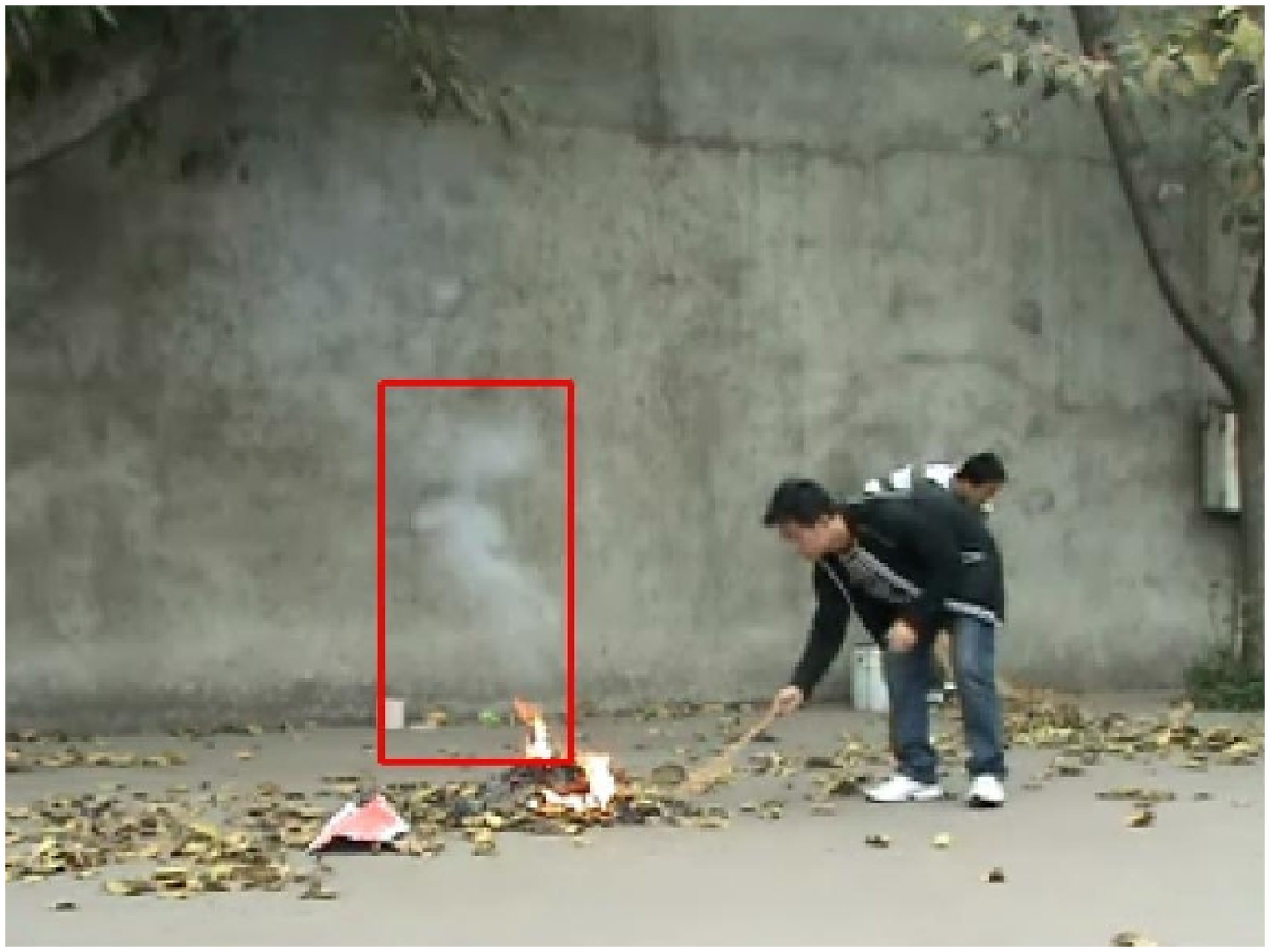}
}
\subfigure[]
{
      \includegraphics[width=0.7in]{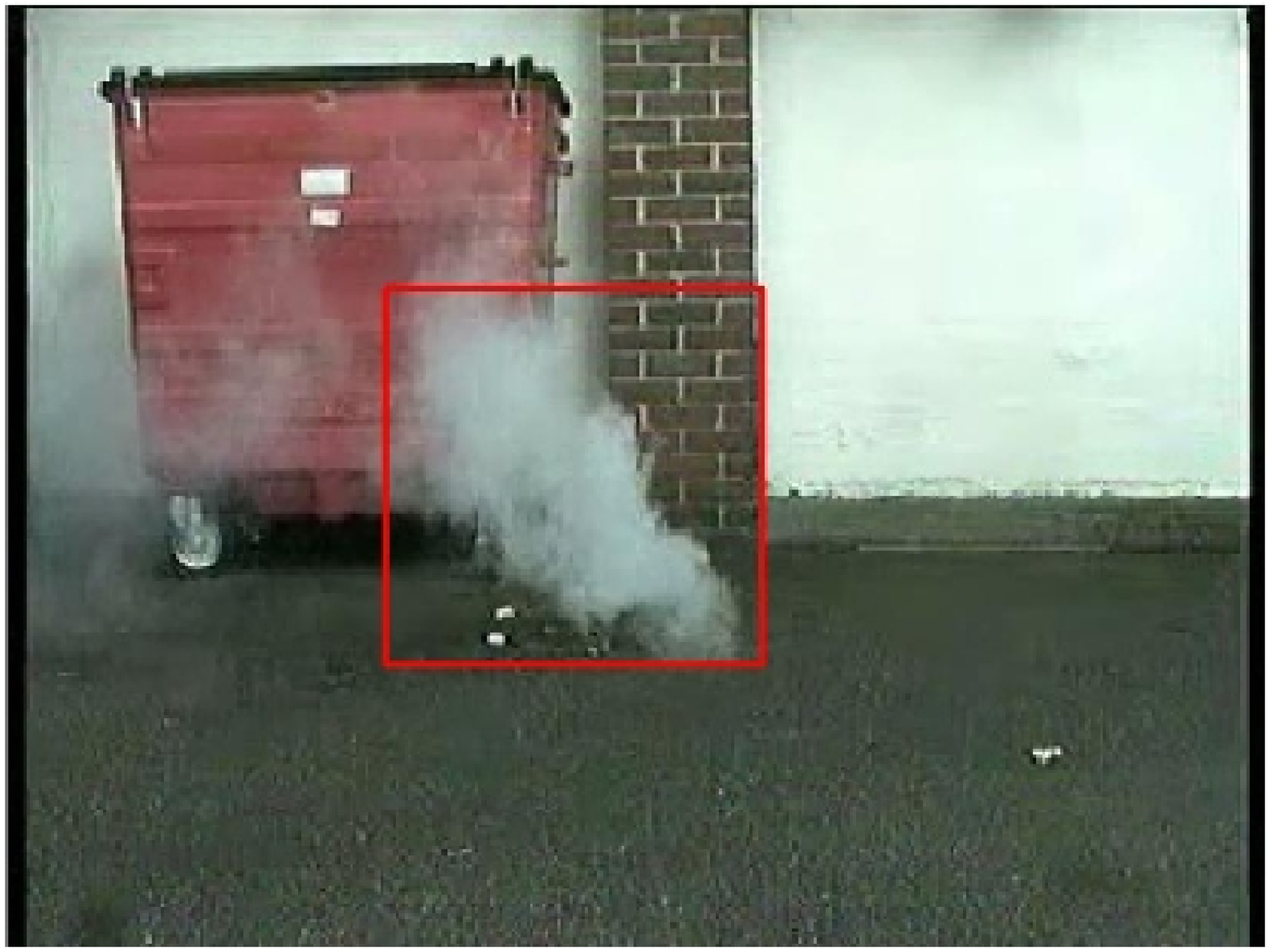}
}
\subfigure[]
{
      \includegraphics[width=0.7in]{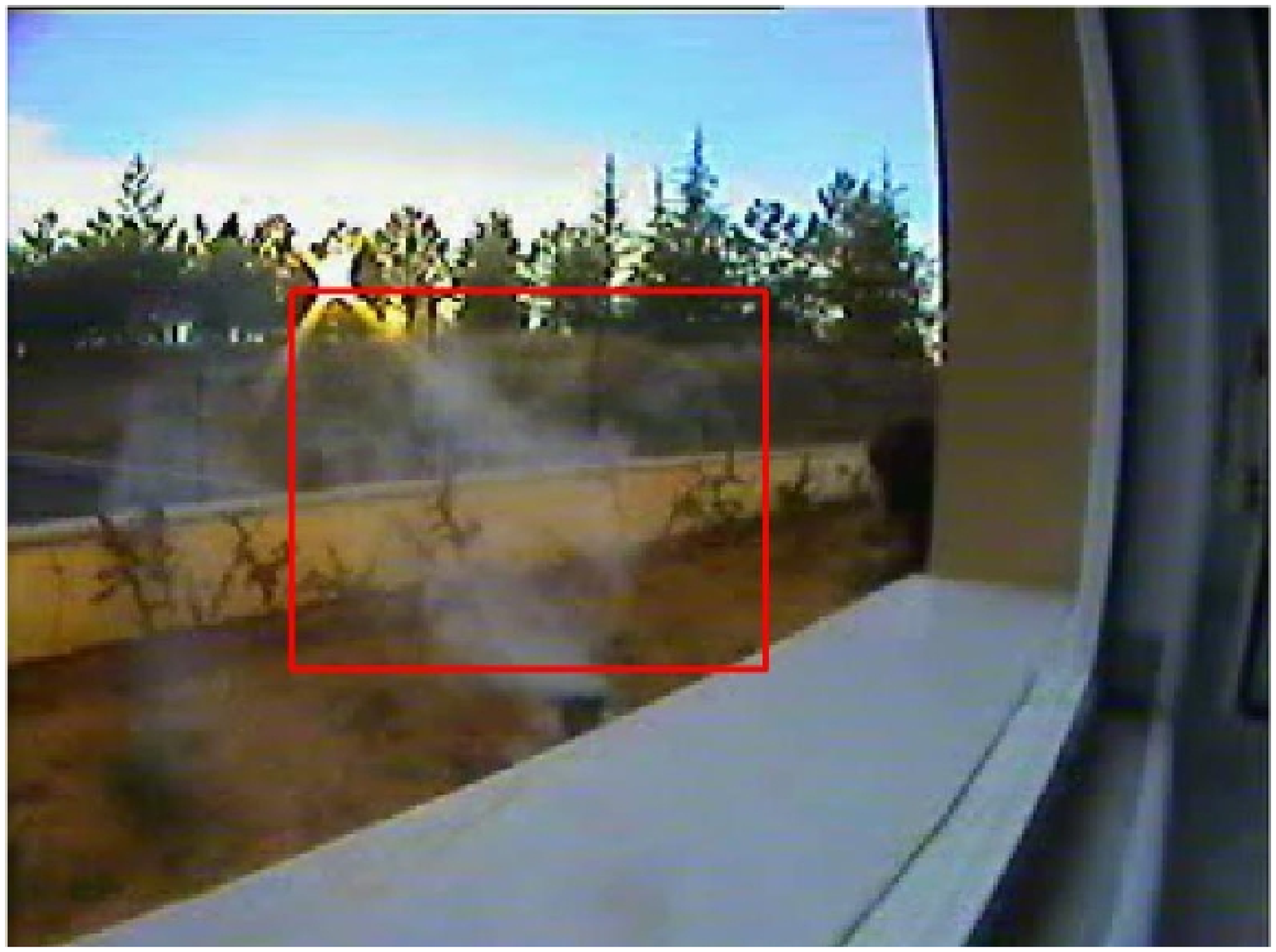}
}
\subfigure[]
{
      \includegraphics[width=0.7in]{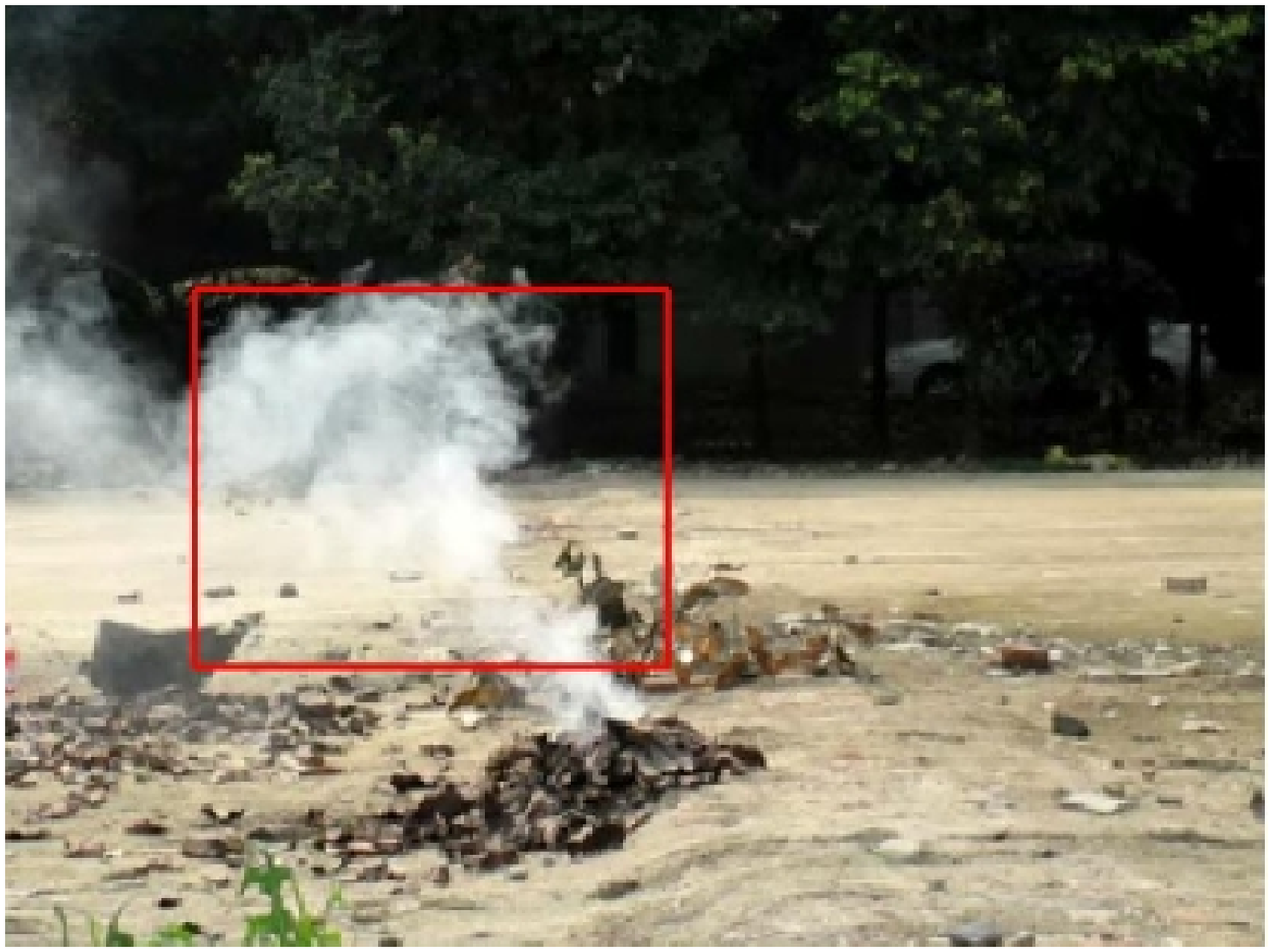}
}
\subfigure[]
{
      \includegraphics[width=0.7in]{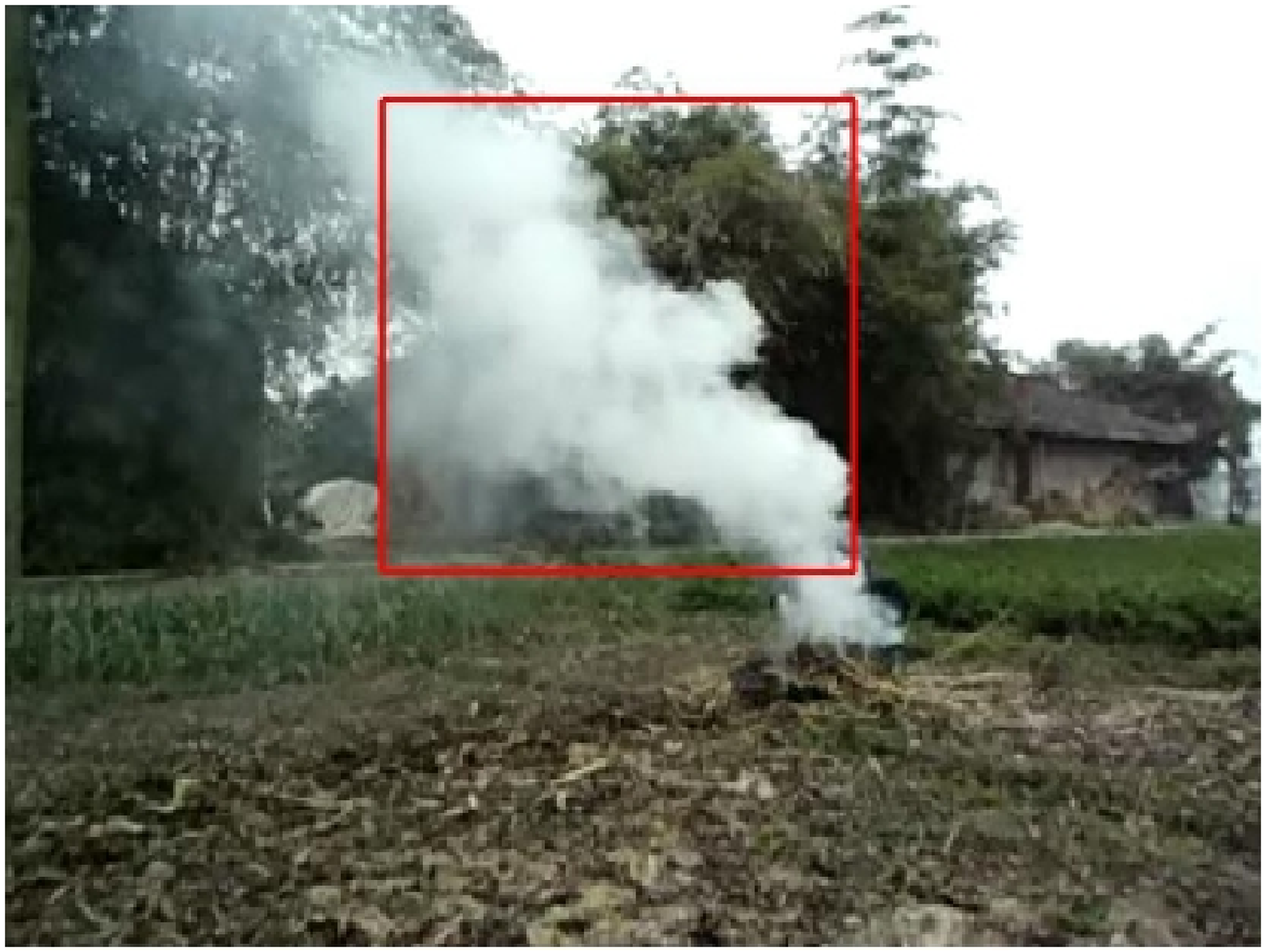}
}
\subfigure[]
{
      \includegraphics[width=0.7in]{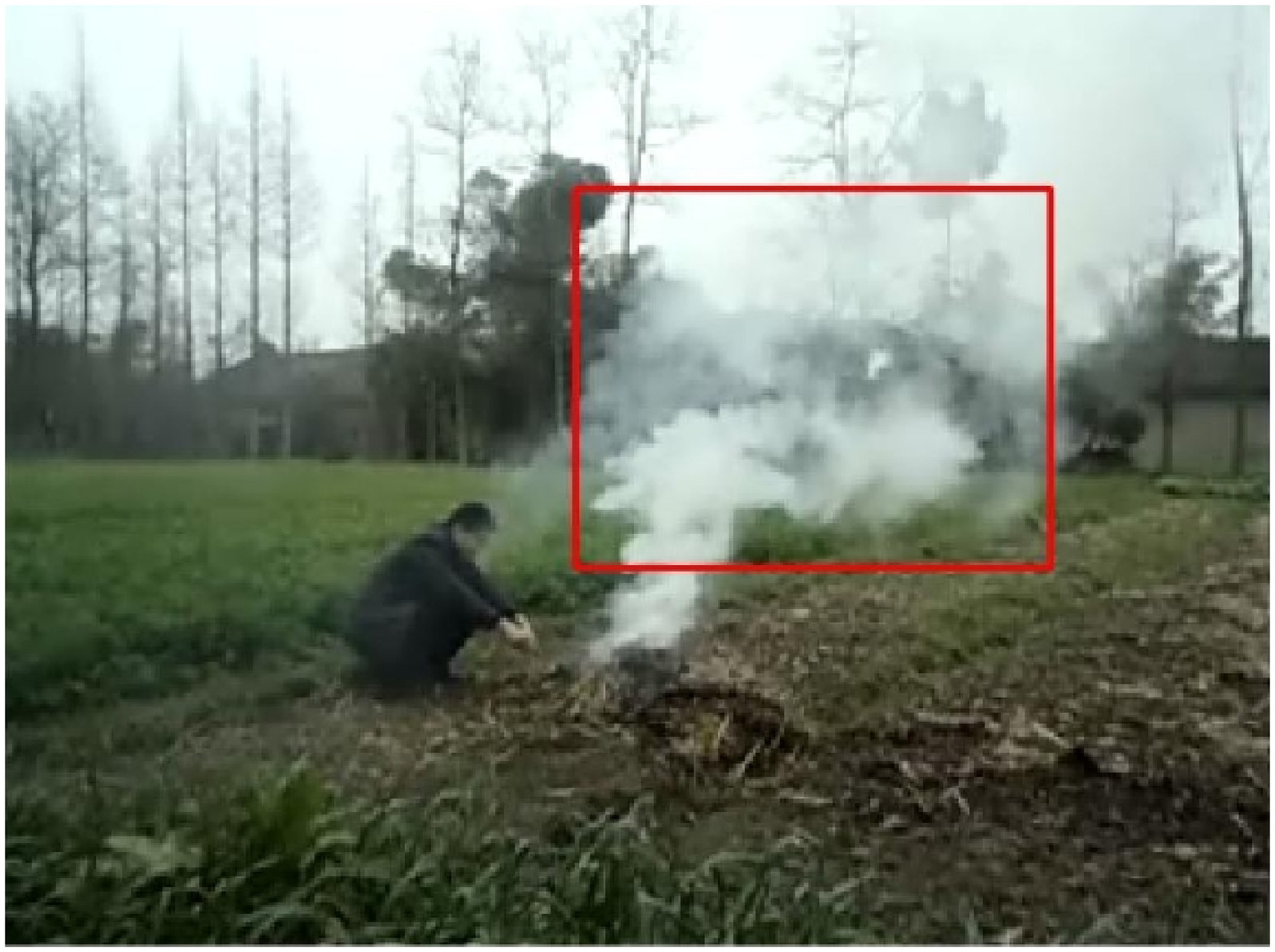}
}
\subfigure[]
{
      \includegraphics[width=0.7in]{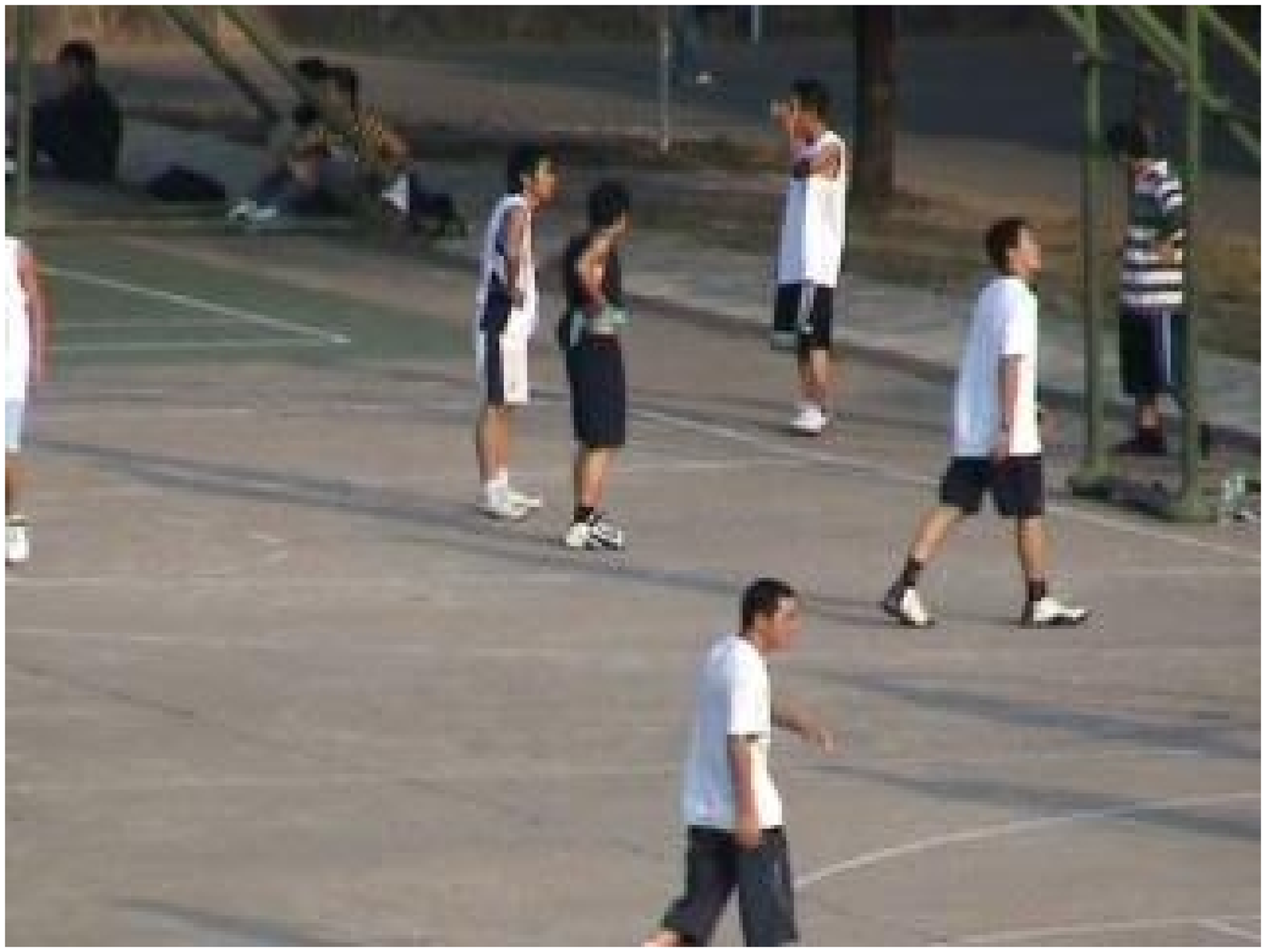}
}
\subfigure[]
{
      \includegraphics[width=0.7in]{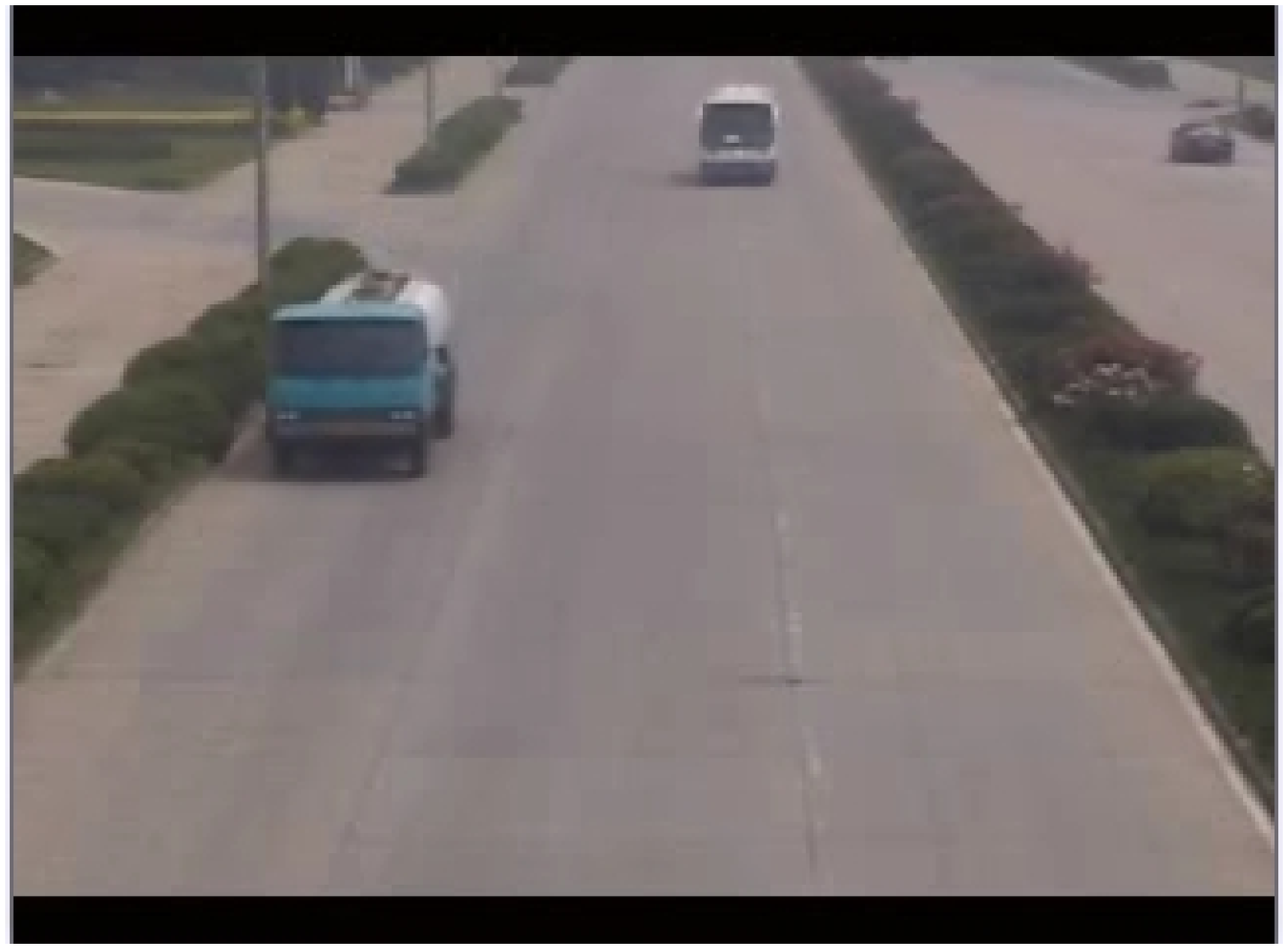}
}
\subfigure[]
{
      \includegraphics[width=0.7in]{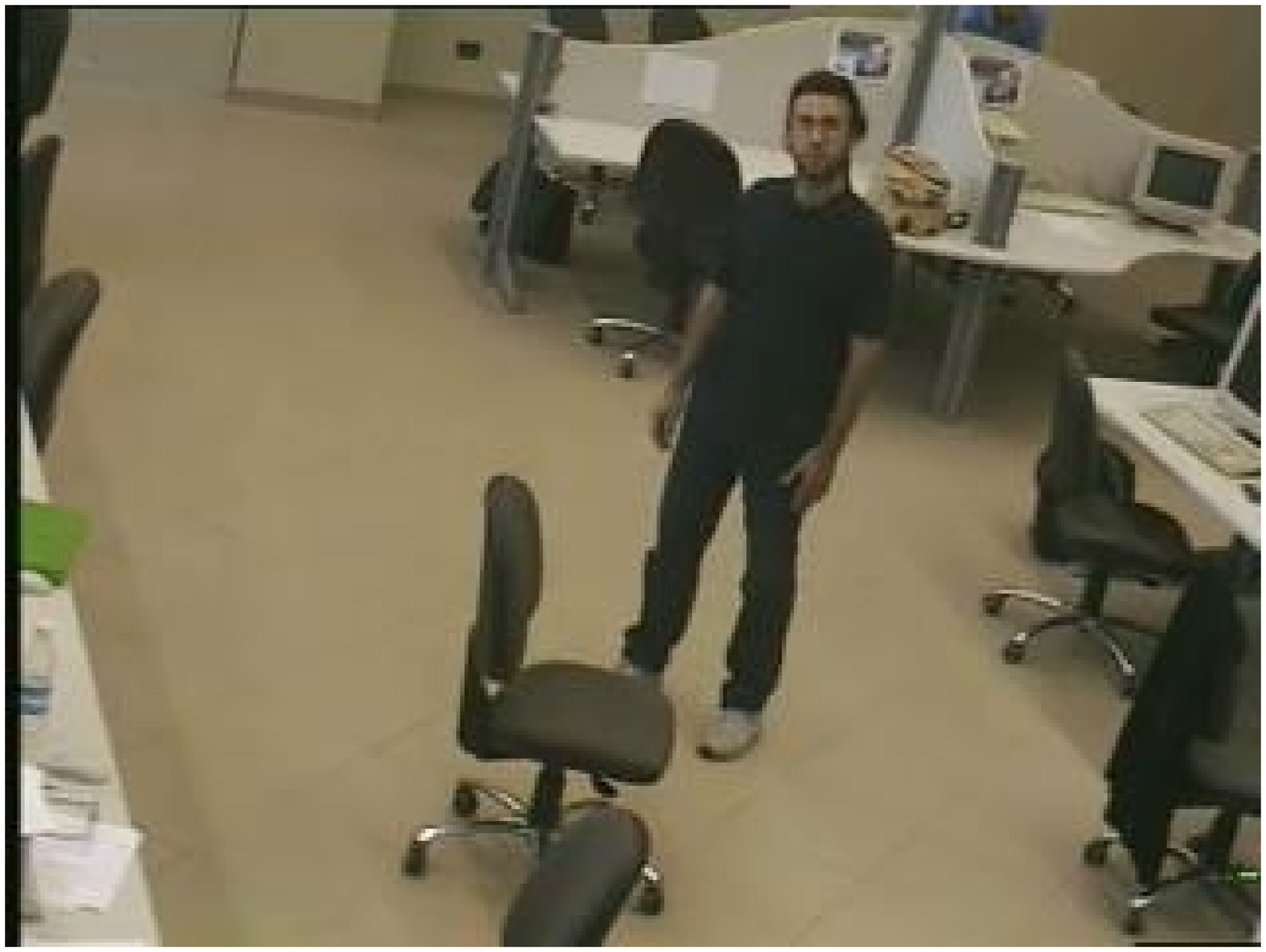}
}
\subfigure[]
{
      \includegraphics[width=0.7in]{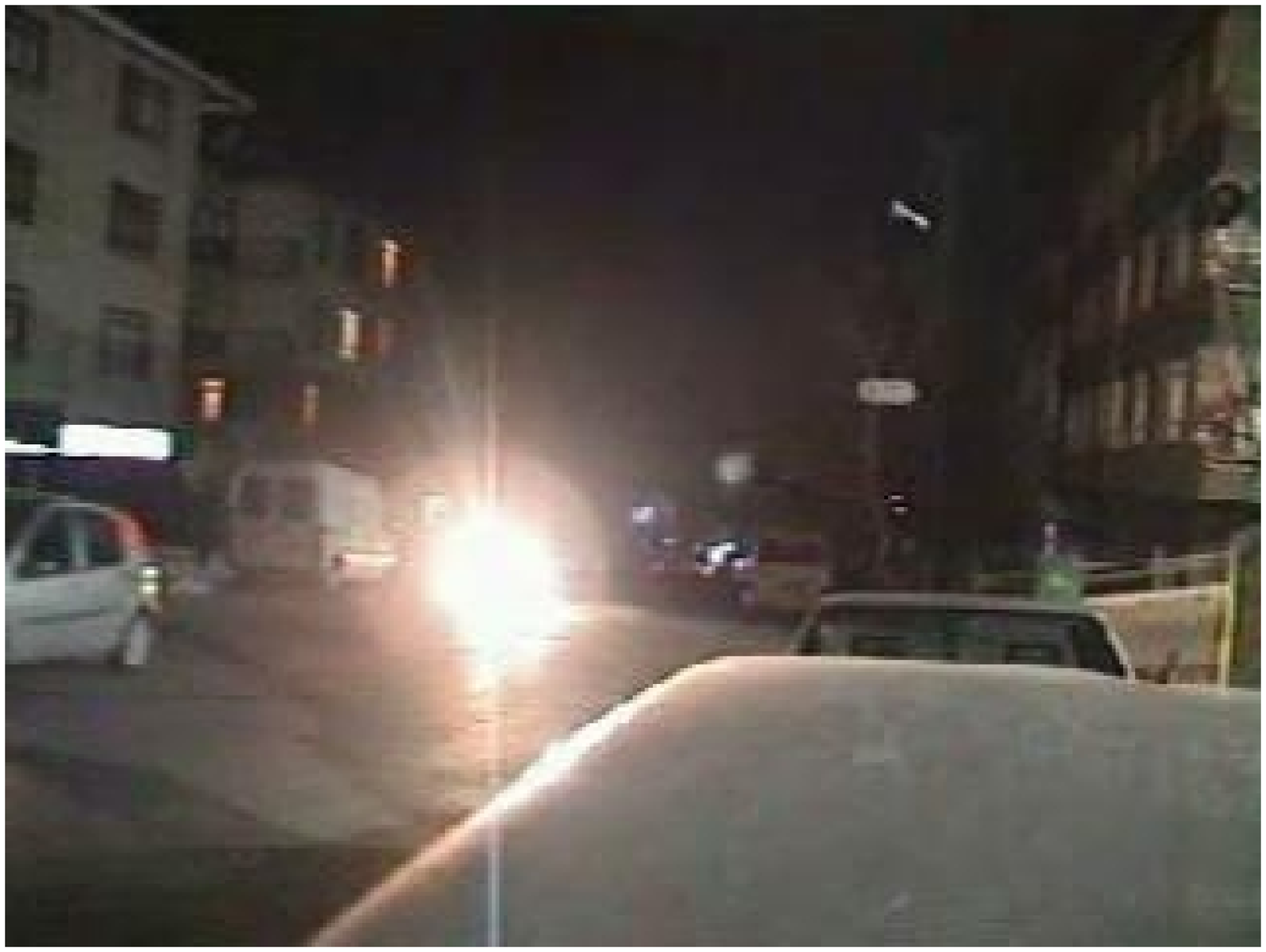}
}
\subfigure[]
{
      \includegraphics[width=0.7in]{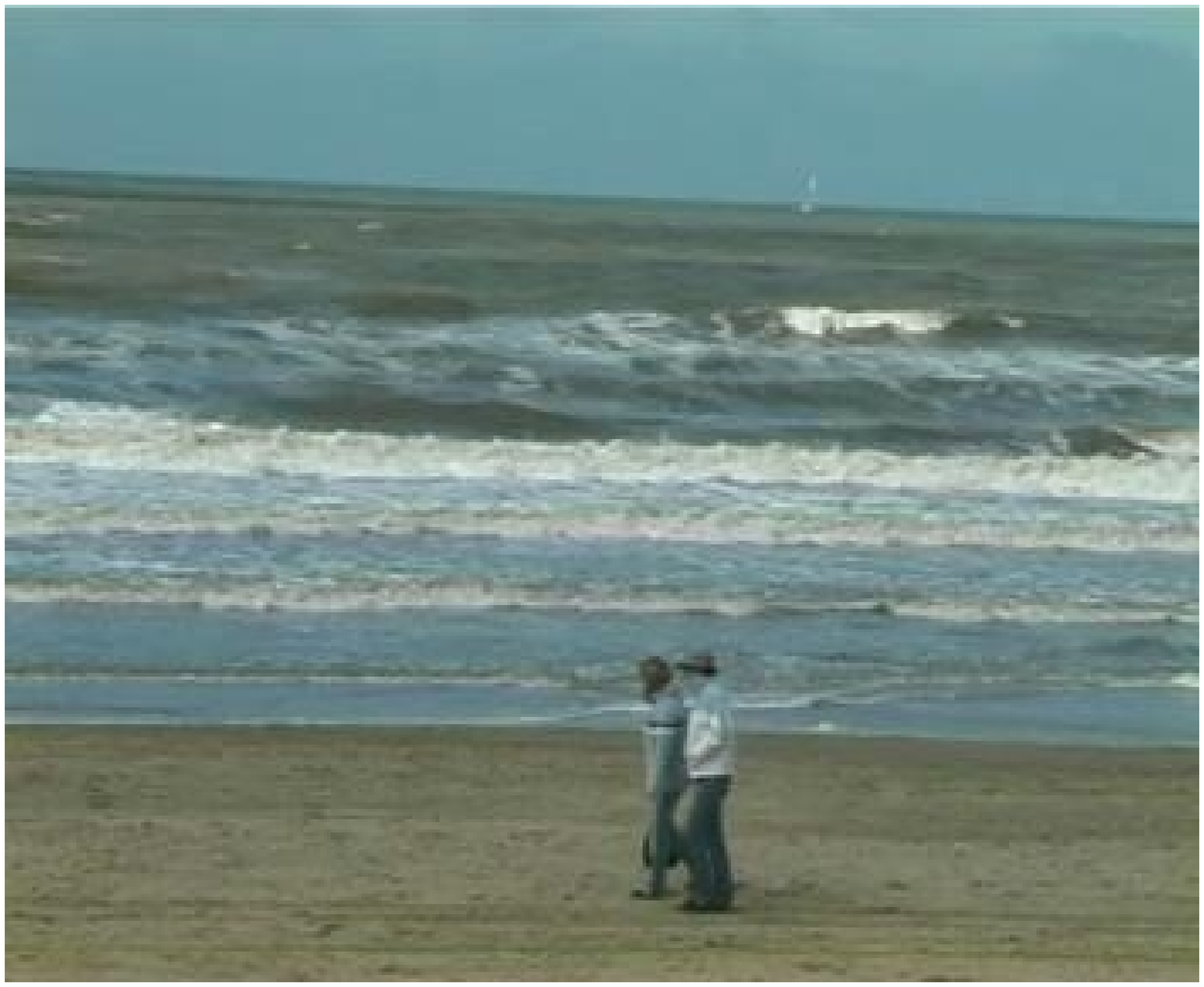}
}
\subfigure[]
{
      \includegraphics[width=0.7in]{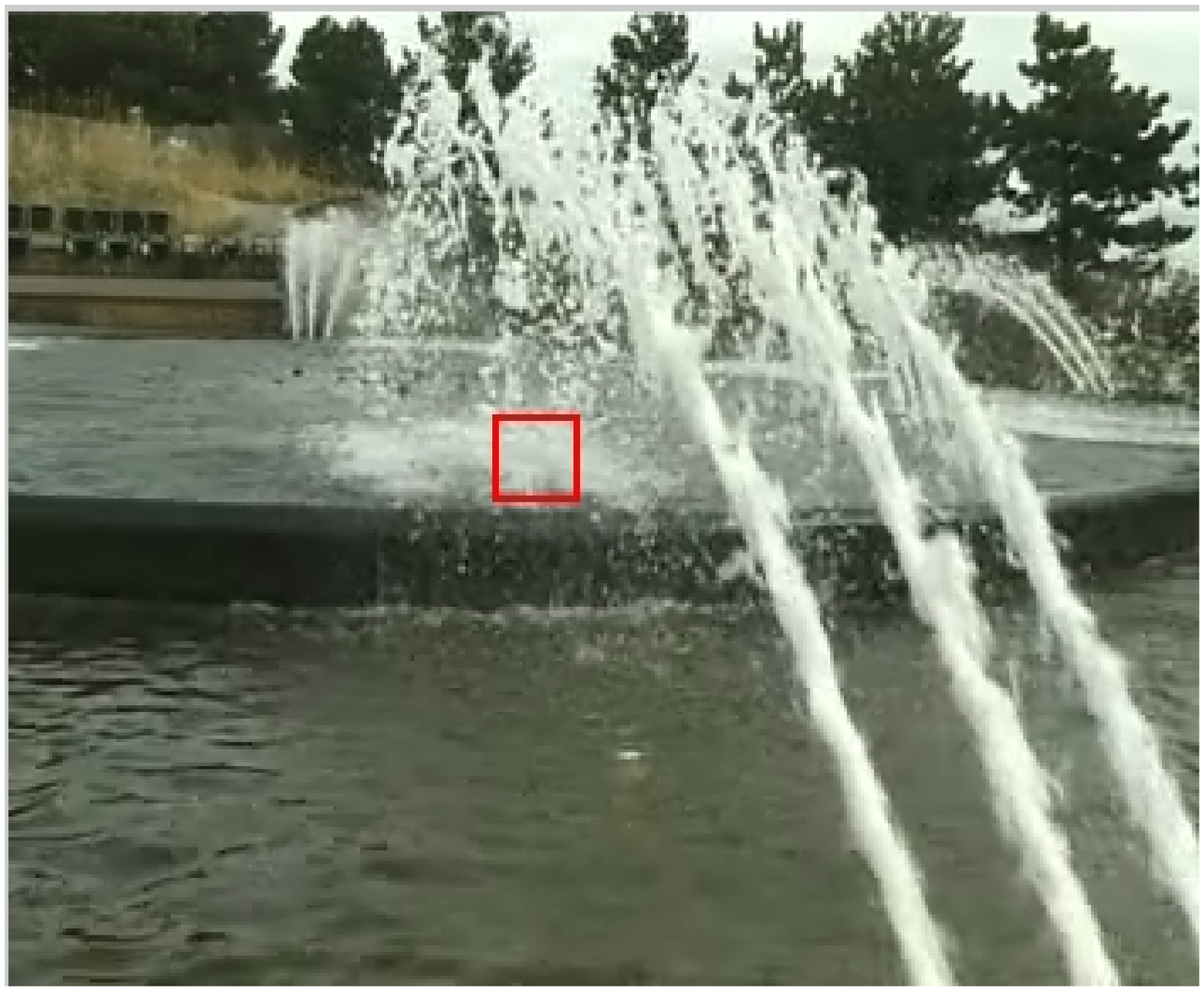}
}
\subfigure[]
{
      \includegraphics[width=0.7in]{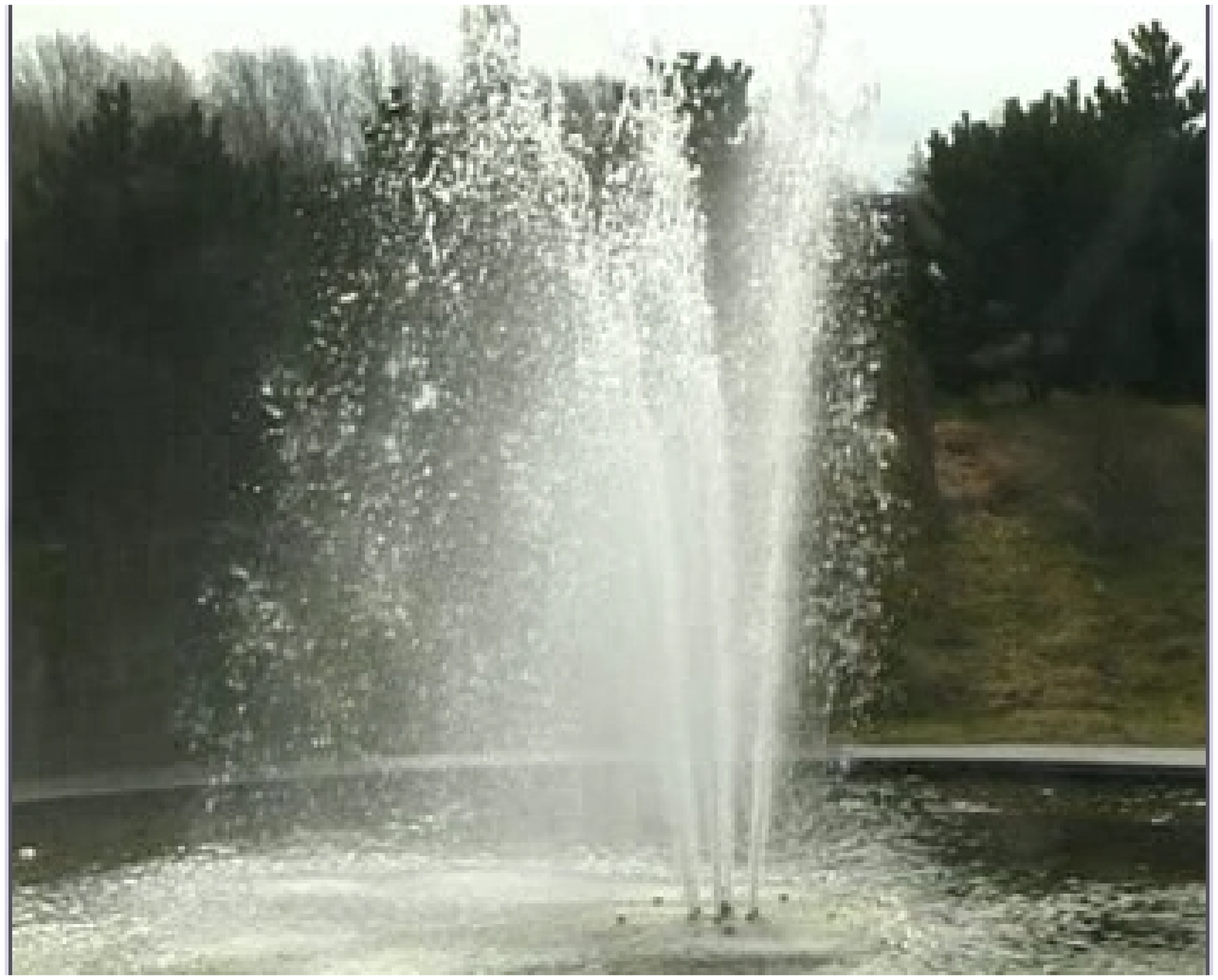}
}
\subfigure[]
{
      \includegraphics[width=0.7in]{final_result17.eps}
}
\caption{Examples of smoke detection results.}
\label{fig:FinalResults}
\end{figure*}

To evaluate the performance of the proposed smoke detection system, it is implemented by Matlab 2009a (as described in section \ref{sec:method}) and tested using a data set with 72 smoke videos and 79 non-smoke videos were used (see Fig. \ref{fig:TestVideos} for examples). Some videos of the data set were downloaded from the internet and the others were recorded by the authors. Different kinds of burning matters in various scenarios are considered. For instances, video (a) presents cotton rope smoke captured outdoors; video (b) records the smoke of burning leaves, the video (c) shows a video containing white smoke by burning dry leaves in front of groves of bamboo; video Video (d) contains a white smoke near a wasted basket. Fig. \ref{fig:FinalResults} illustrates the examples of the smoke detection results. We can see that suspected smoke areas (indicated by red rectangles) were well detected and located in the real smoke regions and the proposed method seldom produced false alarm.

\begin{table*}
\centering
\caption{Compared with other methods using smoke videos}
\label{tab:CSV}
\begin{tabular}{| p{2.5cm}| p{1.0cm}| p{1.0cm}| p{1.0cm}| p{1.0cm}| p{1.0cm}| p{1.0cm}| p{1.0cm}| p{1.0cm}|}
\hline \scriptsize Alarm at frame number & \scriptsize Video(a) & \scriptsize Video(b) &\scriptsize Video(c) &\scriptsize Video(d) &\scriptsize Video(e) &\scriptsize Video(f) &\scriptsize Video(g) &\scriptsize Video(h)\\
\hline \scriptsize Paper \cite{Yuan2012double} Method(2)&\scriptsize 75 &\scriptsize 415 &\scriptsize 509 &\scriptsize 89 &\scriptsize No alram &\scriptsize 78 &\scriptsize 74 &\scriptsize 73\\
\hline \scriptsize Proposed &\scriptsize 15 &\scriptsize 330 &\scriptsize 274 &\scriptsize 61 &\scriptsize 105 &\scriptsize 41 &\scriptsize 15 &\scriptsize 19\\
\hline
\end{tabular}
\end{table*}

\begin{table*}
\centering
\caption{Compared with other methods using non-smoke videos}
\label{tab:CNSV}
\begin{tabular}{| p{2.5cm}| p{1.0cm}| p{1.0cm}| p{1.0cm}| p{1.0cm}| p{1.0cm}| p{1.0cm}| p{1.0cm}| p{1.0cm}|}
\hline \scriptsize Number of false alarms & \scriptsize Video(i) & \scriptsize Video(j) &\scriptsize Video(k) &\scriptsize Video(l) &\scriptsize Video(m) &\scriptsize Video(n) &\scriptsize Video(o) &\scriptsize Video(p)\\
\hline \scriptsize Paper \cite{Yuan2012double} Method(2)&\scriptsize 0 &\scriptsize 0 &\scriptsize 0 &\scriptsize 0 &\scriptsize 0 &\scriptsize 180 &\scriptsize 110 &\scriptsize 73\\
\hline \scriptsize Proposed &\scriptsize 0 &\scriptsize 0 &\scriptsize 0 &\scriptsize 0 &\scriptsize 0 &\scriptsize 3 &\scriptsize 0 &\scriptsize 0\\
\hline
\end{tabular}
\end{table*}
\subsection{Comparison}
The proposed smoke detection system is compared with the method (2) presented in \cite{Yuan2012double}, which is demonstrated achieving the best performance in regard of most state-of-the-arts \cite{Toreyin2005wavelet}, \cite{yuan2008fast}. We follow the evaluation criterions adopted in \cite{Yuan2012double} including: first alarm at frame number and the number of false alarms. The videos shown in Fig. \ref{fig:TestVideos} are used for testing, the reminder of the videos are employed for training. Tabel \ref{tab:CSV} lists the first alarm comparison results of eight smoke videos. The smaller number of alarm frame indicates the earlier smoke alarm. For all the videos, the proposed method achieves earlier alarms than method \cite{Yuan2012double}. On the other hand, Tabel \ref{tab:CNSV} gives false alarm comparison results in eight non-smoke videos. Both methods haven't false alarms in Video (i), (j), (k), (I)and (m).It produces much fewer false alarms in Video(n), (o) and (p) in contrast with \cite{Yuan2012double}. Therefore, in summary, the proposed method provides better performance in smoke detection. Additionally, the average process time is 0.512267 seconds for each smoke frame and 0.451138 seconds for each non-smoke frame on a computer with Intel Core2 P8400 2.26GHZ CPU and 2GB RAM. Since a C++ implementation will 15-30 times faster than the Matlab version, the proposed algorithm can be employed for real-time vision tasks.
\section{Conclusion} \label{sec:conclusion}
In this paper, we developed a video based smoke detection system for early fire surveillance. Different kinds of texture features are evaluated using a general frame work name HEP to find the best one for smoke detection. A novel spatial-temporal feature combining block based Inter-Frame Difference ($BIFD_{CM}$) and improved LBP-TOPs (EOH-TOP, BGC3-TOP and RTU-TOP) is proposed to analyze the dynamic characteristics of the video smoke. The performance of the proposed features  is evaluated utilizing SVM classifier. Additionally, the Smoke Histogram Image (SHI) is adapted to reduce the false alarm. Experimental results show better detection accuracy and false alarm resistance are achieved compared with the state-of-the-art technologies.
\section*{Acknowledgments}
This work is supported by the National Natural Science Foundation of China (No.61003143) and the Fundamental Research Funds for Central Universities (No.SWJTU12CX094).

\end{document}